\documentclass[11pt]{article}

\usepackage{graphicx}%
\usepackage{multirow}%
\usepackage{amsmath,amssymb,amsthm,amsfonts}
\usepackage{makecell}
\usepackage{mathrsfs}%
\usepackage[title]{appendix}%
\usepackage{xcolor}%
\usepackage{textcomp}%
\usepackage{manyfoot}%
\usepackage{booktabs}%
\usepackage[ruled, lined, linesnumbered, commentsnumbered]{algorithm2e}
\usepackage{bbding} 
\usepackage{pifont} 

\usepackage{listings}%
\usepackage{natbib}
\usepackage{lmodern} 
\usepackage{anyfontsize} 
\usepackage[title]{appendix}%
\usepackage{xcolor}%
\usepackage{pdflscape}
\usepackage{kpfonts}
\usepackage{url}
\usepackage{epstopdf}
\usepackage{wrapfig}
\usepackage[colorlinks, linkcolor=blue, anchorcolor=blue, citecolor=blue]{hyperref}
\usepackage[margin=1in]{geometry}
\usepackage[normalem]{ulem}
\usepackage[export]{adjustbox}
\usepackage{mathtools, cuted}
\usepackage{enumerate}
\usepackage{enumitem}
\usepackage{subfigure}
\usepackage{rotating}

\def\cX{{\cal X}}

\def\cZ{{\cal Z}}

\newcommand{\bz}{{\bf z}}

\newcommand{\bX}{{\bf X}}
\newcommand{\bx}{{\bf x}}

\newcommand{\bmu}{\mbox{\boldmath{$\mu$}}}

\newcommand{\bc}{\begin{center}}
\newcommand{\ec}{\end{center}}
\newcommand{\be}{\begin{equation}}
\newcommand{\ee}{\end{equation}}
\newcommand{\ba}{\begin{array}}
\newcommand{\ea}{\end{array}}
\newcommand{\bean}{\begin{eqnarray*}}
\newcommand{\eean}{\end{eqnarray*}}
\newcommand{\bea}{\begin{eqnarray}}
\newcommand{\eea}{\end{eqnarray}}
\newcommand{\ben}{\begin{enumerate}}
\newcommand{\een}{\end{enumerate}}
\newcommand{\bed}{\begin{itemize}}
\newcommand{\eed}{\end{itemize}}
\newcommand{\bs}{\begin{slide}}
\newcommand{\es}{\end{slide}}

\newcommand{\hochkomma}{$^{,}$}

\begin{document}

\title{Improving SMOTE via Fusing Conditional VAE for Data-adaptive Noise Filtering}

\author{Sungchul Hong\footnote{Department of Statistics, University of Seoul.}, Seunghwan An\footnote{Department of Statistical Data Science, University of Seoul.}, and Jong-June Jeon\footnotemark[1] \hochkomma \footnote{Corresponding author. \url{jj.jeon@uos.ac.kr}}}

\date{First version: January, 2024 \\ 
Latest version: \today}
\maketitle

\abstract{Recent advances in a generative neural network model extend the development of data augmentation methods. However, the augmentation methods based on the modern generative models fail to achieve notable performance for class imbalance data compared to the conventional model, Synthetic Minority Oversampling Technique (SMOTE). We investigate the problem of the generative model for imbalanced classification and introduce a framework to enhance the SMOTE algorithm using Variational Autoencoders (VAE). Our approach systematically quantifies the density of data points in a low-dimensional latent space using the VAE, simultaneously incorporating information on class labels and classification difficulty. Then, the data points potentially degrading the augmentation are systematically excluded, and the neighboring observations are directly augmented on the data space. Empirical studies on several imbalanced datasets represent that this simple process innovatively improves the conventional SMOTE algorithm over the deep learning models.
Consequently, we conclude that the selection of minority data and the interpolation in the data space are beneficial for imbalanced classification problems with a relatively small number of data points.}\\
\noindent \textit{keywords}: SMOTE, Imbalanced classification, Variational autoencoder, Oversampling, Noise filtering

\section{Introduction} 
Class imbalance problem is a common challenge in constructing a classification model \citep{he2009learning}.  
The empirical risk minimization method tends to underrepresent the features of observations belonging to the minor class, thereby inducing bias in the learning model and detrimentally affecting its performance.
 Oversampling the minor class is a popular technique to mitigate the class imbalance problem. Oversampling involves generating synthetic instances for the minor class and resampling to balance the class distribution. Under lack of data, it is often regarded as an essential preprocessing step in various domains, including medical diagnosis, defect detection, and chemical classification \citep{mazurowski2008training, malhotra2019empirical, idakwo2020structure}.

Synthetic Minority Oversampling Technique (SMOTE) \citep{chawla2002smote} generates new minor class samples by interpolating between existing minor class samples and their nearest minority neighbors. Due to its simplicity and effectiveness, SMOTE has become a standard approach in the field of imbalanced classification \citep{fernandez2018smote}. Building on the success of SMOTE, several extensions have been proposed to further enhance classification performance, especially under scenarios involving specific data structures or mislabeling.
These extensions include Borderline-SMOTE \citep{Han2005BorderlineSMOTEAN}, and ADASYN \citep{he2008adasyn}. These methods aim to address the limitations of SMOTE and are discussed in detail in Section \ref{sec:rw}. 

The advent of effective dimensionality reduction and generative model learning for high-dimensional data has facilitated easier generation of data belonging to the minor class, leading to a significant expansion beyond conventional oversampling methods. Some of these techniques employ neural networks to enhance the oversampling performance \citep{Liu2018DeepDF, Dablain2021DeepSMOTEFD, shin2022adanoise}. Additionally, deep learning approaches efficiently augment unstructured datasets such as image and text \citep{tian2021re, dablain2023efficient}. These techniques leverage the power of deep neural networks to generate synthetic samples that are more representative of the minor class, improving the performance of models on image and text classification tasks. 

Accordingly, deep generative models (DGMs), such as Variational AutoEncoders (VAEs) \citep{kingma2013auto} and Generative Adversarial Networks (GANs) \citep{goodfellow2014generative},  are adapted to oversampling. Specifically, the VAE maps high-dimensional data into a lower-dimensional latent space by feature extraction. These approaches effectively encode the underlying characteristics of imbalanced datasets by utilizing the latent space \citep{Dai2019GenerativeOW, fajardo2021oversampling}. However,  \cite{Camino2020OversamplingTD} claims that oversampling based on DGMs does not yield substantial improvements over SMOTE, particularly in tabular datasets. They conjecture the limitation of DGM-based oversampling techniques is attributed to the inaccurate estimation of the density function in an imbalanced dataset. They raise doubts about whether learning the distribution of the major class would significantly contribute to identifying the distribution of the minor class. Motivated by the empirical evidence of \cite{Camino2020OversamplingTD}, we investigate the pattern of the augmented samples of DGMs with the Escherichia coli dataset (also known as \texttt{ecoli} dataset) \citep{misc_ecoli_39} to understand why the DGMs often fail to improve the prediction model.

\begin{figure}[t!]
    \centering
    \subfigure[Conditional VAE]{\includegraphics[width=0.48\linewidth]{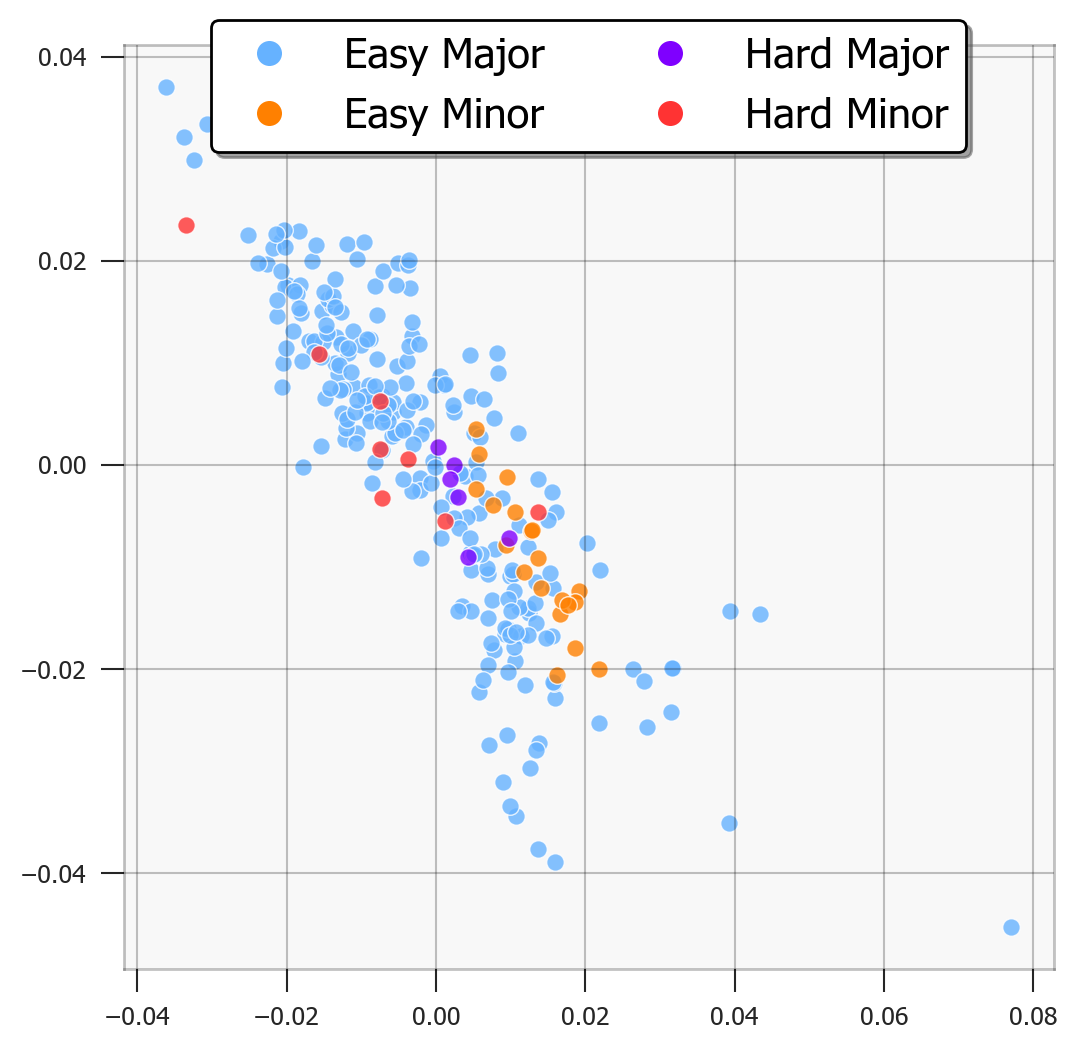}}
    \subfigure[DDHS]{\includegraphics[width=0.47\linewidth]{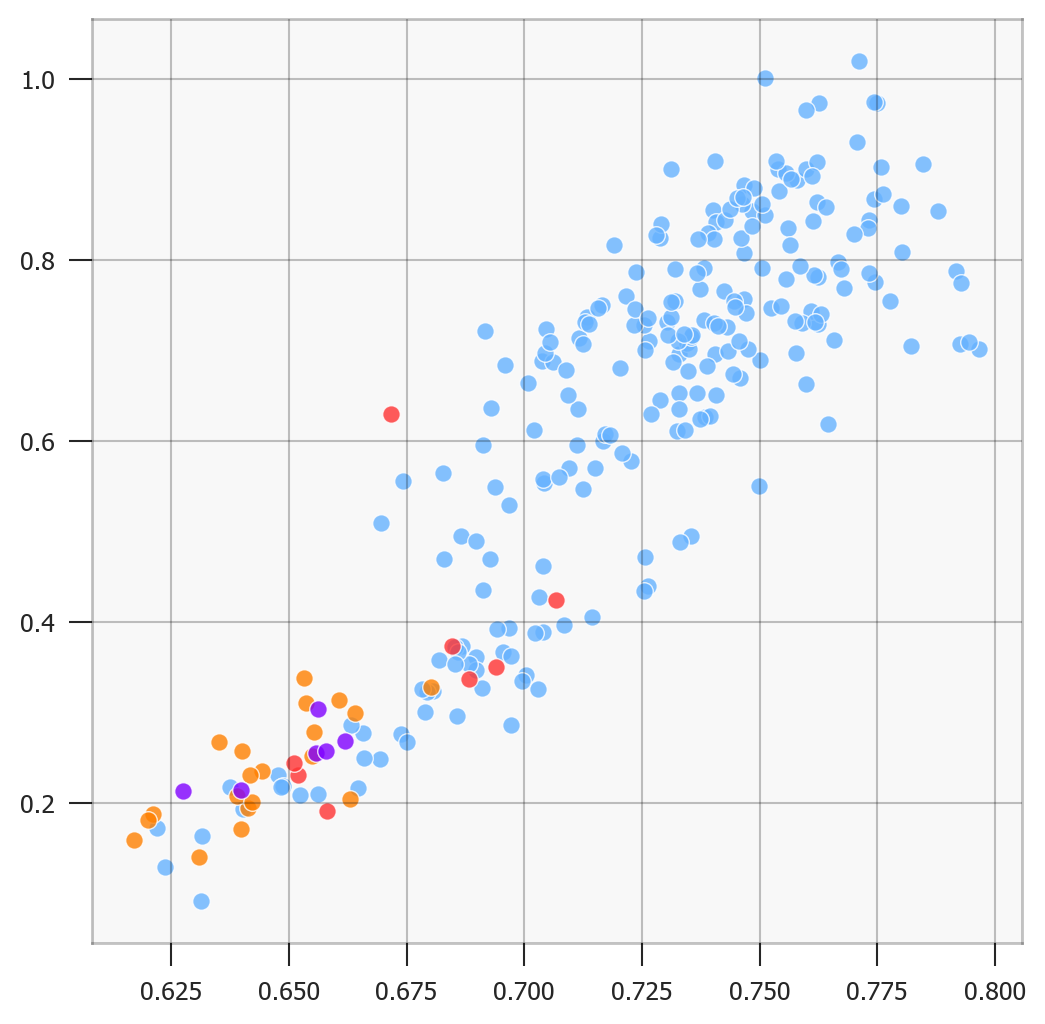}}
    \caption{Visualization of latent space of Conditional VAE and DDHS.}
    \label{fig:motiv}
\end{figure}

First, we visualize the latent space obtained by typical DGMs, the conditional VAE, and the Autoencoder (AE)-based DDHS \citep{Liu2022LearningFI}.
Figure \ref{fig:motiv} displays observations on the trained latent space. The observations are labeled as easy-major, hard-major, easy-minor, and hard-minor classes according to their classification difficulty and the class majority. The classification difficulty is quantified as a local misclassification error obtained by the K-nearest neighbor algorithm (for more detailed descriptions, refer to Section \ref{sec:vae}). We can see that the minority samples are overlapped with the major samples in the latent space. In addition, Figure \ref{fig:motiv} shows that the embedding latent space is unlikely to contain the information of the classification difficulty. Thus, SMOTE or neighborhood sampling methods often learn a distribution of major samples as that of minority samples. 

Second, the estimation of distribution for minority samples is vulnerable to noise or outliers. In particular, when the minor class exhibits within-class imbalance, naive distributional learning can mislead to oversampling, pursuing the generation of so-called high-quality synthetic instances. This challenging problem is called the small-disjunct problem \citep{he2009learning, azhar2022investigation}. In the conventional kernel density estimation studies, the adaptive bandwidth selection method is recommended for more accurate distributional learning \citep{van2003adaptive}. In both Figure \ref{fig:motiv} (a) and (b), we can see a hard-minority sample distant from the center of the minority samples.  Interpolation on the latent space between this observation and other neighborhoods produces synthetic samples far from the minor group.

The main idea underlying our oversampling method revolves around contrasting the pattern of interest. The proposed oversampling consists of three steps. 
The algorithm explores the latent space in which the majority and classification difficulty are embedded. To disentangle the mode of each class in the latent space, we employ the VAE \citep{an2024customization} capable of aligning an encoded feature of a specific class at a pre-determined prior mode. Next, with the trained latent structure, we exclude the samples that do not contain neighboring information belonging to the same class \citep{liu2023goggle}.  
A density-based filtering removes isolated observations, regarded as outliers in the latent space. 
Finally, we employ SMOTE directly to the filtered observations, not latent variables. 

The advantages of our method are listed as follows: 
\ben
    \item The proposed method involves disentangling the latent space based on various features under consideration for data augmentation with a state-of-the-art VAE. Thus, we can gain insights into the distinguished nature of labels, effectively separating latent variables based on a predefined prior distribution.
    \item We enhance the existing methods to identify and remove a noise sample by considering proximity in the separated latent space. Through filtering, we can prevent a noise sample from being involved in SMOTE so that we partly solve the low-quality problem, such as unrealistic synthetic samples and ignoring variable dependencies, by DGMs \citep{liu2023goggle}. 
    \item We demonstrate the efficiency and superiority of our proposed method. Numerical studies show that our simple disentangling and filtering step can effectively improve the classification performance of the Area Under the Precision-Recall Curve (AUPRC) and the Area Under the Curve (AUC).
\een

The remainder of this paper is organized as follows: Section \ref{sec:rw} introduces the existing oversampling method containing classical and deep learning-based approaches. Section \ref{sec:pr} explains the proposed algorithm by dividing it into two parts: customizing latent space and oversampling with filtering. Section \ref{sec:ns} shows the numerical results using synthetic and real datasets with various visualizations. Concluding remarks and limitations of this paper follow in Section \ref{sec:cl}.    

\section{Related Work} \label{sec:rw}

\subsection{Overview} 

An ideal oversampling method is to sample from the true distribution of the minor class. Thus, estimating a target distribution is a natural and intuitive approach to developing an oversampling method.
However, this can be particularly challenging in high-dimensional settings under a lack of data. Despite the notable advances of a generative model driven by deep learning, developing a novel oversampling method is discouraging.  It is important to note that deep learning-based approaches still face challenges in accurately estimating the density of the minor class, and this limitation can impede their effectiveness in oversampling tasks. We will investigate this issue in more detail in Section \ref{sec:ns}. Thus, it is common for researchers to define the specifics of their task and then apply a suitable oversampling method, as demonstrated in survey works \citep{he2009learning, azhar2022investigation}. For binary classification problems, it is generally assumed that generated synthetic samples can aid in finding the optimal decision boundary, leading to oversampling in the vicinity of the minority samples.

SMOTE \citep{chawla2002smote} is a straightforward oversampling method that linearly interpolates between two neighboring minority samples. This method samples from restricted low dimensional space (a subset of low dimensional convex hull) and achieves superior performance \citep{kamalov2020kernel}. However, SMOTE is still vulnerable to noise and is often unable to effectively address the issue of small disjuncts, which refer to small-sized clusters within the minor class and incur false negatives. 

The proposed oversampling method is an extension of existing research enhancing the performance of SMOTE. The novelty distinguished from the prior studies lies in filtering minority samples on a low-dimensional latent space, reflecting the pre-defined characteristics of data.
In particular, our proposed method focuses on selecting appropriate minority samples for oversampling and mitigating the issues associated with small disjuncts, ultimately improving the performance of imbalanced classification. In the following subsection, we introduce an extended study of SMOTE, presenting an oversampling method that utilizes deep learning as a comparative model.

\subsection{Literature review}
This section provides a comprehensive review of oversampling literature, specifically focusing on data-level oversampling approaches, including SMOTE-based, AE-based, and VAE-based methods. 

\textbf{SMOTE-based methods.}
SMOTE \citep{chawla2002smote} generates new synthetic samples by linearly interpolating between minority samples and their K-nearest neighbors among the minority samples. SMOTE is a simple and powerful augmentation method, but its theoretical properties are rarely known.
\citet{elreedy2023theoretical} analyzes the distribution of the augmented samples by SMOTE. Because SMOTE does not perform well on highly noisy datasets, \cite{Han2005BorderlineSMOTEAN} proposes Borderline-SMOTE incorporating a selection step to identify necessary minority samples known as ``borderline samples''. These samples are chosen based on neighborhood information, specifically the ratio of major samples to the total number of neighbors. 
Another approach to address class imbalance is ADASYN \citep{he2008adasyn}. A data-adaptive oversampling method assigns weights to minority samples based on their neighborhood information. This adaptive weighting scheme emphasizes the minority samples that are more challenging to classify correctly. Another approach to noise is SMOTE-ENN \citep{batista2004study}, which synthesizes minority samples by SMOTE and filters noise out by the edited nearest neighborhood (ENN). However, these approaches primarily rely on neighborhood information and do not consider the characteristics on feature levels. To address imbalance within the class, KMSMOTE is proposed by \cite{douzas2018improving}. KMSMOTE constructs clusters regardless of class by K-means clustering and then calculates the ratios of the minority samples in clusters. After clustering, KMSMOTE identifies clusters whose ratio of minority samples is too low and filters them out. However, this method has some limitations in that hyperparameters (e.g., the number of clusters) tuning is required, and its performance is sensitive to them.  

\textbf{AE-based methods.}
The neighborhood information used in SMOTE-based methods is typically induced by Euclidean distance defined on data space. The Euclidean distance is often an inadequate metric representing the proximity between data points, especially in high-dimensional cases. To address this issue, various methods leverage autoencoder architectures to learn lower-dimensional representations of the data. These lower-dimensional representations retain essential information, enabling the reconstruction of the original data. 
Many oversampling methods propose various architectures of AEs and loss functions. \citet{Dablain2021DeepSMOTEFD} proposes the AE-based oversampling algorithm, DeepSMOTE, which employs SMOTE to synthesize the latent variables with the minor label. DeepSMOTE uses a loss function based on U-statistics to minimize within-class variances, aiming to create a desirable latent space. Other methods, such as DFBS \citep{Liu2018DeepDF} and DDHS \citep{Liu2022LearningFI}, leverage the center loss \citep{wen2016discriminative} to learn discriminative latent features. They also propose novel methods for generating high-quality and various features focusing on a minor class. 

\textbf{VAE-based methods.}
Variational AutoEncoders (VAEs) have derived practical applications in generating high-dimensional datasets, and the theoretical foundations of VAEs have led to extensions and adaptations in various domains. For instance, \citet{fajardo2021oversampling} employ Conditional VAEs (CVAEs) to synthesize minority samples, while \citet{Dai2019GenerativeOW} extend VAEs for discriminative feature learning within a contrastive learning framework. In the context of multi-class classification, \citet{solomon2022data} propose a novel VAE framework designed to preserve the data structure based on their class labels. However, similar to AE-based methods, these VAE-based approaches fail to separate the latent space. Additionally, they may suffer from generating poor-quality synthetic samples due to the inherent trade-off between KL-divergence and reconstruction errors.

In contrast to the above literature, our approach introduces a two-step filtering algorithm that can be combined with various oversampling techniques. The filtering algorithm considers not only the local information of the samples themselves but also identifies anomalies within each group in the latent space.
This sophisticated filtering improves classic oversampling algorithms, making them more effective in low- and high-dimensional datasets when compared to deep learning approaches.

\section{Proposed Method} \label{sec:pr}
This section introduces SMOTE-CLS (SMOTE with Customizing Latent Space), a novel oversampling technique that combines SMOTE with a customized latent space generated by a conditional VAE. Section \ref{sec:vae} elaborates on the approach used to handle the sample difficulties and manipulate the latent space. Following that, in Section \ref{sec:os}, we will describe our oversampling methodology, which relies on utilizing the trained latent space. 

\subsection{Customizing Latent Space with VAE} \label{sec:vae}

First, we introduce the notations used throughout this paper. Let $\bx_i \in \mathcal{X}$ and $y_i \in  \mathcal{Y} = \{M, m\}$ for $i=1, \cdots, n$ be the predictors and labels where $M$ and $m$ are labels associated with the major and minor classes, respectively. 
Inspired by \cite{Han2005BorderlineSMOTEAN} and \cite{ sorscher2022beyond} identifying a sample difficulty, we categorize the samples into two distinct groups again: ``easy sample'' and ``hard sample'' based on their local information that easy samples are more prototypical with their neighborhoods while hard samples are less prototypical.

To obtain the information straightforwardly, we employ the K-nearest neighbors (KNN) classifier, denoted as $f_K:\mathcal{X} \mapsto \mathcal{Y}$, and define the relabeling map $g:\mathcal{X} \times \mathcal{Y} \mapsto \mathcal{Y}^* = \{M^*, m^*, M, m \}$ in \eqref{eq: sample_diff}.
\bea \label{eq: sample_diff}
g(\bx, y; f_K) = \begin{cases}  M^*, & \mbox{if} ~ y = M ~ \mbox{and} ~ f_K(\bx) \neq y \\ 
                            m^*, & \mbox{if} ~ y=m ~ \mbox{and} ~ f_K(\bx) \neq y\\ 
                            M, & \mbox{if} ~ y=M ~ \mbox{and} ~ f_K(\bx) = y\\ 
                            m, & \mbox{if} ~ y=m ~ \mbox{and} ~ f_K(\bx) = y\\ 
\end{cases}.
\eea
 Let a pseudo label indicating both the class and the classification difficulty by the KNN be
$$\tilde y_i = g(\bx_i, y_i; f_K)$$ for $i = 1, \cdots, n $. Consequently, the labels of data, $\tilde y_i$s, are refined by hard-major $M^*$, hard-minor $m^*$, easy-major $M$, and easy-minor group $m$, respectively. Accordingly let $D_M^* = \{ i | ~ \tilde y_i = M^* \}$, $D_m^* = \{ i | ~ \tilde y_i = m^* \}$, $D_M = \{ i | ~ \tilde y_i = M \}$, and $D_m = \{ i | ~ \tilde y_i = m \}$.
\begin{figure}[t]
    \centering
    \includegraphics[width=0.7\linewidth]{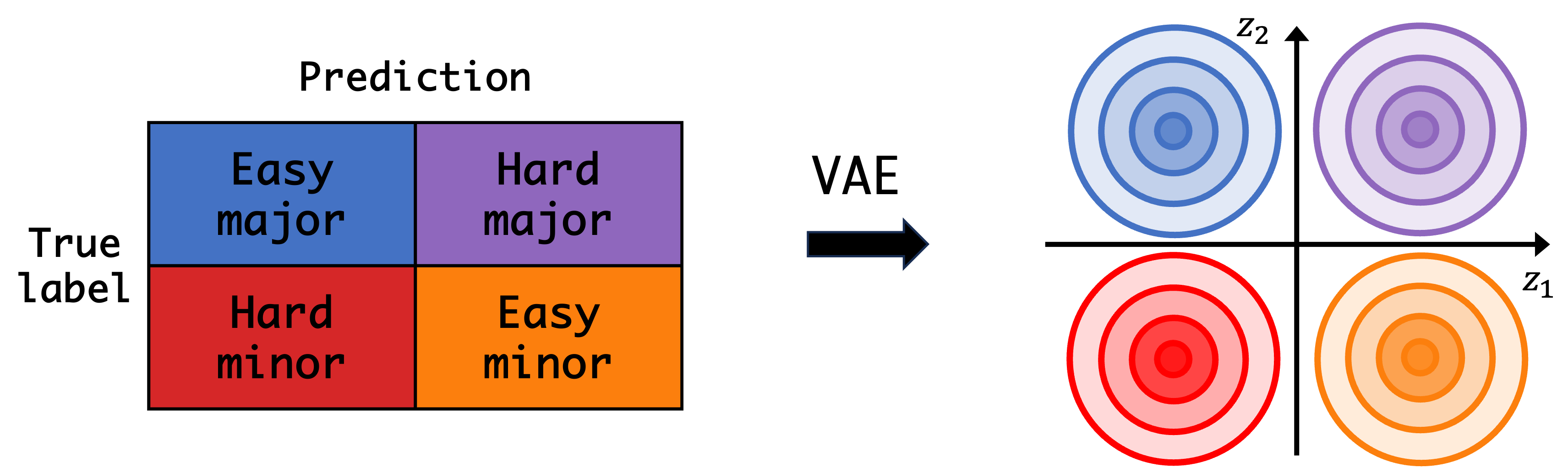}
    \caption{An example of constructing prior distribution. Classification confusion matrix generated by the KNN classifier (left), the prior distribution, $p(\bz|y)$ (right). We set the expectation of prior distribution as follows: $\bmu_{M} = (-1, 1), \bmu_{m} = (1, -1), \bmu_{M^*} = (1, 1)$, and $\bmu_{m^*} = (-1, -1)$.}
    \label{fig: ex_prior}
\end{figure}

Building upon the previous relabelling, we segment the latent space with major and minor labels and their classification difficulty. We assume that easy and hard samples possess distinct characteristics preserved in the latent space within the major and minor classes. Denote the latent space by $\cZ$. For each label $c \in \mathcal{Y}^*$, we assume $\bz | \tilde y = c \sim \mathcal{N}(\bmu_c, diag(s_c^2))$. The marginalized prior distribution $p(\bz)$ is a Gaussian Mixture Model (GMM) as follows: 
\bean
p(\bz) = \sum_{c\in \mathcal{Y}^*} p(\tilde y=c) \cdot  \mathcal{N}(\bmu_c, diag(s_c^2)). 
\eean 

The conditional VAE with the priors allows us to manipulate the latent space based on the classification difficulty defined in  \eqref{eq: sample_diff}. Figure \ref{fig: ex_prior} presents an example of latent space by sample difficulty. In Section \ref{sec:ablation}, we investigate the effect of structural constraint on the prior distribution. In addition, this conditional VAE is different from the conventional one, such as CVAE \citep{sohn2015learning}, because it utilizes the pseudo labels for a specific disentanglement of the latent space. 

We employ a encoder $q_{\phi, \eta}(\bz|\bx)$ to approximate $p(\bz|\bx)$.
\bea \label{eq: posterior2}
q_{\phi, \eta}(\bz|\bx) = \sum_{c\in \mathcal{Y}^*} f_\eta( \tilde y=c|\bx) \cdot q_{\phi}(\bz| \tilde y=c, \bx), \label{eq: proposal} 
\eea
where $f_\eta$ is a classifier with a parameter $\eta$. Each component of Gaussian mixture \eqref{eq: proposal} is defined as $q_{\phi}(\bz | \tilde{y} = c, \bx) = \mathcal{N}(\bz | \bmu_c(\bx; \phi), diag(\sigma^2_{c}(\bx; \phi)))$. Then the evidence lower bound (ELBO) of $p(\bx, \tilde y)$ is derived by
\bean
&& \log p (\bx, \tilde y ; \theta, \sigma) \\ 
&=& \log p(\bx;\theta, \sigma) + \log p(\tilde y|\bx; \theta, \sigma) \\ 
&\geq& \mathbb{E}_q [\log p (\bx|\bz; \theta, \sigma)] - D_{KL}(q_{\eta, \phi}(\bz|\bx) \| p(\bz))  + \log p(\tilde y|\bx ; \theta, \sigma)  \\ 
&\simeq& \mathbb{E}_q [\log p (\bx|\bz; \theta, \sigma)] - D_{KL}(q_{\eta, \phi}(\bz|\bx) \| p(\bz)) +  \log f_\eta(\tilde y|\bx), 
\eean
where the decoder $p(\bx|\bz; \theta, \sigma)$ is assummed as Gaussian distribution, $\mathcal{N}(D(\bz;\theta), diag(\sigma^2))$, and $D(\cdot;\theta)$ is a neural network with parameter $\theta$. \\
Because $D_{KL}(q(\bz|\bx; \eta, \phi) \| p(\bz))$ in the ELBO does not possess a closed-form solution, we use its upper bound, $D_{KL}^U(q_{\eta, \phi}(\bz|\bx) \| p(\bz))$ instead as follows
\bean
&& D_{KL}^U(q_{\eta, \phi}(\bz|\bx) \| p(\bz)) \nonumber  \\ 
&=&  \sum_{c=1}^C f_\eta( \tilde y=c | \bx)D_{KL}(q_{\eta, \phi}(\bz|\bx, \tilde y=c) \| p(\bz| \tilde y=c))+ D_{KL}(f_\eta(\tilde y|\bx)\| p(y))
\eean
Therefore, our objective function of VAE is given by 
\bea \label{eq: obj}
L(\theta, \eta, \phi) = \mathbb{E}_q [\log p (\bx|\bz; \theta, \sigma)] - D_{KL}^U(q_{\eta, \phi}(\bz|\bx) \| p(\bz)) + \log f_\eta(\tilde y|\bx).
\eea

\begin{figure}
    \centering
    \includegraphics[width=0.7\linewidth]{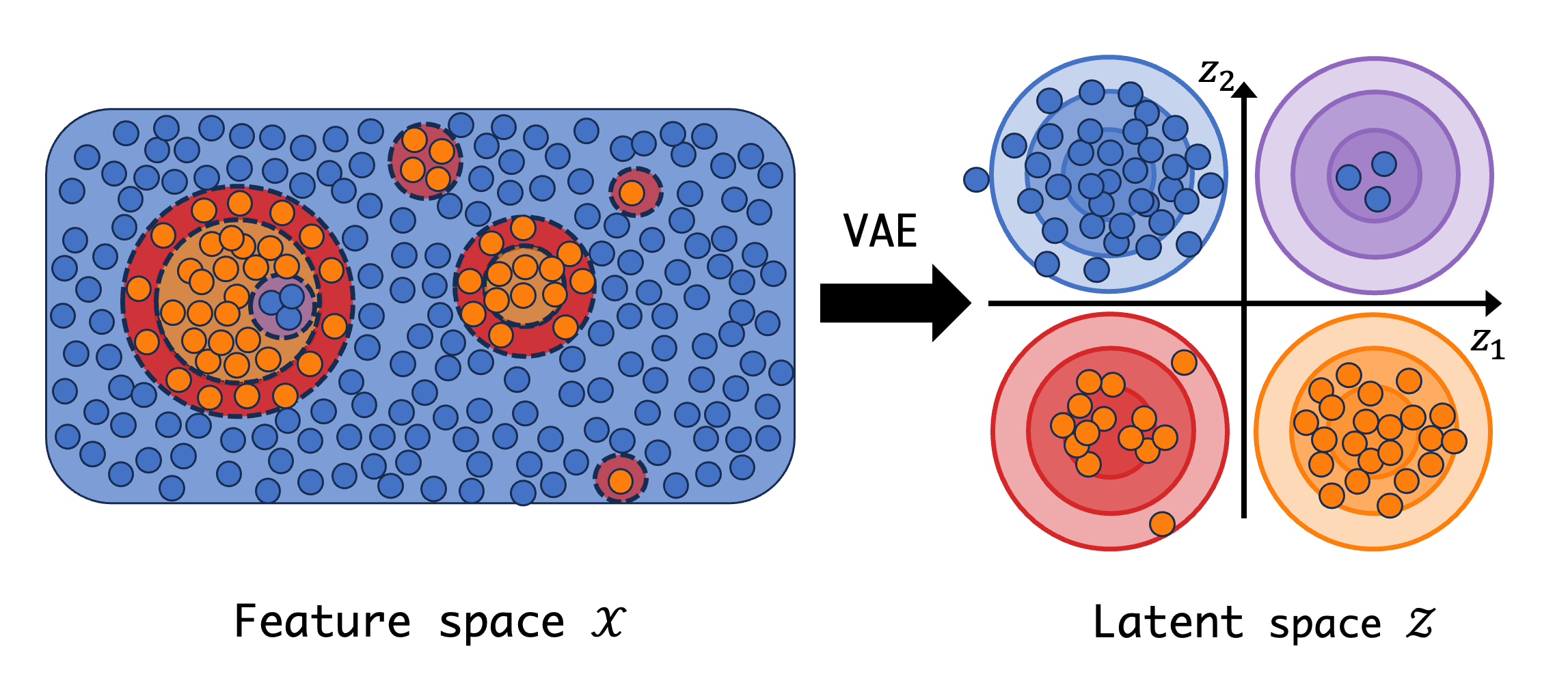}
    \caption{An example of the process of customizing latent space based on the classification difficulty from original feature space $\mathcal{X}$ (left) to latent space $\mathcal{Z}$ (right).  Blue and orange points denote major and minor samples. The classification difficulty defined by \eqref{eq: sample_diff} corresponding to the colors as follows: \textcolor{blue}{blue}: easy-major, \textcolor{orange}{orange}: easy-minor, \textcolor{violet}{purple}: hard-major, \textcolor{red}{red}: hard-minor.}
    \label{fig:vae_process}
\end{figure}

In our approach, we train the VAE to maximize the objective function \eqref{eq: obj}. Following the semi-supervised learning perspective, the classifier $f_\eta$ in \eqref{eq: obj} is simultaneously trained with the VAE, and the latent space is disentangled by the trained classifier predicting $\tilde y$. Since $\tilde y$ contains the information on classification difficulties and labels, the trained latent space can be constructed more elaborately to reflect features of an isolated observation. Furthermore, 
this study investigates the impact of $f_\eta$ on modeling the latent space (see Appendix \ref{app:cls}). We use tree-based classifiers to enhance performance, particularly in tabular datasets, such as a random forest \citep{breiman2001random} or XGBoost \citep{Chen2016XGBoostAS}. This approach improves performance and reduces the computational cost of training neural networks. Figure \ref{fig:vae_process} provides a visual summary of the process discussed in Section \ref{sec:vae}.

\subsection{Oversampling with Filtering} \label{sec:os}
To generate high-quality minority samples, selecting high-quality candidates for oversampling is essential. The definition of `high quality' often involves utilizing distance or density measures to distinguish between instances, as demonstrated in previous studies \citep{Liu2018DeepDF, xie2020gaussian, Liu2022LearningFI}. However, these existing approaches can be overly conservative, focusing solely on easy samples or sensitive to outliers, rendering them vulnerable to noise. For instance, \citet{Liu2018DeepDF} proposes a selection method as a high-quality candidate for oversampling with a distance between a sample and each class centroid. This approach is susceptible to noise samples because the centroid is sensitive to outliers (see Section \ref{sec:simul}), and the centroid inaccurately represents the locational characteristic of a mixture distribution.

\citet{Liu2022LearningFI} defines high-quality minority samples as the minor class observations with a high density. As a result, only easy samples, in our definition labeled \eqref{eq: sample_diff} by $\tilde y = m$, are selected, and hard samples tend to be excluded in the sampled candidates (see Section \ref{sec:simul}). To avoid this problem and simultaneously identify noise samples, we modulate the filtering process with a density function for each of two classes, $m^*$ and $m$ in \eqref{eq: sample_diff}. For example, the latent variables labeled by $m^*$ form clusters by a predefined prior distribution $p(\bz|\tilde y = m^*)$ (See Figure \ref{fig: ex_prior}). Consequently, we identify noise samples as outliers from each cluster, and their noise level is assessed based on each density.

Despite a specification of the prior distribution in VAE, the posterior distribution $q_{\phi, \eta}(\bz|\bx)$ may not perfectly align with the predefined priors. As a solution, previous works such as \cite{pmlr-v80-achlioptas18a} and \cite{dai2018diagnosing} employed a two-stage approach, utilizing GMM on the latent space. Additional estimation of the posterior enables a more accurate diagnosis of latent variables.

Building on the motivation from prior works, we incorporate the Kernel Density Estimation (KDE) on the latent space with the filtering process. We estimate the density of hard and easy-minority samples by KDE to consider the distributional property of each group. Let latent variables from the easy-minority samples be 
$$\bz_i = \bmu_c(\bx_i; \phi)),$$
where $c = \mbox{argmax}_k f_{\eta}(\tilde y = k|\bx_i)$ for $i \in D_m$. Recall that $\bmu_c(\bx_i; \phi))$ and $f_{\eta}(\tilde y = k|\bx_i)$ are the mean vector and classifier in \eqref{eq: posterior2}. Let $\hat{f}_m(\bz)$ be the estimated density of the latent variables of easy-minority samples by KDE. The bandwidth of KDE is determined using a conventional bandwidth selection method, Scott's rule \citep{Scott1979OnOA} to ensure more accurate density estimation, particularly when dealing with varying numbers of samples \citep{ghosh2004optimal}. Similarly, we estimate the densities of the latent variables from the hard-minority samples, $\hat{f}_m^*(\bz)$ by the KDE. Using threshold $\tau$ to filter out noise, we select minor latent variables whose densities are larger than $\tau$s in each group. 
This difficulty-adaptive filtering scheme ensures the robustness of our algorithm to the parameters of predefined priors. Please refer to Section \ref{sec:ablation} and Appendix \ref{app:thr} for further exploration and details. In the implementation, we set $\tau_1$ and $\tau_2$ below which the observations with density $\hat{f}_m(\bz)$ are excluded in the easy and hard-minor classes. The K-fold cross-validation framework can assist in selecting optimal threshold values $\tau$s empirically (see Section \ref{sec:noise}). 

\begin{figure}[t]
    \centering
    \includegraphics[width=0.99\linewidth]{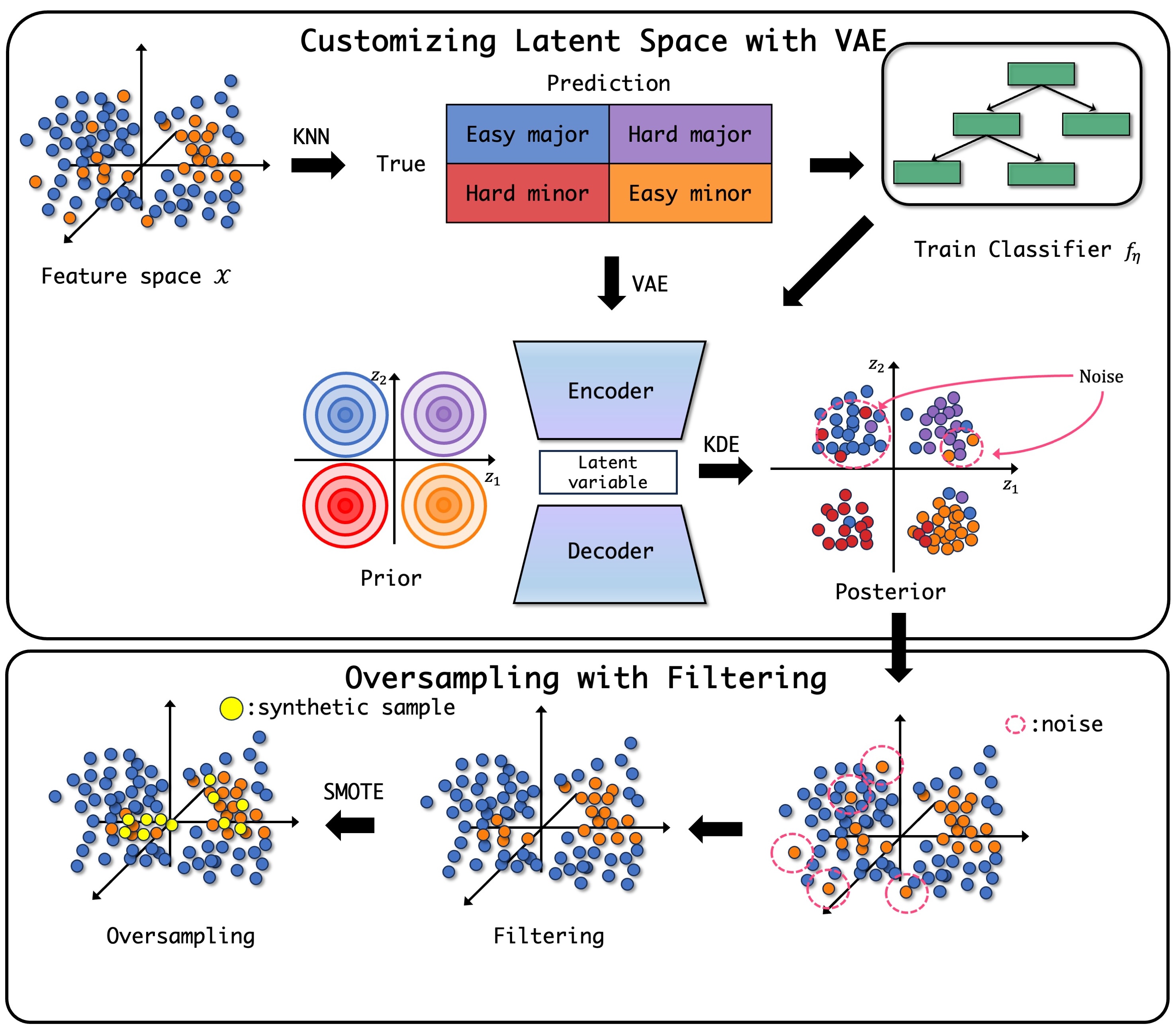}
    \caption{Oversampling process of our proposed method. The entire process is divided into two main stages. The first box (top) outlines the process of customizing the latent space and detecting noise using VAE and KDE. The second box (bottom) illustrates the subsequent noise filtering and oversampling process.}
    \label{fig:process}
\end{figure}

Once the filtering process is completed, let a subset of the minority samples with selected features denoted as $\tilde{\bX}_m$:
\[ \label{eq:gf}
\tilde{\bX}_m = \{\bx_i ~|~ \hat{f}_{m}(\bz_i) > \tau_1, i \in D_m \} \cup \{\bx_i ~|~ \hat{f}_{m^*}(\bz_i) > \tau_2, i \in D_m^* \}.
\]
Finally, we apply the conventional SMOTE on $\tilde{\bX}_m$ to generate synthetic minority samples. The entire process of our method is summarized in Algorithm \ref{alg:1} and Figure \ref{fig:process}. 

\begin{algorithm}[!ht]
\SetKwInOut{Input}{Input}\SetKwInOut{Output}{Output}\SetKwInOut{Data}{Data}
\SetAlgoLined
\Input{Model parameters: $\theta, \phi$; Prior distribution $p(\bz)$; Learning rate: $\alpha$; KNN classifier: $f_{K}$; Classifier: $f_\eta$; KL-divergence weight: $\beta$;  Thresholds: $\tau_1$, $\tau_2$; oversampling ratio: $\rho$.}
\Output{Augmented minor set.}
\textbf{Identifying Sample's Difficulty}\\
\For{$i = 1, \dots, n$}{
    $\tilde{y}_i = g(\bx_i, y_i; f_K)$
}
Let $D_m^* = \{ i | ~ \tilde y_i = m^* )\}$, $D_m = \{ i | ~ \tilde y_i = m )\}$, 
$D_M^* = \{ i | ~ \tilde y_i = M^* )\}$, and $D_M = \{ i | ~ \tilde y_i = M )\}$. \\
\textbf{Train the Classifier $f_\eta$}. \\ 
\textbf{Train the VAE:}
\For{$e \leftarrow$ epochs}{
    Draw a random subset $\mathcal{B} \subset \{1, \dots, n\}$ \\ 
    \For{$j \in \mathcal{B}$}{
        $\bz_j \sim q_{\eta, \phi}(\bz | \bx_j)$\\ 
        $\hat{\bx}_j \leftarrow p(\bx|\bz_j; \theta, \sigma)$ \\
    }
    $L = \sum_{j\in\mathcal{B}}|\bx_j-\hat{\bx}_j| + \beta \cdot D_{KL}^U(q_{\eta, \phi}(\bz|\bx_j) \| p(\bz))$\\
    $\theta \leftarrow \theta - \alpha \frac{\partial L}{\partial \theta}$ \\ 
    $\phi \leftarrow \phi - \alpha \frac{\partial L}{\partial \phi}$
}
\textbf{Group Adaptive Filtering:} \\
Define the latent variable of $\bx$, $\bz$ as follows: \newline
$\bz = \bmu_c(\bx; \eta)$ where $c=\mbox{argmax}_k f_{\eta}(\tilde y = k|\bx)$.
Estimate $\hat{f}_m$ and $\hat{f}_m^*$, the densities of latent variables of easy and hard-minority samples, by KDE.
Filter out noise samples based on probability densities with thresholds $\tau$s and define the selected features of minor set $\tilde{\bX}_m$ as follows:
\[
\tilde{\bX}_m = \{\bx_i ~|~ \hat{f}_{m}(\bz_i) > \tau_1, i \in D_m \} \cup \{\bx_i ~|~ \hat{f}_{m^*}(\bz_i) > \tau_2, i \in D_m^* \}.
\]

\textbf{Oversampling with SMOTE:} \\
$\hat{\bX}_m \leftarrow \tilde{\bX}_m$ \\
\While{$|\hat{\bX}_m| < \rho \cdot |D_M^* \cup D_M|$}{
$\hat{\bX}_m \leftarrow \hat{\bX}_m \cup \{SMOTE(\bx)\}, \bx \in \tilde{\bX}_m$ \\ 
}
\textbf{Return:} $\hat{\bX}_m$
\caption{SMOTE-CLS Algorithm}
\label{alg:1}
\end{algorithm}

\section{Numerical Studies} \label{sec:ns}
In this section, we conduct numerical studies using synthetic, tabular, and image datasets. The simulation study with the synthetic dataset aims to investigate how our proposed method identifies noise samples within the minor group and effectively filters them out. For the tabular datasets, we compare the imbalanced classification performance of our method against other competitive oversampling techniques.
Finally, we examine the robustness of our filtering algorithms against labeling noise using image datasets.
Our experiments are implemented with \textsf{Python} on a Mac equipped with an Apple M2 Max processor and 32GB of RAM. 

\subsection{Simulation study} \label{sec:simul}
\begin{figure}[t]
    \centering
    \subfigure[Synthetic samples with noises]{\includegraphics[width=0.49\linewidth]{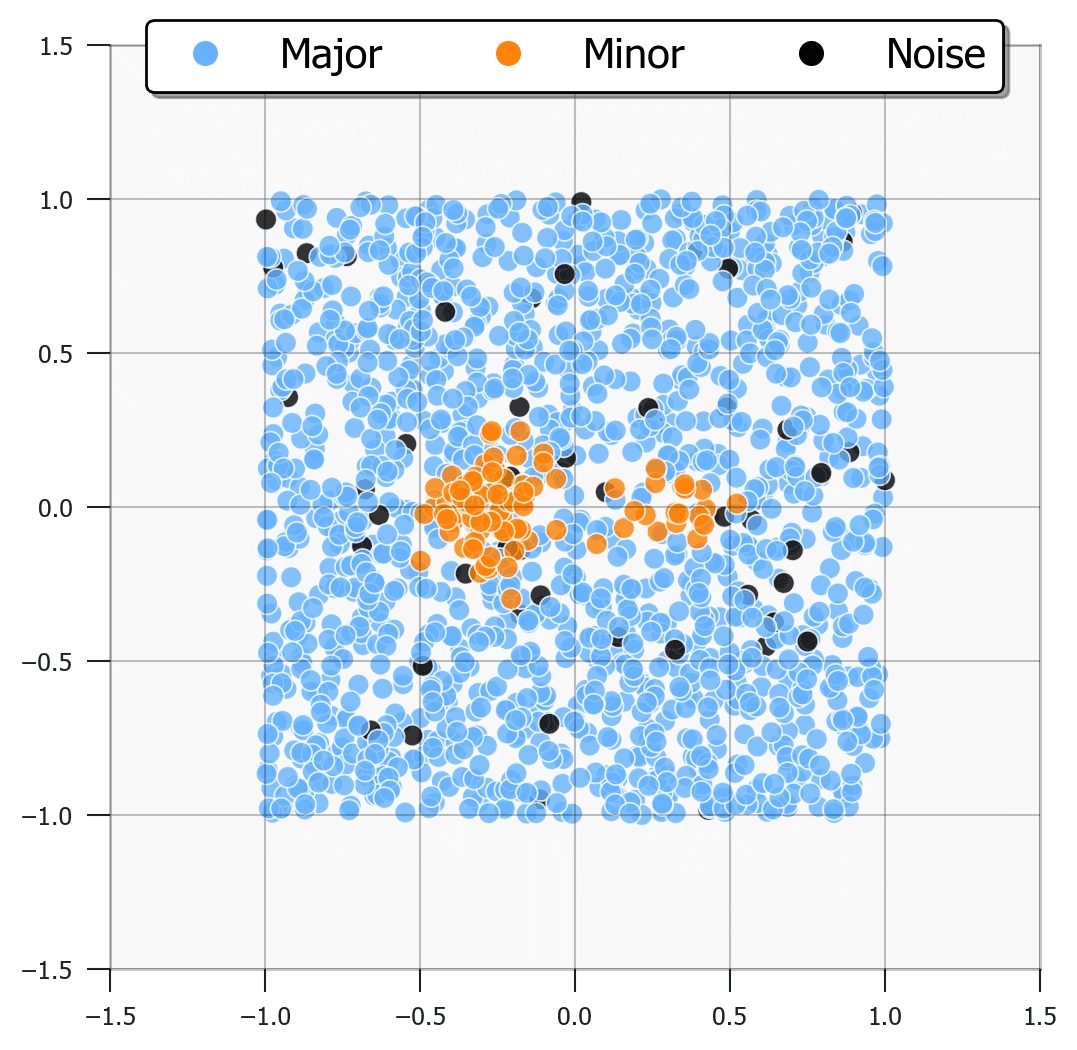}}
    \subfigure[Decision boundary of Bayes classifier]{\includegraphics[width=0.49\linewidth]{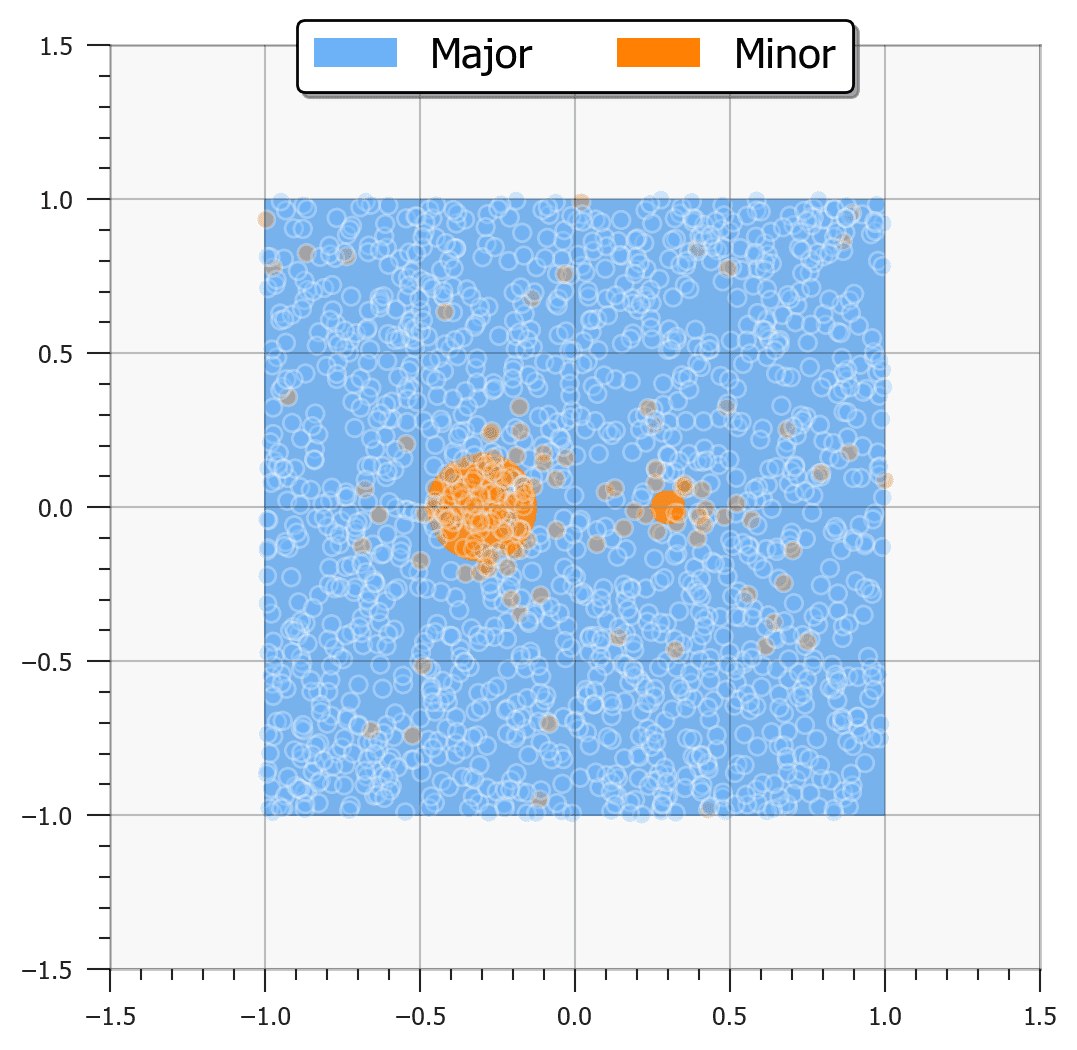}}
    \caption{Visualization of the synthetic dataset. The sky blue, orange, and black points are major, minor, and noise samples.}
    \label{fig:simul}
\end{figure}

We consider a scenario in which the minor class exhibits within-class imbalance that comprises both large and small-sized clusters \citep{he2009learning, liu2023novel}. The primary challenge in the small-disjunct problem lies in distinguishing small-sized clusters from noise instances.

To produce our synthetic minority samples, we generate data from two Gaussian distributions as follows: 
\bean
G_1 = \mathcal{N}((-0.3, 0)^\top, diag(0.01)),
G_2 = \mathcal{N}((0.3, 0)^\top, diag(0.01)). 
\eean
Subsequently, we generate 80 samples from $G_1$ and 20 samples from $G_2$. We generate 1,500 samples from a uniform distribution on $[-1, 1]^2$ for the major class. Through label-swapping, we artificially generate 50 noise samples from the major set. These noise samples can be regarded as label-contaminated or outliers in the minor group.

Figure \ref{fig:simul} (a) displays the samples generated according to our simulation settings, while (b) illustrates a true Bayes decision boundary. In Figure \ref{fig:simul} (b), you can observe that samples from $G_1$ dominate the decision region of the minor class, and the ones from $G_2$ and noise samples are excluded from the decision region. Through various oversampling methods, we investigate how the decision boundary changes. We conduct a comparative analysis of our proposed method, SMOTE-CLS, against other deep learning approaches, including DFBS \citep{Liu2018DeepDF}, DeepSMOTE \citep{Dablain2021DeepSMOTEFD}, and DDHS \citep{Liu2022LearningFI} and employ a decision tree as the baseline classifier. To achieve a balanced label ratio, 1400 minority samples are generated by oversampling techniques. The prior distribution of SMOTE-CLS 
has mean vectors as $\bmu_{M} = (-1, 1), \bmu_{m} = (1, -1), \bmu_{M^*} = (1, 1)$, and $\bmu_{m^*} = (-1, -1)$, and variance as $s_c^2 = 0.1$ for $c \in \mathcal{Y}^*$. Additionally, we set $K=5$ for the KNN classifier and set $\tau_1$ and $\tau_2$ to be selected  90\% and 60\% samples in $D_m$ and $D^*_m$ according to \eqref{eq:gf}. XGBoost is employed as the function $f_\eta$. 

\begin{figure}[!ht]
    \centering
    \subfigure[DFBS]{\includegraphics[width=0.24\linewidth]{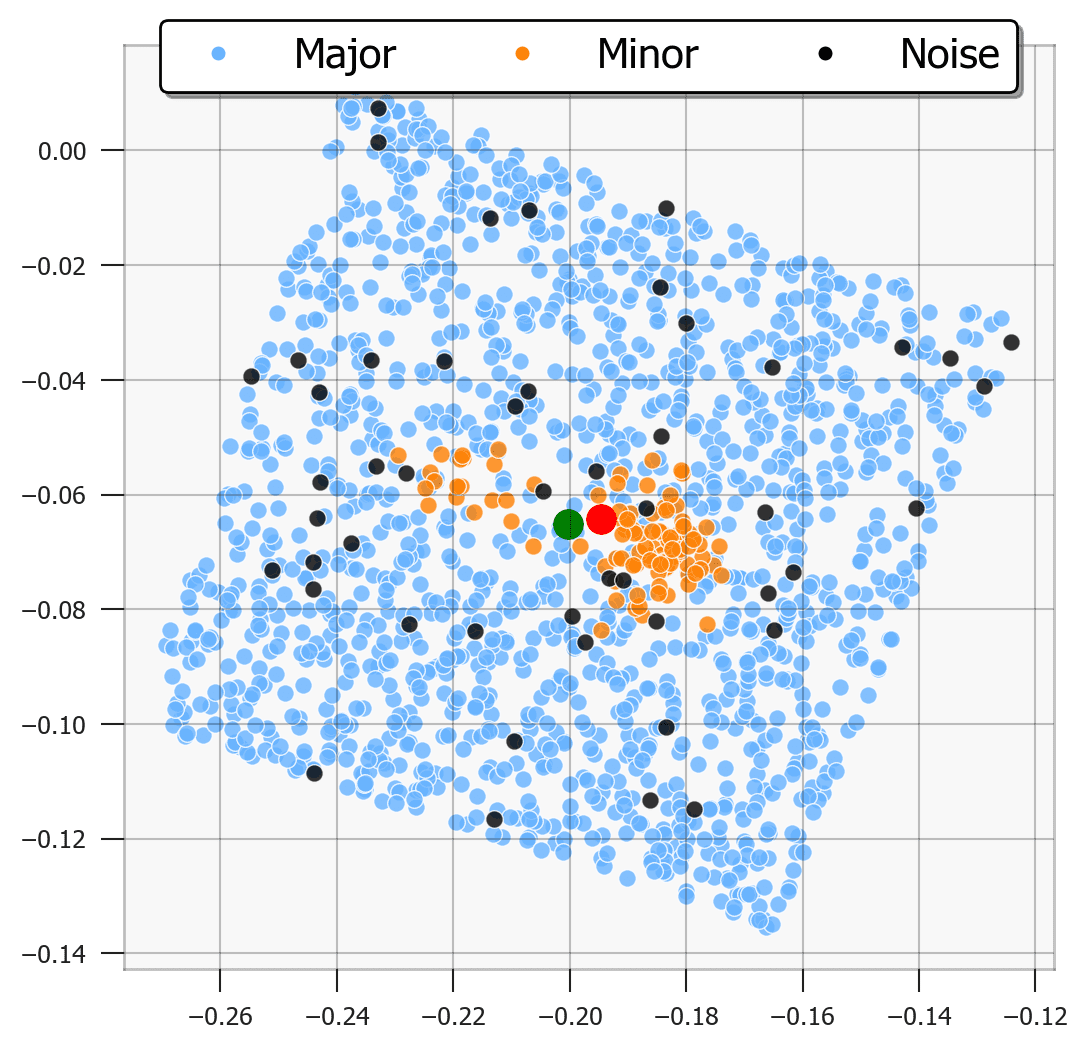}}
    \subfigure[DeepSMOTE]{\includegraphics[width=0.24\linewidth]{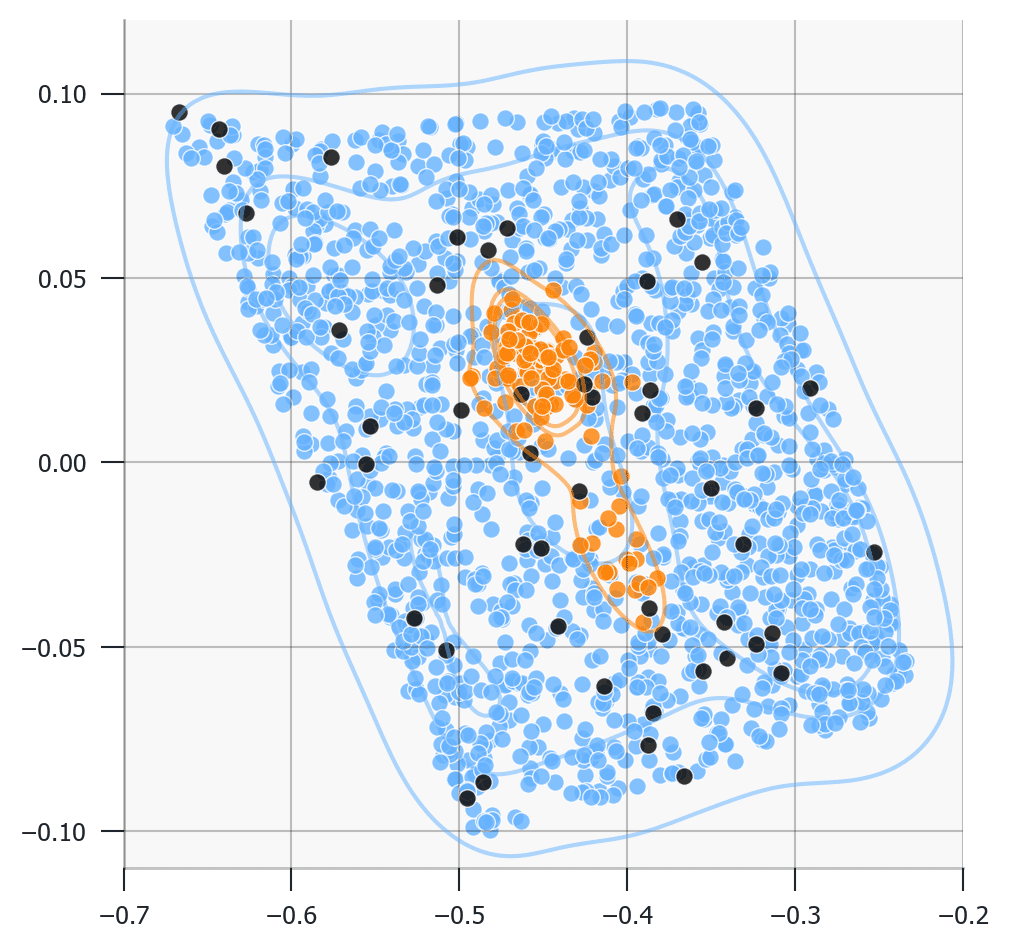}}
    \subfigure[DDHS]{\includegraphics[width=0.24\linewidth]{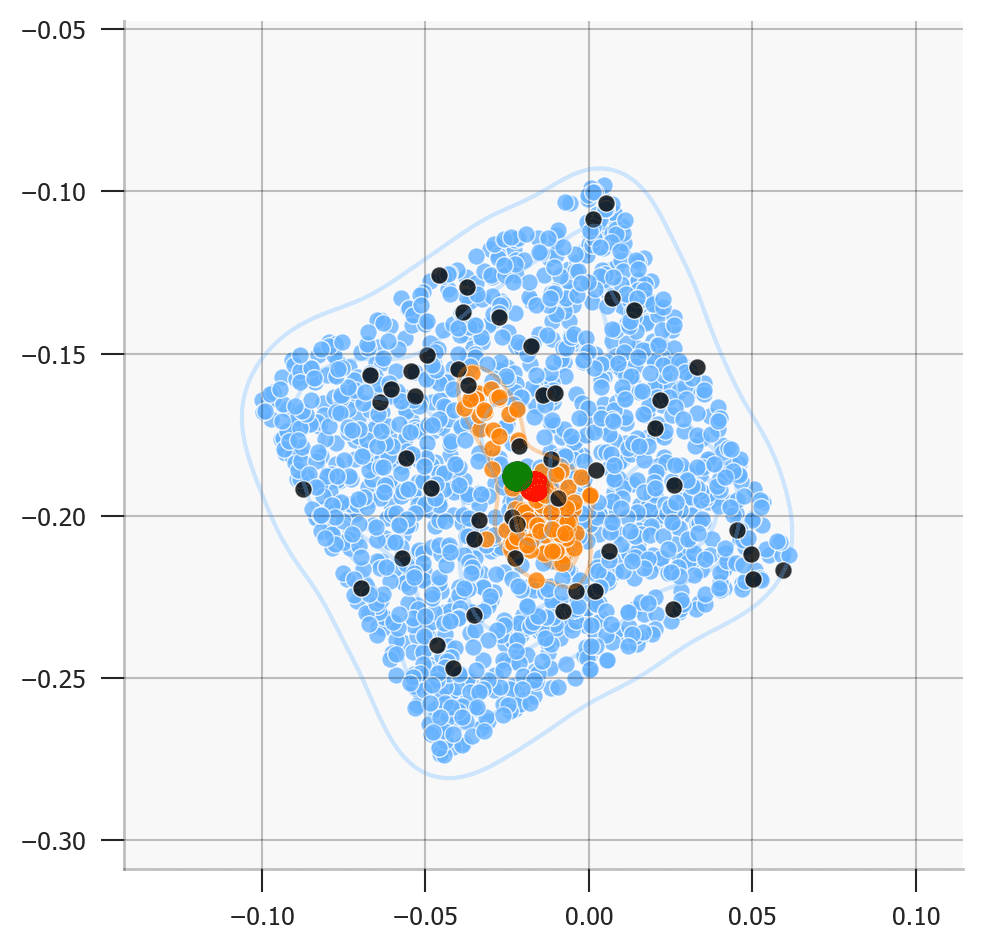}}
    \subfigure[SMOTE-CLS]{\includegraphics[width=0.24\linewidth]{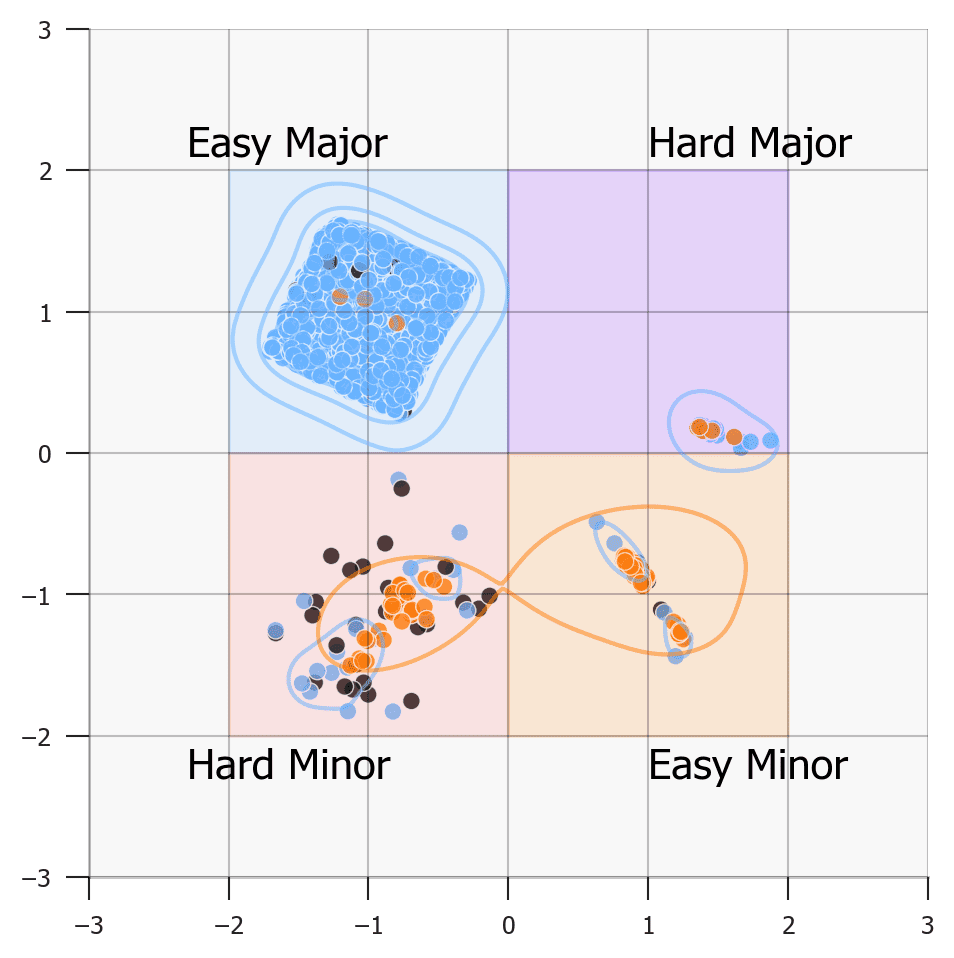}}
    \subfigure[DFBS]{\includegraphics[width=0.24\linewidth]{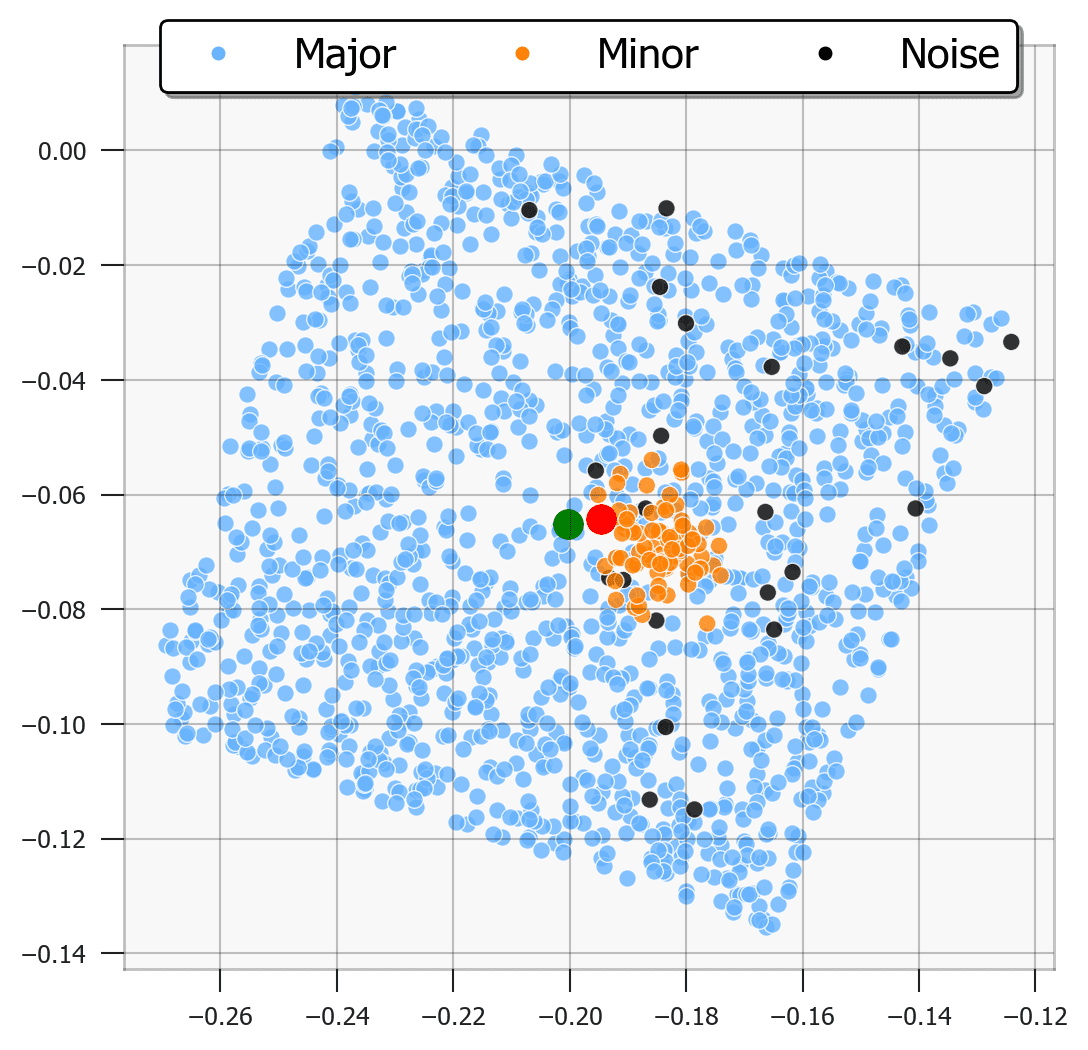}}
    \subfigure[DeepSMOTE]{\includegraphics[width=0.24\linewidth]{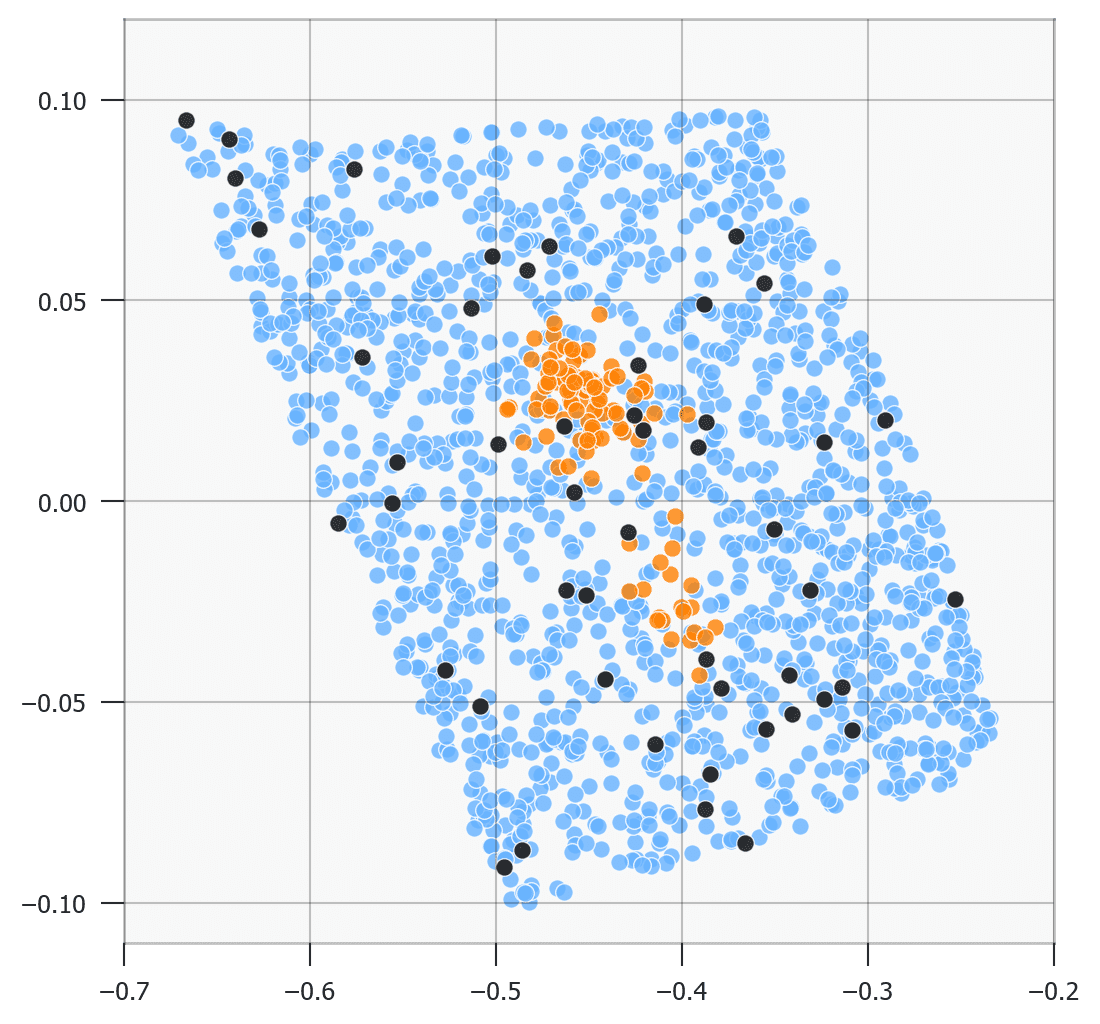}}
    \subfigure[DDHS]{\includegraphics[width=0.24\linewidth]{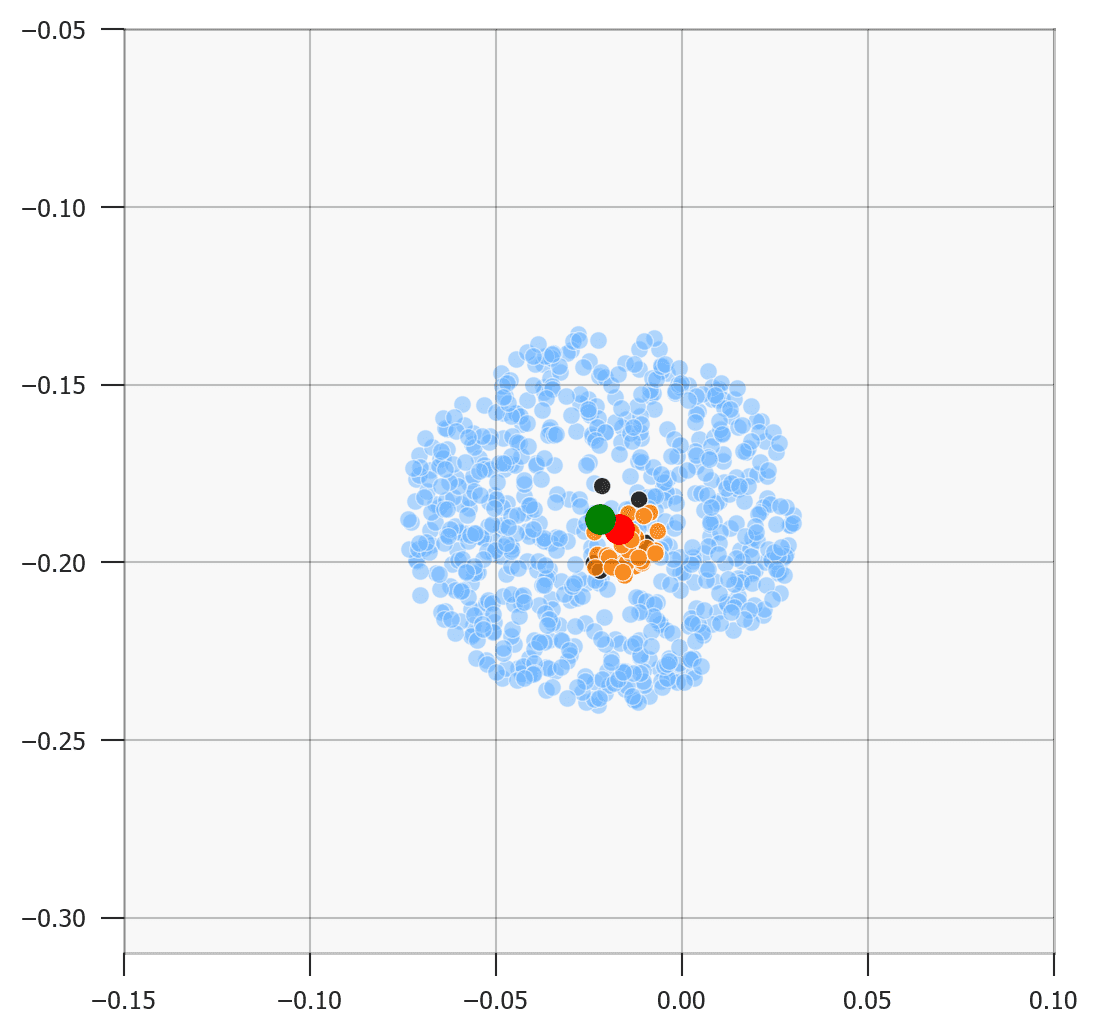}}
    \subfigure[SMOTE-CLS]{\includegraphics[width=0.24\linewidth]{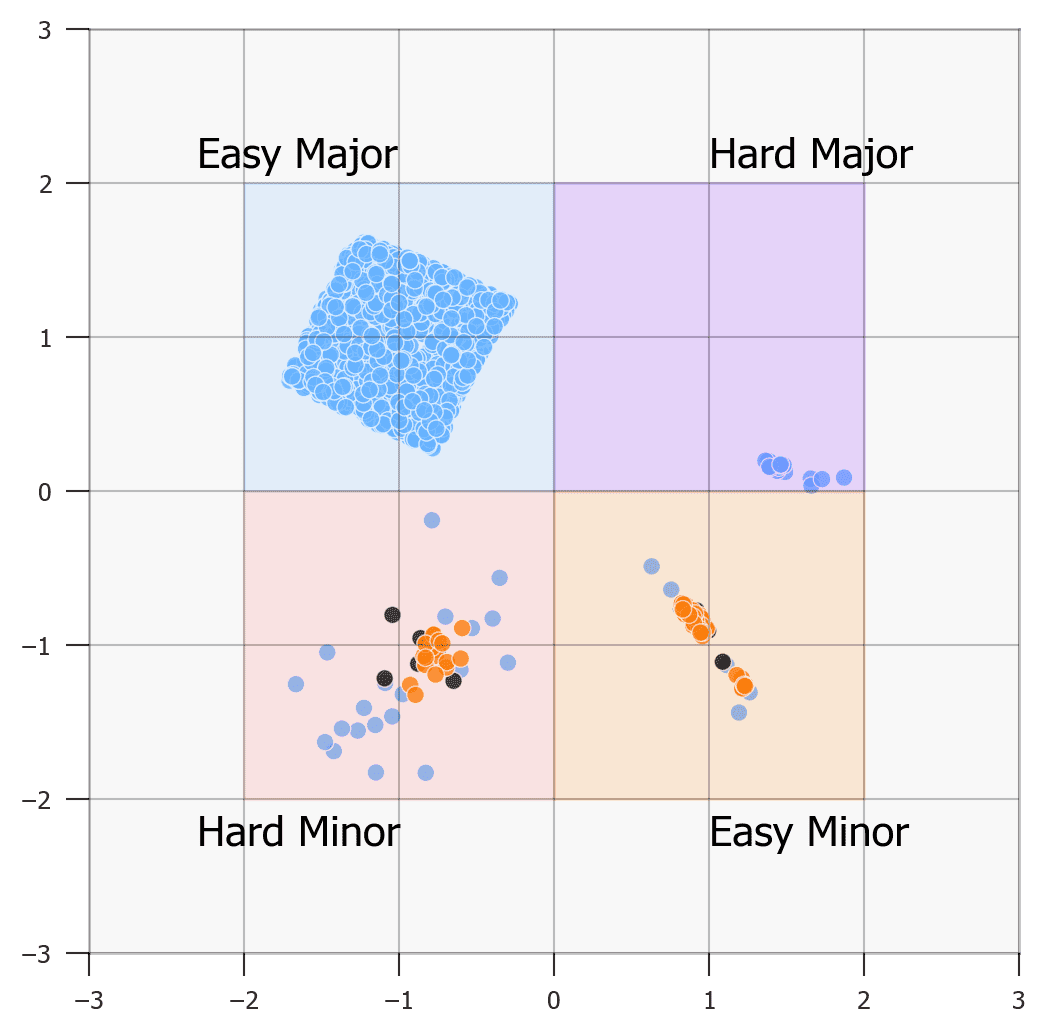}}
    \subfigure[DFBS]{\includegraphics[width=0.24\linewidth]{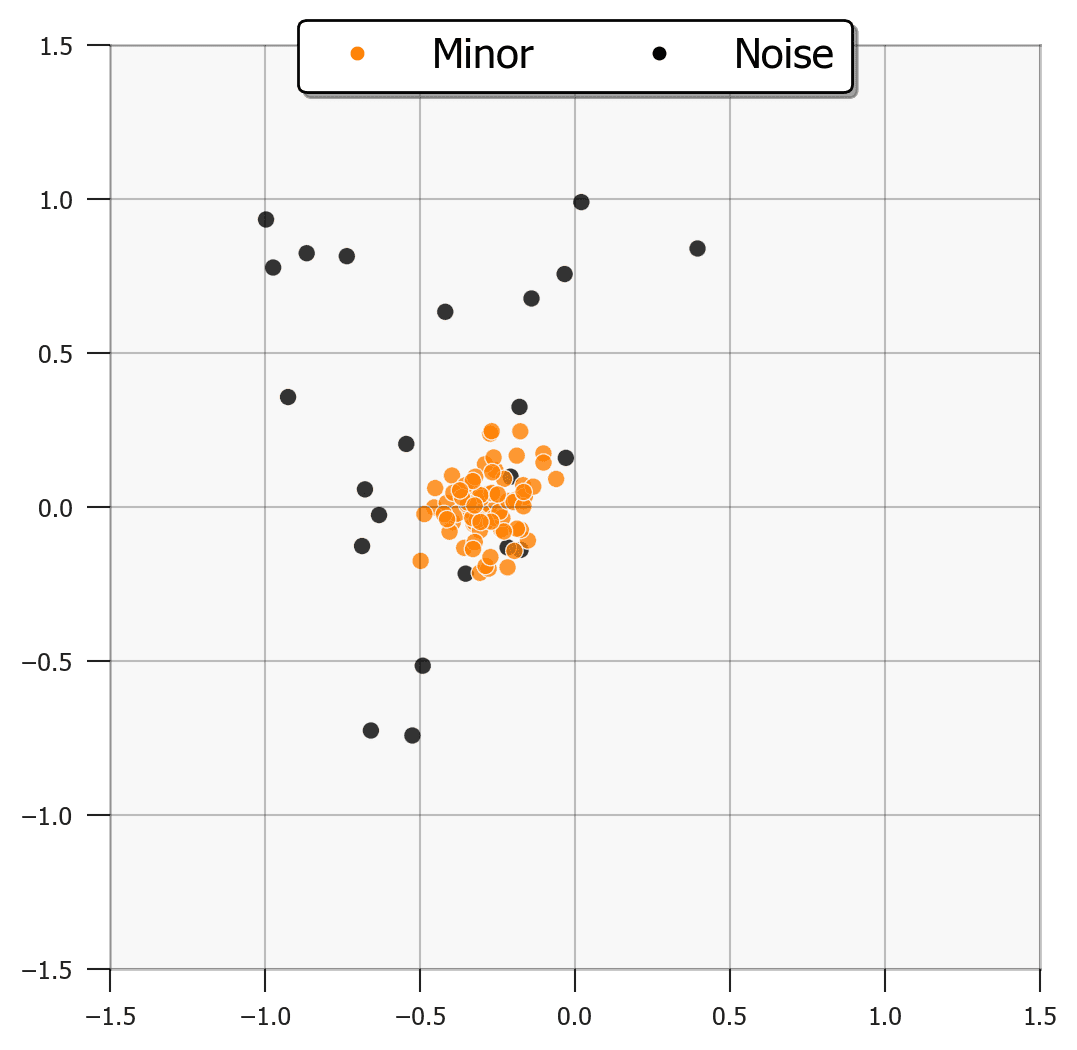}}
    \subfigure[DeepSMOTE]{\includegraphics[width=0.24\linewidth]{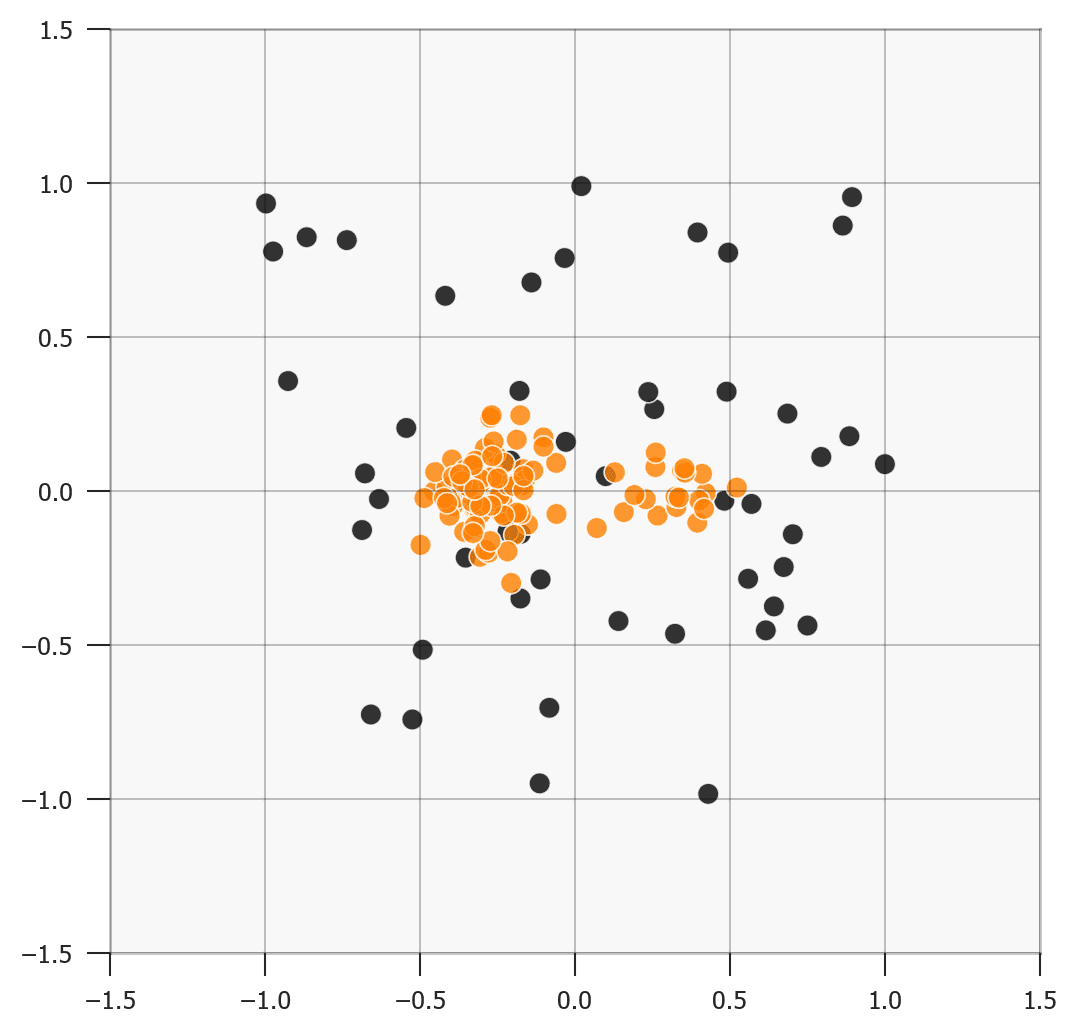}}
    \subfigure[DDHS]{\includegraphics[width=0.24\linewidth]{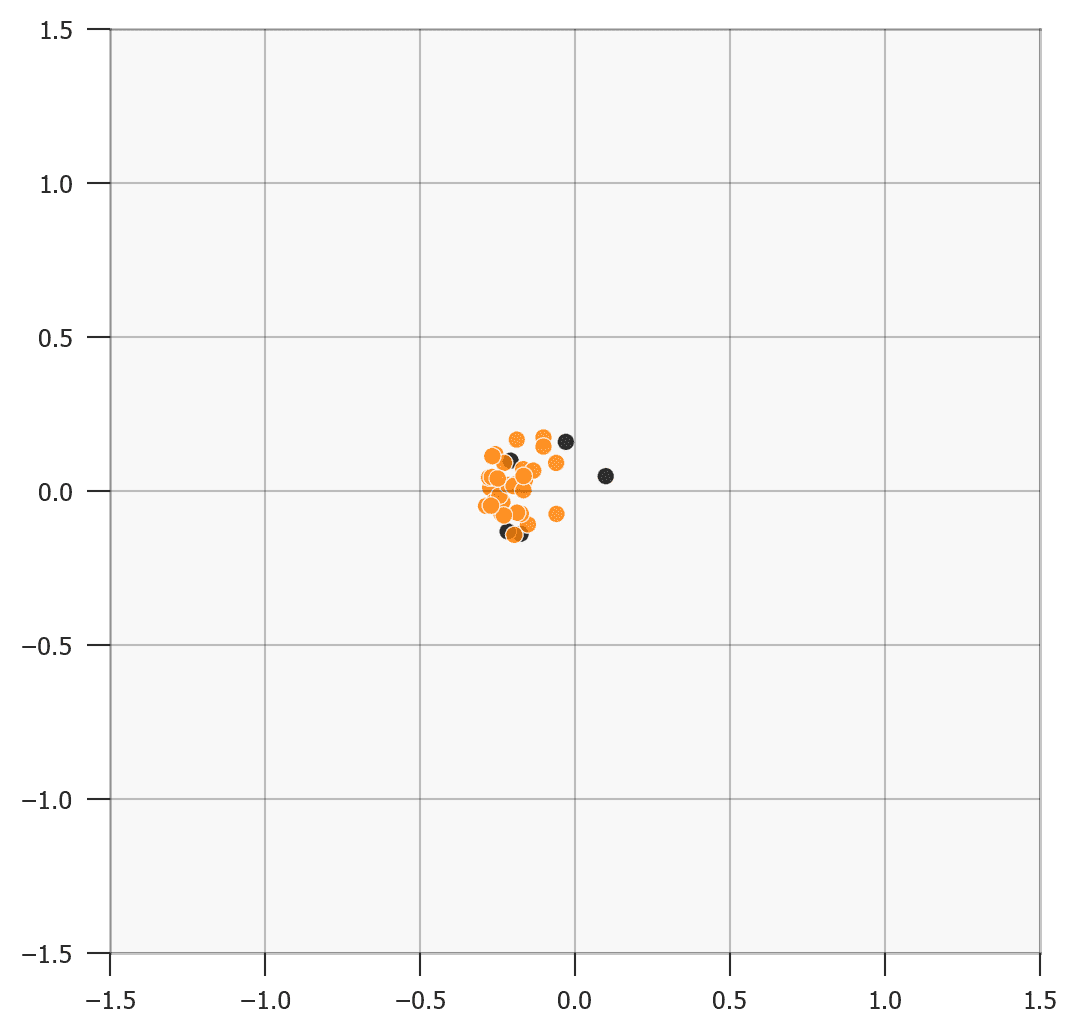}}
    \subfigure[SMOTE-CLS]{\includegraphics[width=0.24\linewidth]{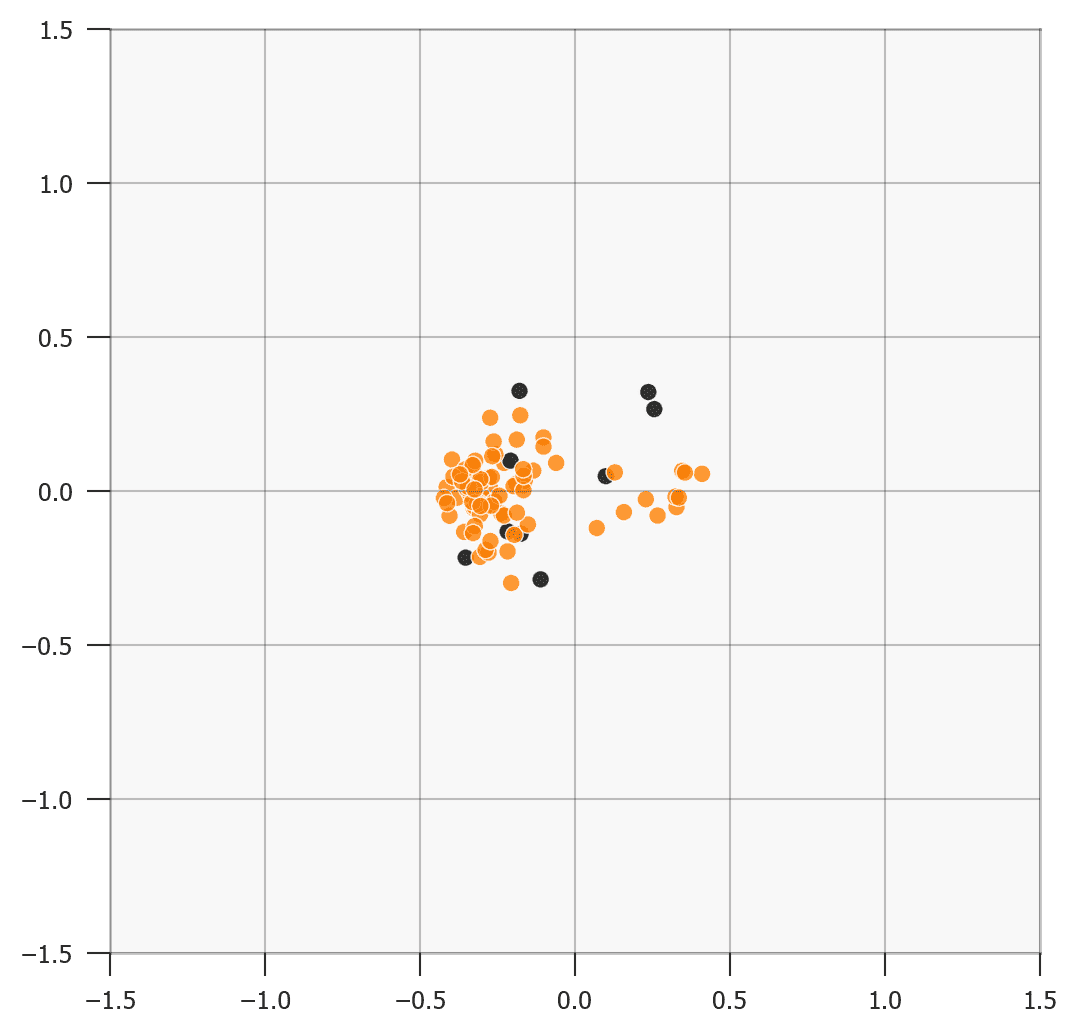}}
    \subfigure[DFBS]{\includegraphics[width=0.24\linewidth]{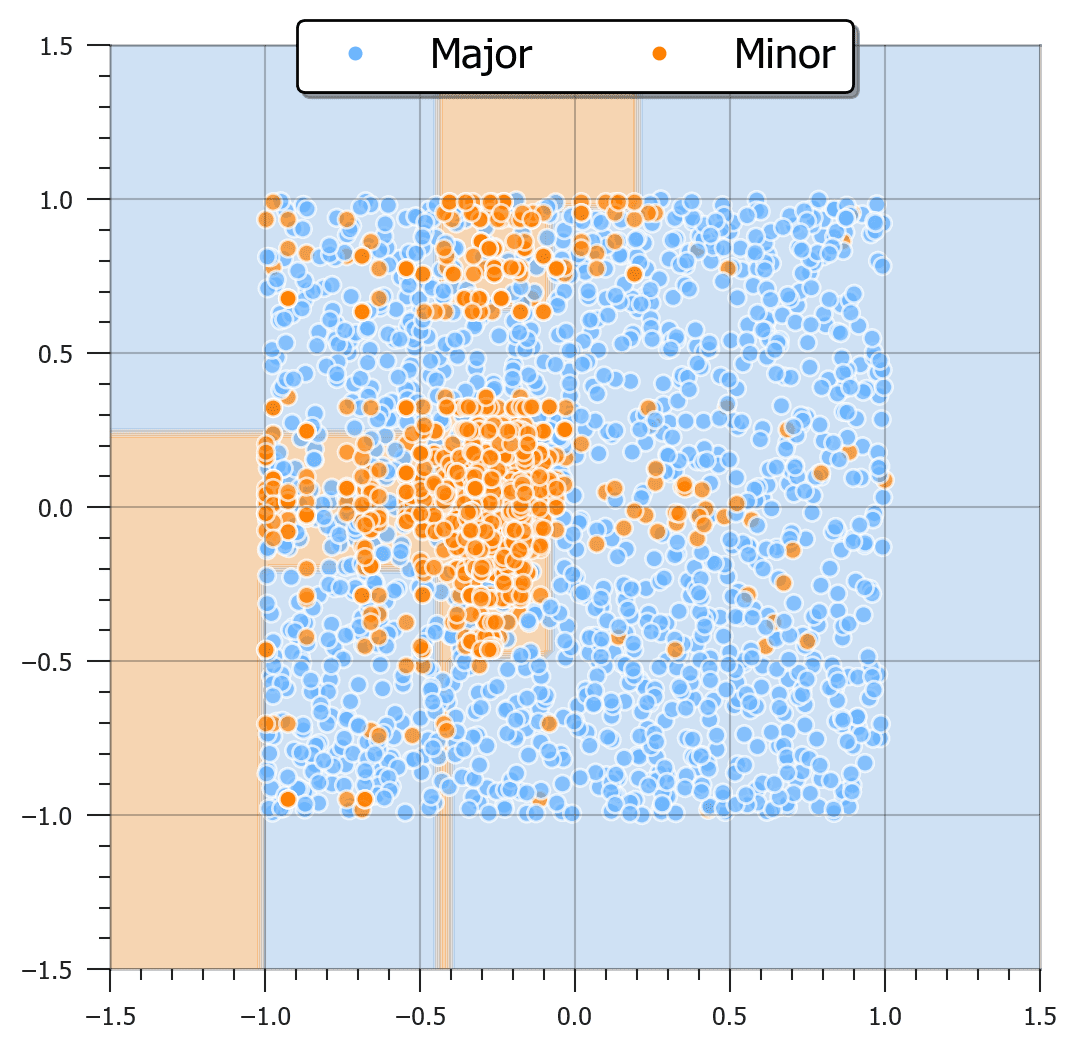}}
    \subfigure[DeepSMOTE]{\includegraphics[width=0.24\linewidth]{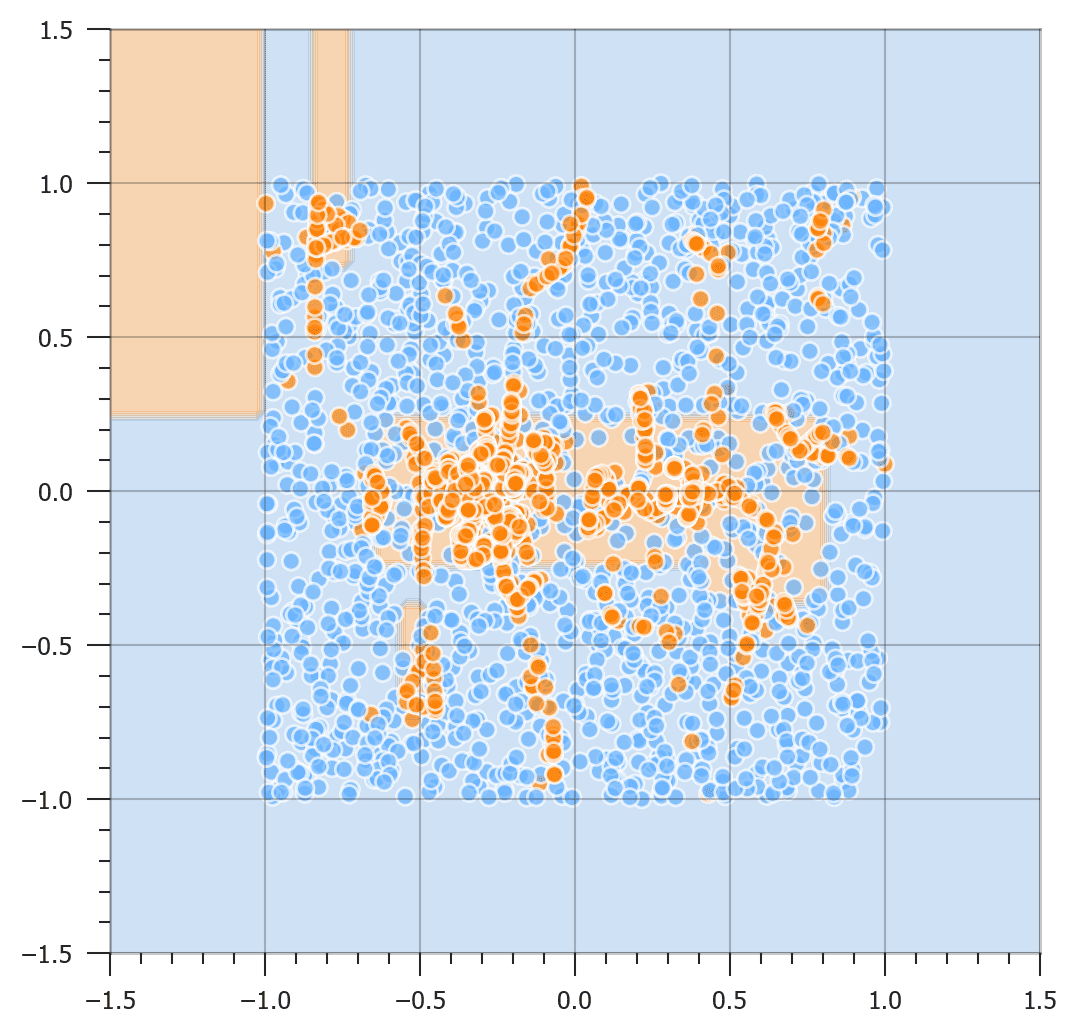}}
    \subfigure[DDHS]{\includegraphics[width=0.24\linewidth]{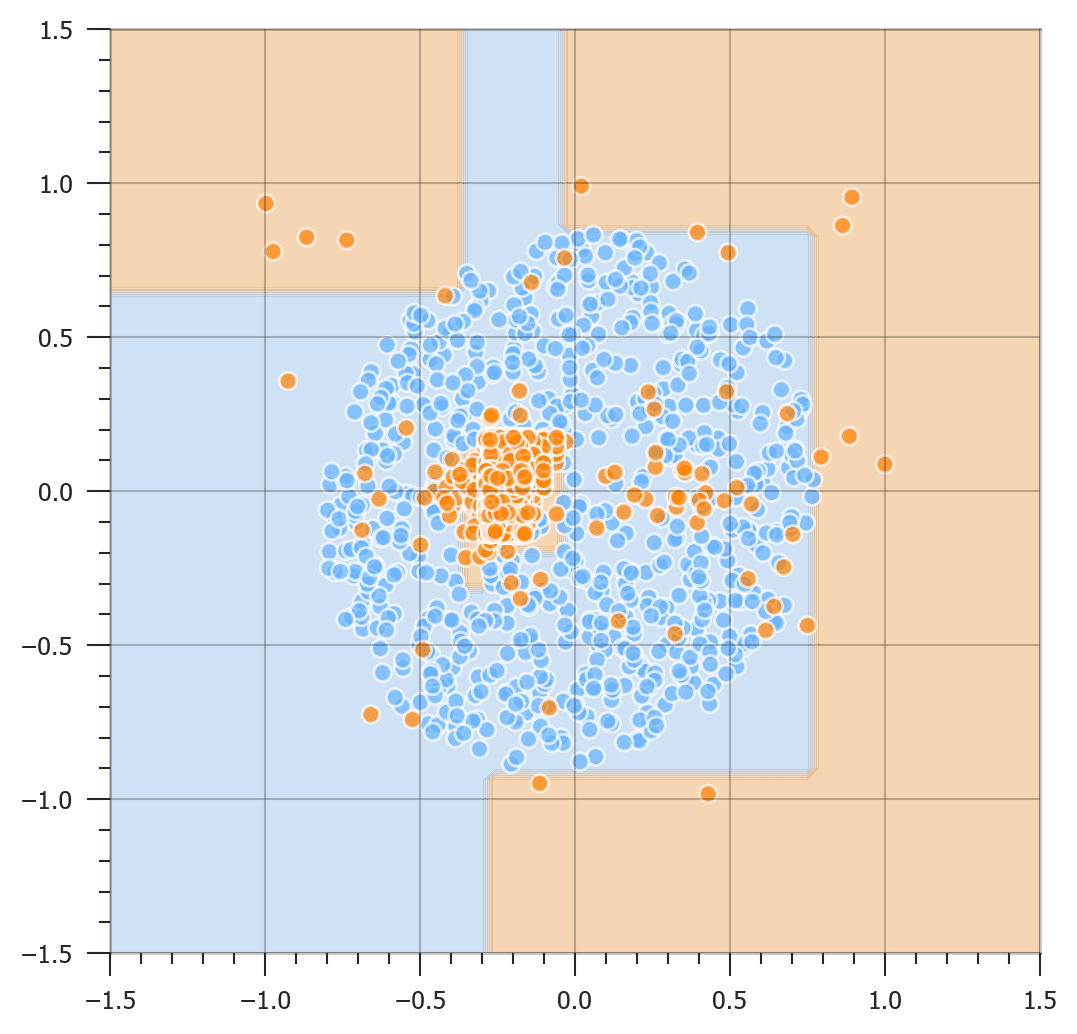}}
    \subfigure[SMOTE-CLS]{\includegraphics[width=0.24\linewidth]{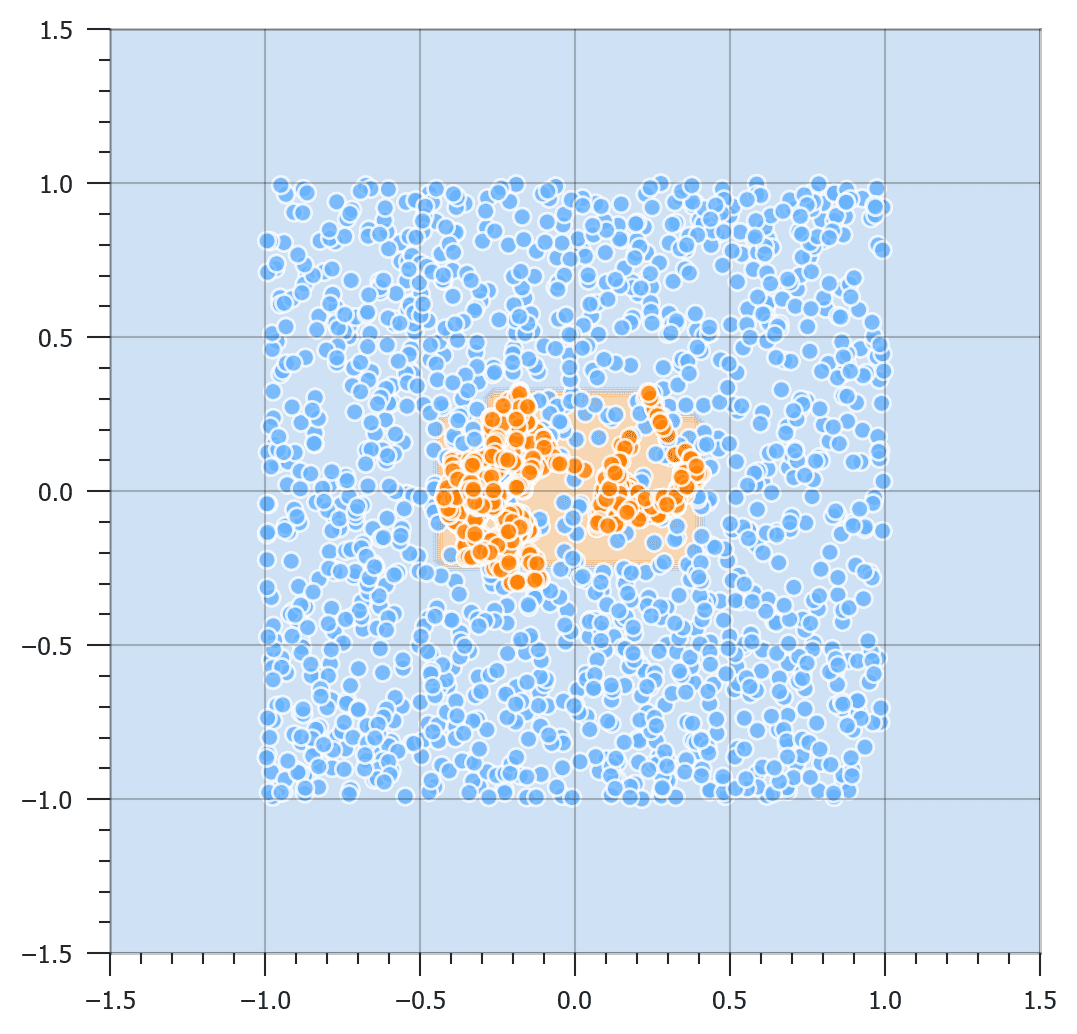}}
    \caption{Visualization of the oversampling process in the simulation study. (a) - (d): the trained latent variables. (e) - (h): the selected latent variables by each filtering method. (i) - (l): the selected minority samples for oversampling. (m) - (p): the oversampling results and decision boundary using decision trees. In the visual representation of DFBS and DDHS, green and red points denote the centers of major and minor sets, respectively. Note that the centers of each class do not exhibit a clear separation.}
    \label{fig:simul_results}
\end{figure}

Figure \ref{fig:simul_results} visually represents the oversampling process employed by each step. Specifically, the first row of Figures \ref{fig:simul_results} illustrate the latent space generated by four oversampling methods. Light blue points denote the latent features of the major class, and orange points indicate those of the minor class. Black points are noise variables labeled as the minor class. In Figure \ref{fig:simul_results} (a) and (c), the centroids of the major and the minor classes, which are used in their filtering process, are displayed by the blue and red points, respectively.
Unlike the other methods, SMOTE-CLS spatially aligns latent features into four distinct groups based on the predefined priors $p(\bz)$ of Figure \ref{fig: ex_prior}. In the third quadrant of \ref{fig: ex_prior} (d), we can see that the noise samples scatter around the minority samples. Therefore, it is visually easy to identify uncommon samples within each group in Figure \ref{fig:simul_results} (d). Most noise samples predominantly consist of hard-minority samples, yet they are noticeably distant from the conditional expectation, $\bmu_{m^*} = (-1, -1)$. 

The second row of Figure \ref{fig:simul_results}  visualizes the latent variables ($\bz$) after each filtering process, and the third row of Figure \ref{fig:simul_results} shows the filtered minor set ($\tilde{\bX}_m$). 
Note that $\bz$ denotes the latent variables without specifying a particular model. Only DeepSMOTE does not have its filtering process such that Figure \ref{fig:simul_results} (b) is equal to Figure \ref{fig:simul_results} (f). First, DFBS defines the candidate set for oversampling, which satisfies the following condition:
\bea \label{eq:dfbs_filtering}
\tilde{\bX}_m = \{\bx_i ~|~ \|\bz_i - \bar{\bz}_M \| > \|\bz_i - \bar{\bz}_m \|, y_i = m\}, 
\eea
where $\bar{\bz}_M$ and $\bar{\bz}_m$ denote the centers of latent variables of major and minor sets, respectively, presented as green and red points in Figure \ref{fig:simul_results} (a). \\
The center of each class computed with Euclidean distance is biased to outliers or large-sized clusters. Therefore, as shown in Figures \ref{fig:simul_results} (e) and (i), the selected latent variables and minority samples predominantly come from the large-sized cluster, and even noise samples near the large-sized cluster are chosen. This outcome is because the center of the minor class is slightly biased toward the large-sized cluster. Consequently, the latent variables from the small-sized cluster are excluded from the candidate set for oversampling. 

Figures \ref{fig:simul_results} (f) and (j) pertain to DeepSMOTE, which does not involve a filtering process; it selects all minority samples, including noise ones, as candidates. Because of the unconditional oversampling, its decision boundary in Figure \ref{fig:simul_results} (n) is not a desired result.  

Especially, DDHS selects the minor subset for oversampling and the major subset for training the classifier based on the probability densities. In this paper, we only focus on the selection of a minor subset, and its filtering algorithm for only the minor set can be written as follows: 
\bea \label{eq:ddhs_filtering}
\tilde{\bX}_m = \{\bx_i ~ | ~ \hat{f}_{m,m*}(\bz_i) > \tau_3, y_i = m \}, 
\eea
where $\hat{f}_{m,m*}$ is an estimated density of minor latent variables by KDE, and set $\tau_3$ to select 75\% minor samples. In a similar way, DDHS selects the 50\% major samples by $\hat{f}_{M,M*}$

In Figure \ref{fig:simul_results} (g) and (k), corresponding to the results of DDHS, only minority samples with higher densities in the latent space are selected. As a result, samples from the small and large clusters are excluded from the candidate set for oversampling. Additionally, DDHS keeps the minority samples not selected for oversampling to train the classifier while excluding the major ones not selected. As a result, in Figure \ref{fig:simul_results} (o), noise samples dominate the decision boundary region where the major samples have been removed.

In contrast, as shown in Figures \ref{fig:simul_results} (h) and (l), SMOTE-CLS effectively filters out noise samples from the minor class while also selecting candidates from both the large and small-sized clusters. Precisely, in our approach, we consider both similarity and difficulty when embedding samples into the latent space. This results in more balanced and effective oversampling, as the selected latent variables include representatives from both clusters rather than being dominated by samples from the larger cluster. This approach allows us to generate synthetic samples that better capture the underlying distribution of the minor class, ultimately improving classification performance.

Finally, the last row of Figures \ref{fig:simul_results}  represents the oversampling outcomes with augmented minor set $\hat{\bX}_m$ on the data space. DFBS and DeepSMOTE appear susceptible to noise samples, while DDHS fails to augment samples from the $G_2$ because the filtering process excludes the small disjunct population. In contrast, only SMOTE-CLS exhibits a decision boundary that closely aligns with the true generative models $G_1$ and $G_2$. Through this simulation study, SMOTE-CLS has robustness to noise and outperforms other oversampling methods in within-class imbalance classification.

\subsection{Ablation Studies} \label{sec:ablation}
\begin{table}[t]
\caption{Summary of the models in the ablation study. 
}
  \centering
  \begin{tabular}{lccc}
    \toprule
    Model & Disentanglement & Difficulty segmentation  & Adaptive filtering \\
    \midrule
    w/o dis  & \color{red}{\ding{55}} & \color{red}{\ding{55}} & \color{red}{\ding{55}}  \\
    w/o seg & \color{blue}{\ding{51}} & \color{red}{\ding{55}}  & \color{red}{\ding{55}} \\ 
    w/o af  & \color{blue}{\ding{51}} &  \color{blue}{\ding{51}} & \color{red}{\ding{55}} \\ 
    SMOTE-CLS & \color{blue}{\ding{51}} & \color{blue}{\ding{51}} & \color{blue}{\ding{51}} \\
    \bottomrule
  \end{tabular}
\label{tab:ablation}
\end{table}

We conduct ablation studies to assess the significance of each step in Algorithm 1. With the synthetic data used in Section \ref{sec:simul} we demonstrate its ability to filter out noise and select high-quality candidates for oversampling effectively. Our study focuses on three key aspects of the proposed procedure: disentangling latent space (the use of GMM), sample difficulty segmentation (introduction of refined levels in (1)), and group adaptive filtering (Section \ref{sec:os}). Each step is configured, and the generated minority samples in the configured SMOTE-CLS are compared.

Table \ref{tab:ablation} shows the configuration in the ablation study. The `w/o dis' model denotes the VAE without disentangling the latent space, essentially equivalent to the conditional VAE (CVAE). The `w/o seg' model represents the VAE that disentangles the major and minor groups without incorporating the segmentation according to classification difficulty. In this case, we employ the two-component Gaussian mixture distribution as the prior and set the location parameters of each class as $(1, 1)$ for the major class and $(-1, -1)$ for the minor class.

The `w/o af' model denotes the SMOTE-CLS model, excluding the group adaptive filtering. It's worth noting that the other models, excluding SMOTE-CLS, involve the naive density-based filtering algorithm, which employs KDE on the entire minor set and selects a minority sample whose density is higher than a predefined threshold $\tau$. In this study, we set $\tau$ where 40\% of the minority samples are filtered out.

\begin{figure}[ht]
    \centering
    \subfigure[w/o dis]{\includegraphics[width=0.24\linewidth]{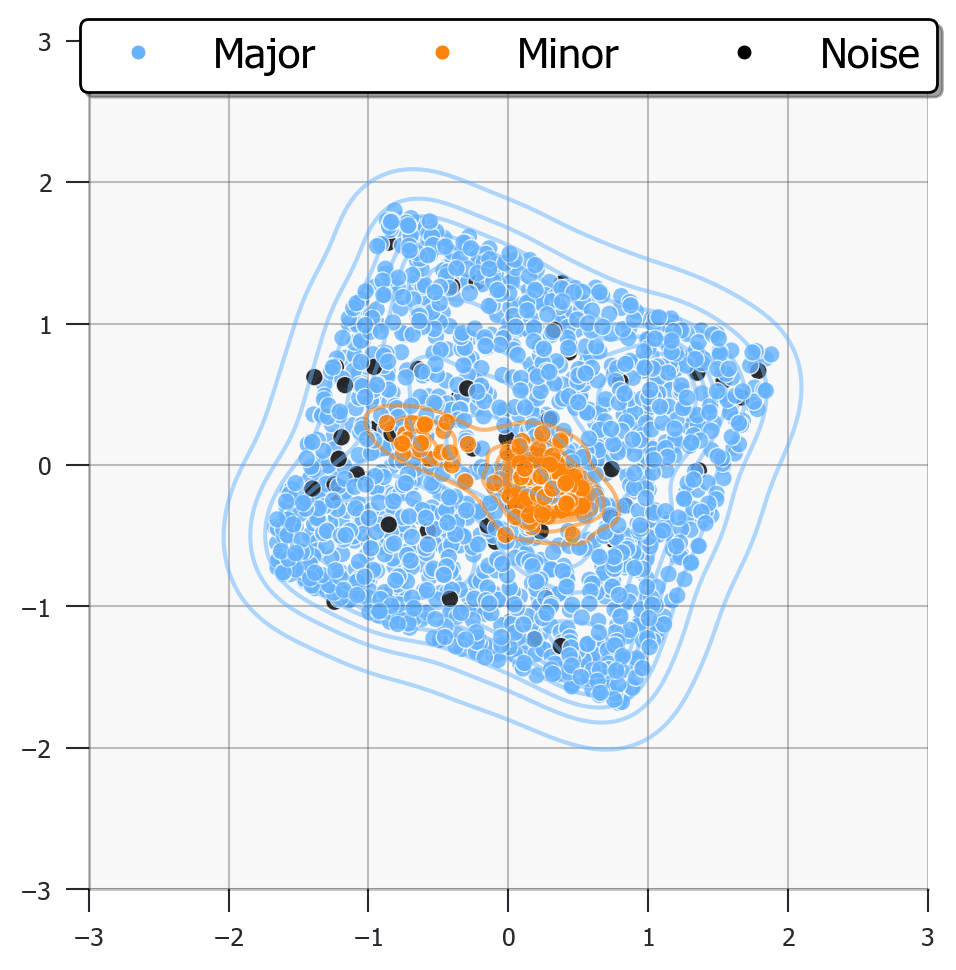}}
    \subfigure[w/o seg]{\includegraphics[width=0.24\linewidth]{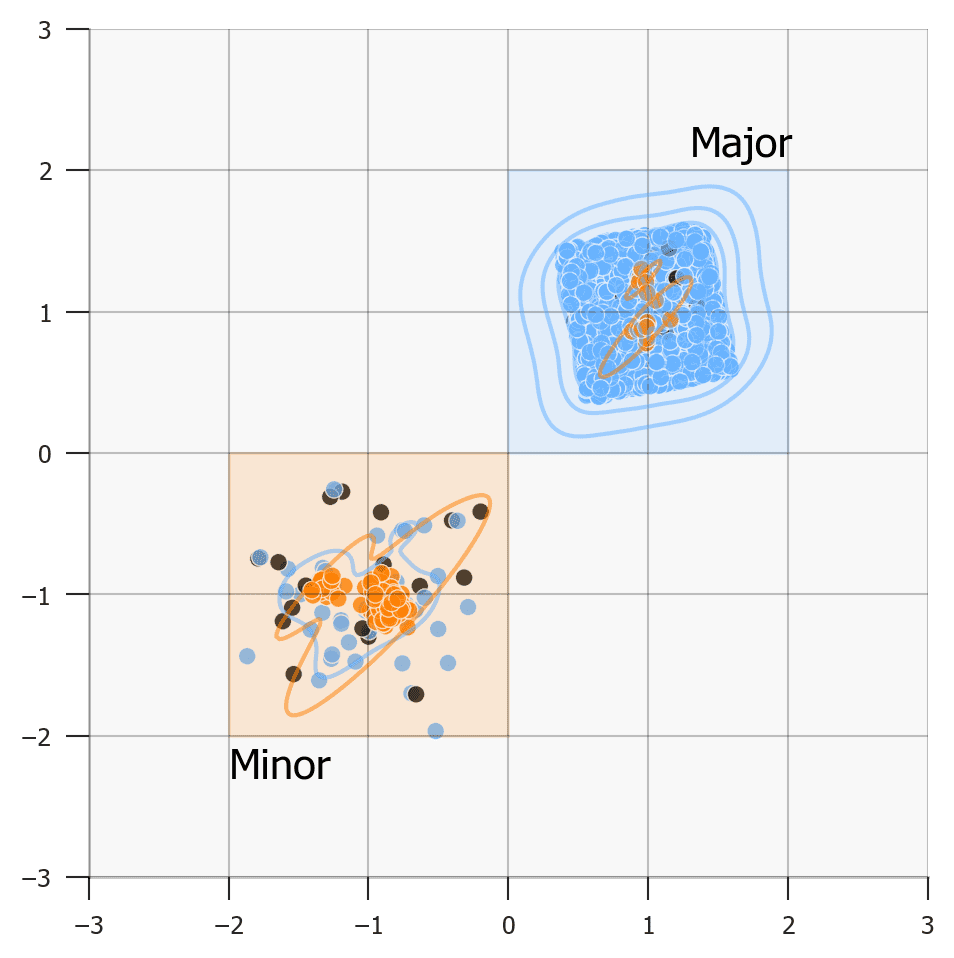}}
    \subfigure[w/o af]{\includegraphics[width=0.24\linewidth]{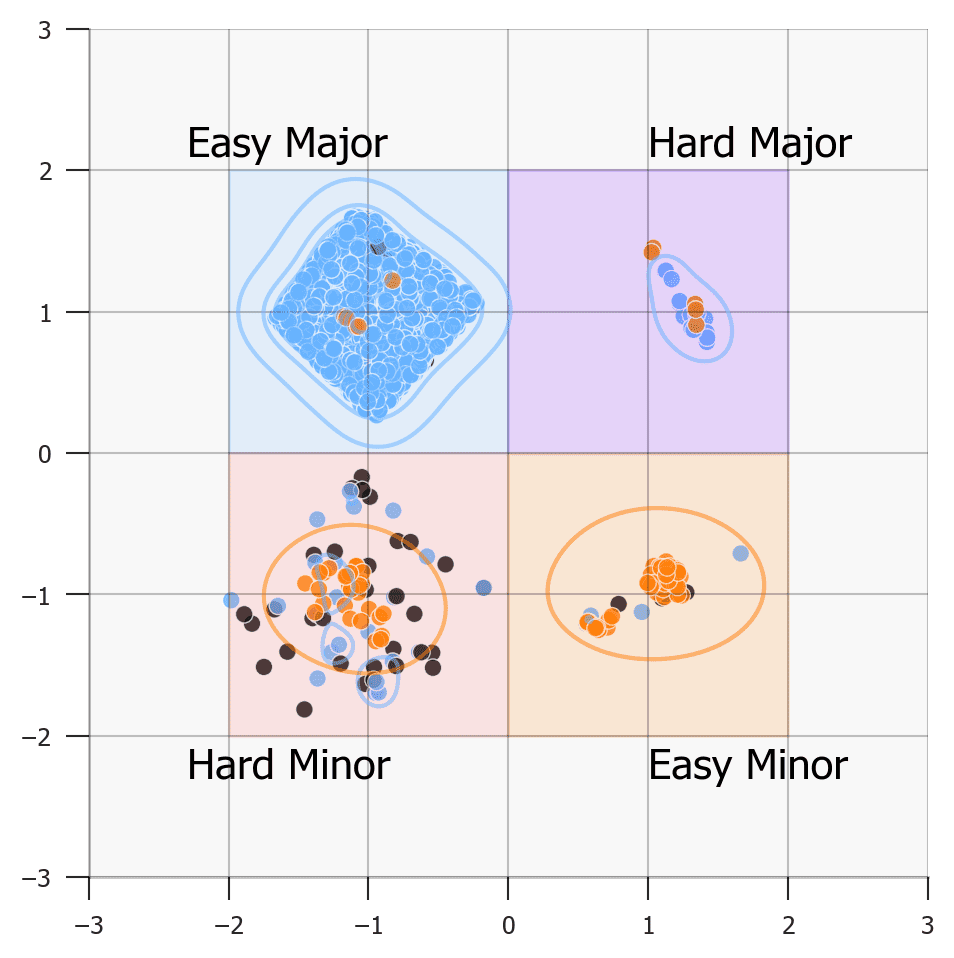}}
    \subfigure[SMOTE-CLS]{\includegraphics[width=0.24\linewidth]{fig/simul/simul_exon.png}}
    \subfigure[w/o dis]{\includegraphics[width=0.24\linewidth]{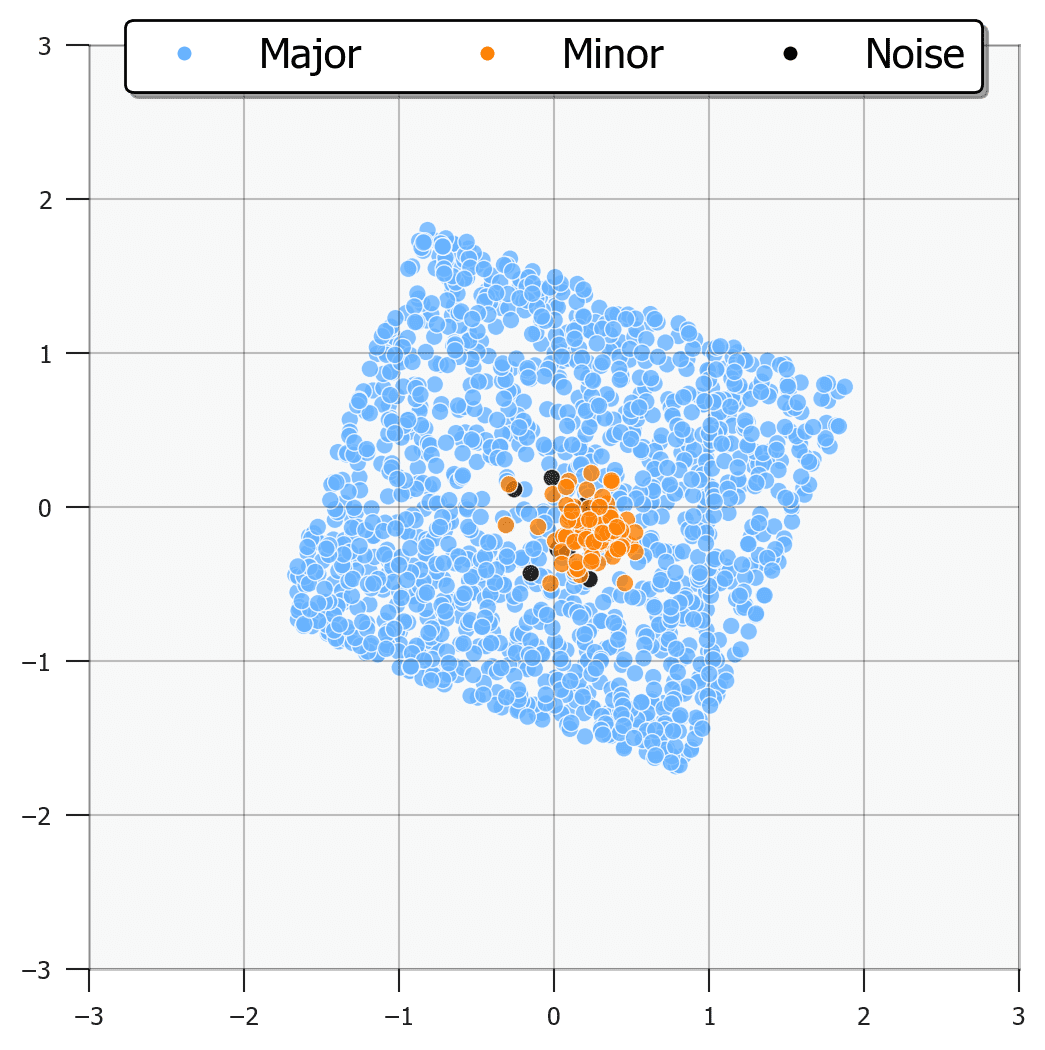}}
    \subfigure[w/o seg]{\includegraphics[width=0.24\linewidth]{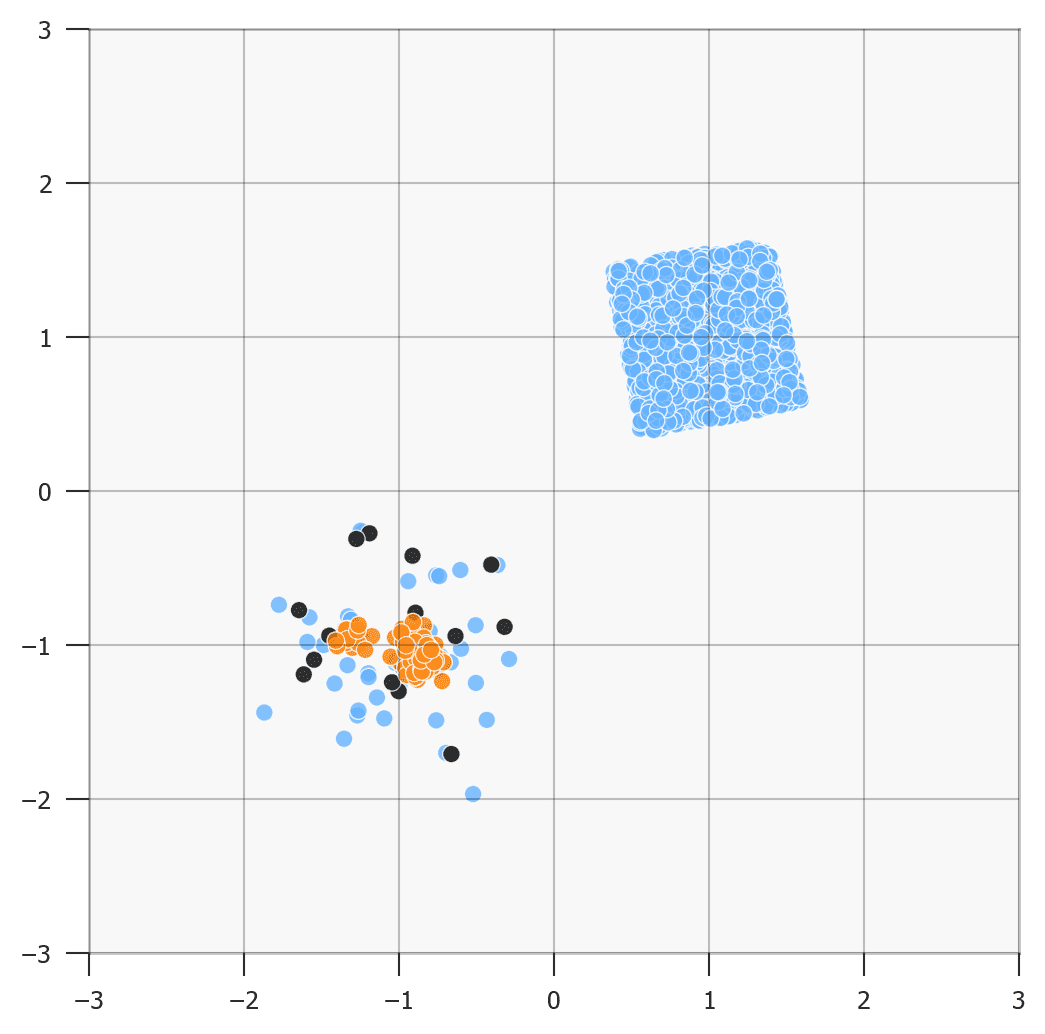}}
    \subfigure[w/o af]{\includegraphics[width=0.24\linewidth]{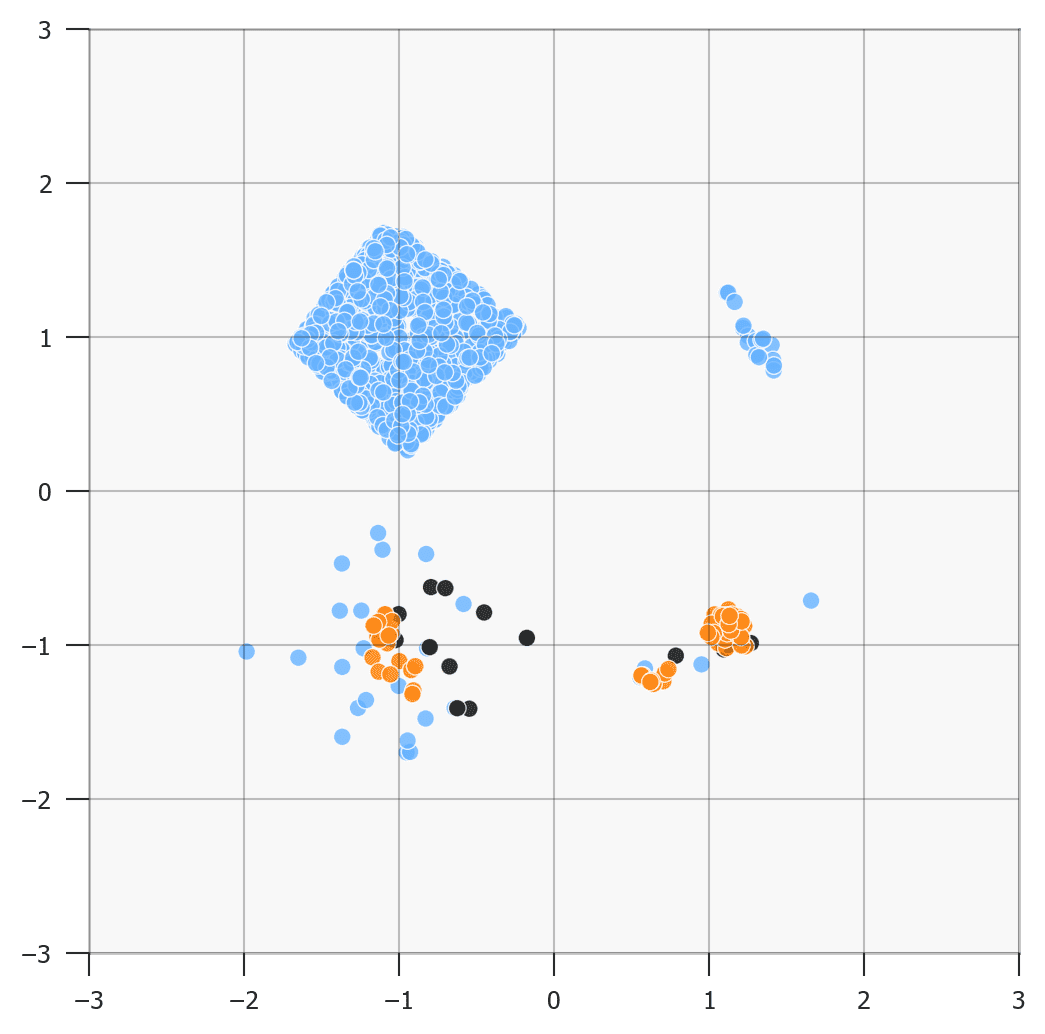}}
    \subfigure[SMOTE-CLS]{\includegraphics[width=0.24\linewidth]{fig/simul/simul_exon_selected_latent.png}}
    \subfigure[w/o dis]{\includegraphics[width=0.24\linewidth]{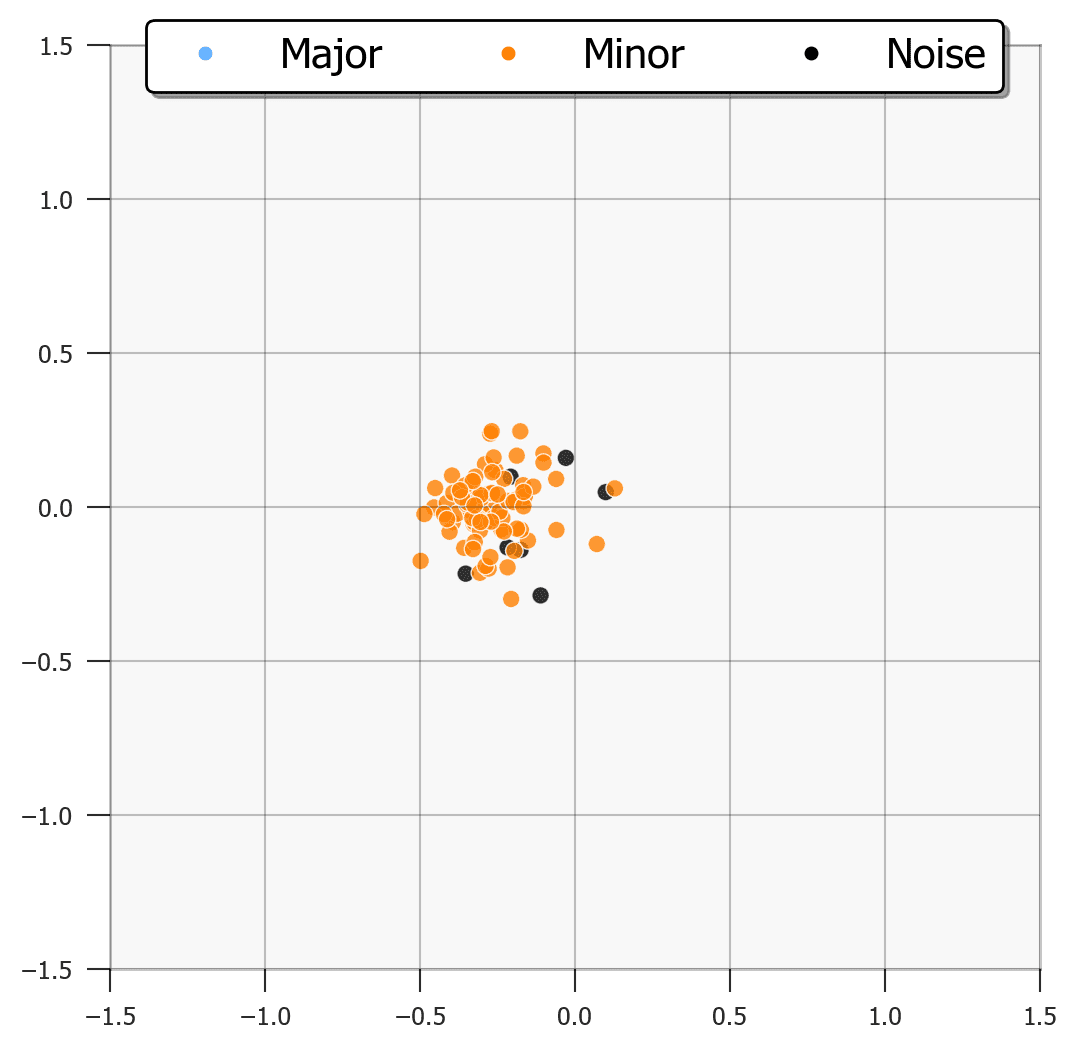}}
    \subfigure[w/o seg]{\includegraphics[width=0.24\linewidth]{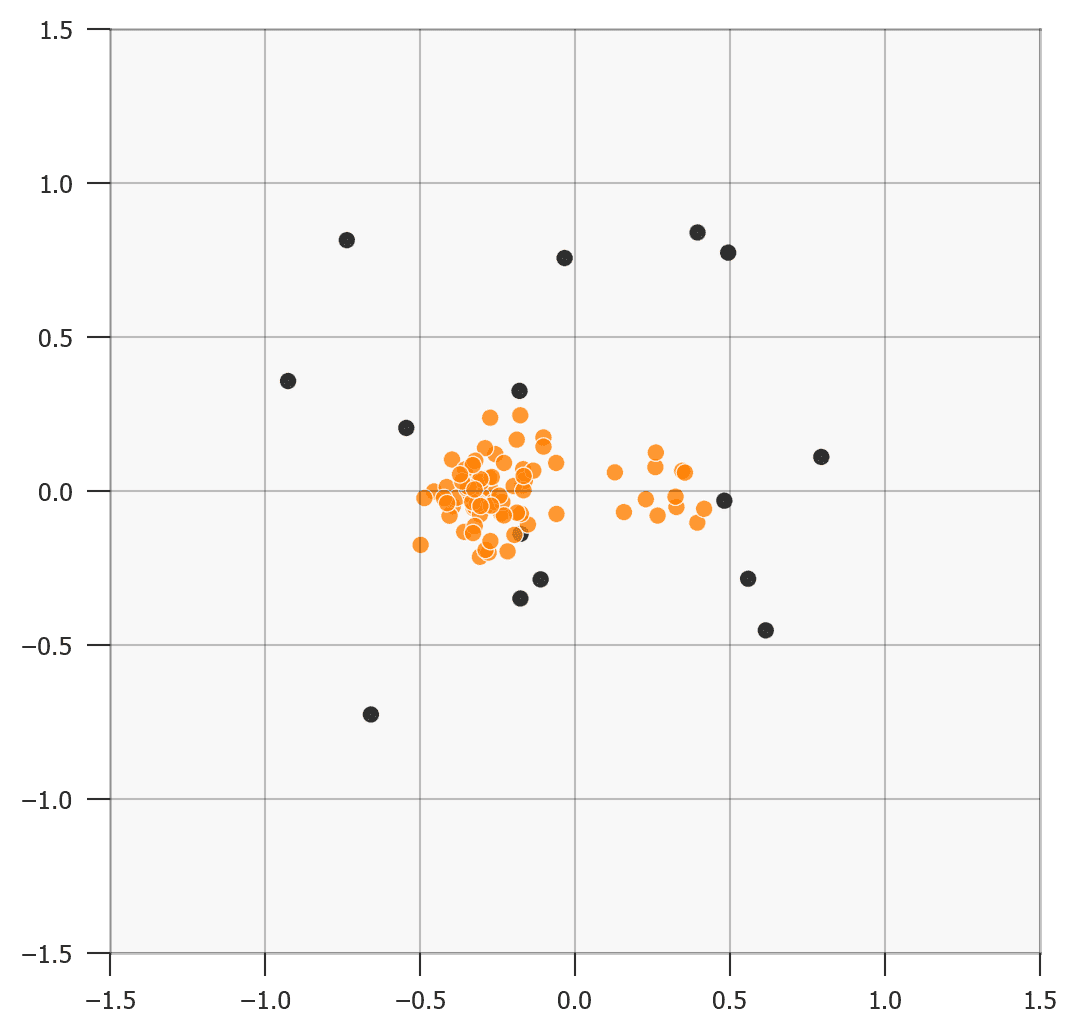}}
    \subfigure[w/o af]{\includegraphics[width=0.24\linewidth]{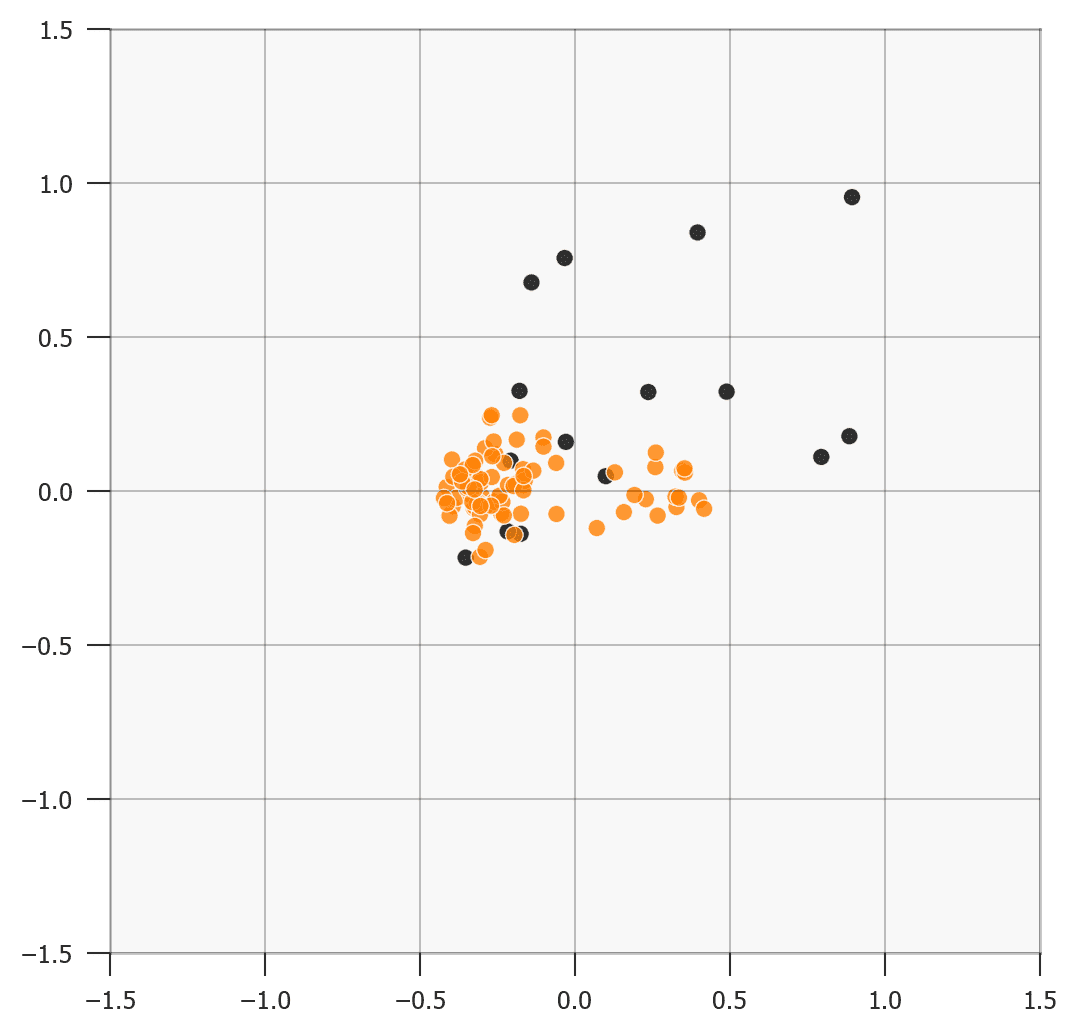}}
    \subfigure[SMOTE-CLS]{\includegraphics[width=0.24\linewidth]{fig/simul/simul_exon_selected_x.png}}
    \caption{Visualization of the ablation study. (a) - (d): the trained latent variables. (e) - (h): the selected latent variables by each filtering method. (i) - (l): the selected minority samples for oversampling.}
    \label{fig:ablation}
\end{figure}

Each column of Figure \ref{fig:ablation} corresponds to the configuration in the SMOTE-CLS for the ablation study. The last column shows the result of the SMOTE-CLS. Each row from top to bottom illustrates the latent variables, the filtered latent variables, and the filtered data points of the minor class. Thus, the bottom row displays what minority samples are selected as high-quality samples under the small disjunct problem existing.

In the first column in Figure \ref{fig:ablation}, the latent space of `w/o dis' is similar to the other AE-based models, such as DFBS, DDHS, and DeepSMOTE,  in Figure \ref{fig:simul_results} that exploits only class labels without additional information or disentangling latent space. As a result, the noise samples, mislabelled data following the major class distribution, are embedded around the minor class, and the samples from the small disjunct are ignored by `w/o dis'. See Figure \ref{fig:ablation} (i). 

As seen in Figure \ref{fig:ablation} (b), when the sample difficulty segmentation is not applied, a significant amount of minority samples are embedded around $(1,1)$, the center of the prior of the major class.  Conversely, noise samples are embedded around $(1,1)$, the center of the prior of the minor class. It leads to undesirable outcomes in the density-based filtering process because these minority samples are excluded. In contrast, many noise samples are selected, as shown in Figures \ref{fig:ablation} (f) and (j). 

In the case of `w/o af' denoting SMOTE-CLS without the group adaptive filtering algorithm, the latent spaces of hard and easy samples are divided by the predefined prior distribution in Figure \ref{fig:ablation} (c). In contrast to Figure \ref{fig:ablation} (b), the minority samples are embedded within the latent space of both hard and easy-minor sets, forming a bimodal distribution. However, the filtering with the KDE cannot exclude the noise samples from the minor class because the noise samples are embedded with the minority samples. The latent variables of noise samples are close to the distribution of the easy-minor class. This happens because these noise samples have higher densities than the minority samples that are farther from the prior distribution of the easy-minor class. See Figures \ref{fig:ablation} (g) and (k). 

These results highlight the need for our proposed method to effectively address challenging issues, such as noise filtering and the selection of small-sized minor sets. Also, contrary to `w/o af', our proposed model is robust to the prior distribution (see Appendix \ref{app:prior}).  Note that SMOTE-CLS does not provide only a statistical detection method for a noise sample. Instead, it renders a general method of specifying a local neighborhood on data space $\cX$ via the disentanglement of latent space $\cZ$. SMOTE-CLS constructs clusters on latent variables via the prior distribution, the predesigned GMM aiming to a specific disentanglement. By relabeling observations, we can separate each mode of $q(\bz|\tilde y = c)$ for $c = M^*, m^*, M, m$ and define a cluster within each subclass. The samples in each cluster convey information about a new neighborhood used in SMOTE. Furthermore, because the criterion of the disentanglement can be variously defined according to the object of data analysis, SMOTE-CLE provides a flexible way to define a neighborhood for improving SMOTE.

\subsection{Oversampling Performance in Real Data Analysis} \label{sec: ra}
\subsubsection{Datasets}
We conduct a comparative study with 12 benchmark imbalanced datasets. These datasets are publicly available in the UCI Machine Learning Repository \citep{uci} and the LIBSVM dataset repository \citep{chang2011libsvm}. Table \ref{tab:dataset} provides detailed descriptions of the benchmark datasets, including the number of samples (ranging from 336 to 7797), the number of features (ranging from 7 to 617), and the imbalanced ratio (IR). IR is calculated as the ratio of the major class size to the minor class size and varies from 3.6\% to 11.6\%.

\begin{table}[t]
\caption{Summary of the datasets.}
  \centering
  \begin{tabular}{llccc}
    \toprule
    \textsf{Dataset} & \textsf{Repository} & \textsf{\# samples} & \textsf{\# features} & \textsf{IR} \\
    \midrule
    \texttt{ecoli} & UCI & 336 & 7 & 11.6\% (301:35)\\
    \texttt{libras\_move} & UCI & 360 & 90 & 7.1\%  (336:24)\\
    \texttt{spectrometer} & UCI & 531 & 93 & 9.3\%  (486:45)\\
    \texttt{oil} & UCI & 937 & 49 & 4.6\% (896:41)\\
    \texttt{yeast} & UCI & 1,484 & 8 & 3.6\% (1433:51) \\
    \texttt{car\_eval} & UCI & 1,728 & 21 &  8.4\% (1594:134)\\
    \texttt{us\_crime} & UCI & 1,994 & 100 & 8.1\% (1844:150)\\
    \texttt{scene} & LIBSVM & 2,407 & 294 & 7.9\% (2230:177)\\ 
    \texttt{abalone} & UCI & 4,177 & 10 & 10.3\% (3786:391) \\ 
    \texttt{optical\_digits} & UCI & 5,620 & 64 & 10.9\% (5066:554)\\    
    \texttt{satimage} & UCI & 6,435 & 36 & 10.8\% (5809:626)\\
    \texttt{isolet} & UCI & 7,797 & 617 & 8.3\% (7197:600)\\  
    \bottomrule
  \end{tabular}
\label{tab:dataset}
\end{table}

\subsubsection{Comparison Models}
Our experiments compare SMOTE-CLS with 9 benchmarks, including SMOTE-based and deep learning-based approaches. Along with the comparative models used in Section \ref{sec:ablation}, we consider five additional models: Baseline (BASE) without oversampling, SMOTE, Borderline-SMOTE (BSMOTE), SMOTE-ENN, KMSMOTE, and CVAE. The SMOTE-based methods are implemented using the \textsf{imbalanced-learn} package \citep{lemaitre2017imbalanced}. The summary of the models is described in Table \ref{tab:models}.

\subsubsection{Evaluation Metrics} \label{sec:metric}
This study uses the Area Under the Precision-Recall Curve (AUPRC) and the Area Under the Curve (AUC) as our evaluation metrics. The AUPRC provides a comprehensive metric by considering Precision-Recall pairs across various thresholds, which is particularly valuable in the context of highly imbalanced classes. The definitions of precision and recall are as follows: 
\bean
\mbox{Precision} &=&  \frac{TP}{TP + FP}\\
\mbox{Recall} &=& \frac{TP}{TP + FN},
\eean
where $TP$, $FP$, and $FN$ denote the number of positive class samples correctly predicted by the classifier as a positive class, the number of negative class samples incorrectly predicted by the classifier as a positive class, and the number of positive class samples incorrectly predicted by the classifier as a negative class, respectively. In this study, we let the minor class be the positive. We also report the AUC, a general metric for assessing classification performance, taking into account all thresholds on the receiver operating characteristic curve. Both metrics are not dependent on a classification threshold value, thus being capable of measuring overall performance. Especially in imbalanced datasets, AUPRC is more informative than AUC \citep{davis2006relationship, hancock2023evaluating}.  

\begin{table}[t]
\caption{Summary of the oversampling techniques.}
  \centering
  \begin{tabular}{lll}
    \toprule
    \textsf{Model} & \textsf{Reference} & \textsf{Methods} \\
    \midrule
    BASE    & - & without oversampling  \\
    \midrule
    SMOTE     & \cite{chawla2002smote}          & SMOTE \\
    BSMOTE    & \cite{Han2005BorderlineSMOTEAN} & SMOTE + filtering \\
    SMOTE-ENN  & \cite{batista2004study}         & SMOTE + filtering \\
    KMSMOTE   & \cite{douzas2018improving}      & SMOTE + cluster-based filtering \\ 
    \midrule
    \multirow{2}{*}{DFBS} & \multirow{2}{*}{\cite{Liu2018DeepDF}}& Autoencoder + distance-based filtering + \\
    & & feature random combination \\
    DeepSMOTE & \cite{Dablain2021DeepSMOTEFD} & Autoencoder + SMOTE on latent variables \\ 
    \multirow{2}{*}{DDHS} & \multirow{2}{*}{\cite{Liu2022LearningFI}} & Autoencoder + KDE-based filtering + \\ 
    & & feature random combination\\ 
    \midrule
    CVAE & \cite{fajardo2021oversampling} & Conditional VAE \\ 
    \multirow{2}{*}{SMOTE-CLS} & \multirow{2}{*}{Ours} & VAE with customized latent space  + \\
    & &  KDE-based filtering + SMOTE \\ 
    \bottomrule
  \end{tabular}
\label{tab:models}
\end{table}

\subsubsection{Experimental Design}
We randomly split the datasets into training and test sets with an 8:2 ratio. Subsequently, the minority samples are augmented in the training set by each oversample method. We set the balanced dataset's oversampling ratio $\rho = 1$, ensuring an equal number of minor and major samples. To evaluate the oversampling method, we fit a classifier with the augmented training set and measure evaluation metrics using the test set. The above process is repeated ten times, and the mean and standard errors of the evaluation metrics are reported. Our experiments employ the random forest as our baseline classifier. 

For the training process of SMOTE-CLS, we configure the dimension of the latent variable to be 4 for datasets with more than 90 features and 2 for the remaining datasets. Suppose we denote the dimension of the latent variable as $h \in \{2, 4\}$. In that case, the architecture of the encoder and decoder are designed by the feedforward network with the dimension of the hidden layers given by $(8h, 4h, 2h, h, 2h, 4h, 8h)$ in order. The prior distribution of SMOTE-CLS is characterized by mean vectors: $\bmu_{M} = (-1, 1)$, $\bmu_{m} = (1, -1), \bmu_{m^*} = (-1, -1)$, and $\bmu_{M^*} = (1, 1)$, and variance: $s_c^2 = 0.1$ for $c \in \mathcal{Y}^*$. Like the simulation study, we set $K=5$ for the KNN classifier $f_K$ and employ XGBoost as $f_\eta$ in \eqref{eq: obj}. 

\subsubsection{Experimental Results}

\begin{sidewaystable}[ph!]
\caption{Classifcation performance (AUPRC) of different oversampling models using the random forest on the imbalanced datasets. The numbers within parentheses indicate their rankings, excluding BASE. The most favorable value is bolded.}
  \centering
{\footnotesize\setlength{\tabcolsep}{1pt} 
\begin{tabular}{lccccccccccc}
    \toprule
    \textsf{Dataset} & BASE & SMOTE & BSMOTE & SMOTE-ENN & KMSMOTE& DFBS & CVAE & DeepSMOTE & DDHS  & SMOTE-CLS \\
    \midrule
    \texttt{ecoli} & $0.736_{\pm0.020}$ & $0.674_{\pm 0.018}~(7)$ & $0.649_{\pm 0.034}~(8)$ & $0.720_{\pm0.031} ~ (4)$ & $0.711_{\pm0.019}~(6)$ & $0.639_{\pm 0.016}~(9)$ & $0.714_{\pm 0.025} ~ (5)$ & $0.736_{\pm 0.025}~(2)$ & $\mathbf{0.776_{\pm 0.019}~(1)}$ & $0.722_{\pm 0.026}~(3)$ \\
    \texttt{libras\_move} & $0.531_{\pm0.015}$ & $0.424_{\pm 0.042}~(9)$ & $0.437_{\pm 0.024}~(7)$ & $0.428_{\pm0.041}~(8)$ & $0.460_{\pm 0.029}~(6)$ & $0.583_{\pm 0.019}~(3)$ & $0.597_{\pm 0.030} ~ (2)$ & $\mathbf{0.637_{\pm 0.027}~(1)}$ & $0.543_{\pm 0.029}~(5)$ & $0.544_{\pm 0.032}~(4)$ \\
    \texttt{spectrometer} & $0.731_{\pm0.015}$ & $0.761_{\pm 0.015}~(4)$ & $0.741_{\pm 0.009}~(6)$ & $0.761_{\pm 0.011}~(3)$ & $0.756_{\pm 0.007}~(5)$ & $0.726_{\pm 0.023}~(8)$ & $0.737_{\pm 0.018}~(7)$ & $\mathbf{0.783_{\pm 0.017}~(1)}$ & $0.556_{\pm 0.015}~(9)$ & $0.780_{\pm 0.017}~(2)$ \\
    \texttt{oil} & $0.651_{\pm0.024}$ & $0.563_{\pm 0.017}~(8)$ & $0.619_{\pm 0.025}~(5)$ & $0.548_{\pm 0.020}~(9)$ & $0.610_{\pm 0.026}~(6)$ & $0.683_{\pm 0.014}~(2)$ & $0.565_{\pm 0.040}~(7)$ & $0.633_{\pm 0.024}~(4)$ & $\mathbf{0.712_{\pm 0.023}~(1)}$ & $0.662_{\pm 0.020}~(3)$ \\    
    \texttt{yeast} & $0.191_{\pm 0.018}$ & $0.340_{\pm 0.025}~(4)$ & $0.351_{\pm 0.022}~(3)$ & $0.330_{\pm 0.021}~(5)$ & $\mathbf{0.368_{\pm 0.021}~(1)}$ & $0.172_{\pm 0.006}~(8)$ & $0.238_{\pm 0.014}~(6)$ & $0.180_{\pm 0.015}~(7)$ & $0.158_{\pm 0.010}~(9)$ & $0.357_{\pm 0.022}~(2)$ \\
    \texttt{car\_eval} & $0.904_{\pm 0.012}$ & $0.920_{\pm 0.012}~(6)$ & $0.920_{\pm 0.011}~(5)$ & $0.921_{\pm 0.009}~(4)$ & $0.930_{\pm 0.005} ~ (2)$ & $0.927_{\pm 0.008} ~ (3)$ & $0.905_{\pm 0.015}~(8)$ & $0.909_{\pm 0.017}~(7)$ & $0.891_{\pm 0.008}~(9)$ & $\mathbf{0.931_{\pm 0.004}~(1)}$\\
    \texttt{us\_crime} & $0.386_{\pm 0.014}$ & $0.413_{\pm 0.009}~(5)$ & $\mathbf{0.443_{\pm 0.010}~(1)}$ & $0.395_{\pm 0.015}~(8)$ & $0.418_{\pm 0.006}~(4)$ & $0.428_{\pm 0.009}~(2)$ & $0.404_{\pm 0.007}~(6)$ & $0.399_{\pm 0.011}~(7)$ & $0.361_{\pm 0.010}~(9)$ & $0.425_{\pm 0.007}~(3)$\\
    \texttt{scene} & $0.264_{\pm 0.014}$ & $0.287_{\pm 0.009}~(4)$ & $0.295_{\pm 0.006}~(3)$ & $0.259_{\pm 0.007}~(7)$ &$0.296_{\pm 0.005}~(2)$  & $0.261_{\pm 0.005}~(6)$ & $0.233_{\pm 0.006}~(8)$ & $0.226_{\pm 0.008}~(9)$ & $0.266_{\pm 0.007}~(5)$ & $\mathbf{0.305_{\pm 0.008}~(1)}$ \\
    \texttt{abalone} & $0.284_{\pm 0.004}$ & $0.314_{\pm 0.004}~(2)$ & $0.312_{\pm 0.003}~(4)$ & $0.306_{\pm 0.003}~(6)$ & $0.235_{\pm 0.003}~(9)$ & $0.302_{\pm 0.004}~(7)$ & $0.313_{\pm 0.009}~(3)$ & $0.309_{\pm 0.006}~(5)$ & $0.265_{\pm 0.004}~(8)$ & $\mathbf{0.326_{\pm 0.004}~(1)}$\\
    \texttt{optical\_digits} & $0.914_{\pm 0.004}$ & $0.926_{\pm 0.004}~(2)$ & $0.876_{\pm 0.007}~(7)$ & $0.914_{\pm 0.006}~(3)$ & $0.905_{\pm 0.006}~(4)$ & $0.882_{\pm 0.004}~(6)$ & $0.870_{\pm 0.006}~(8)$ & $0.864_{\pm 0.006}~(9)$ & $0.888_{\pm 0.003}~(5)$ & $\mathbf{0.929_{\pm 0.003}~ (1)}$ \\
    \texttt{satimage} & $0.711_{\pm 0.004}$ & $0.683_{\pm 0.006}~(4)$ & $0.577_{\pm 0.007}~(7)$ & $0.574_{\pm 0.011}~(8)$ & $0.532_{\pm 0.008}~(9)$ & $0.656_{\pm 0.004}~(6)$ & $0.723_{\pm 0.003}~(2)$ & $\mathbf{0.725_{\pm 0.004}~(1)}$ & $0.679_{\pm 0.007}~(5)$ & $0.697_{\pm 0.003}~ (3)$ \\
    \texttt{isolet} & $0.796_{\pm 0.010}$ & $0.789_{\pm 0.009}~(2)$ & $0.697_{\pm 0.009}~(9)$ & $0.740_{\pm 0.012}~(6)$ & $0.763_{\pm 0.010}~(5)$ & $0.724_{\pm 0.008}~(7)$ & $0.768_{\pm 0.010}~(4)$ & $0.769_{\pm 0.008}~(3)$ & $0.698_{\pm 0.008}~(8)$ & $\mathbf{0.798_{\pm 0.009}~ (1)}$ \\
    Average rank & - & 4.75 & 5.42 & 5.92 & 5.00 & 5.58 & 5.5 & 4.58 & 6.17 & \textbf{2.08} \\ 
    \bottomrule
  \end{tabular}
}
\label{tab:auprc}
\end{sidewaystable}

\begin{sidewaystable}[ph!]
  \caption{Classifcation performance (AUC) of different oversampling models using the random forest on the imbalanced datasets. The numbers within parentheses indicate their rankings, excluding BASE. The most favorable value is bolded.}
  \centering
{\footnotesize\setlength{\tabcolsep}{1pt} 
\begin{tabular}{lcccccccccc}
    \toprule
    \textsf{Dataset} & BASE & SMOTE & BSMOTE & SMOTE-ENN & KMSMOTE & DFBS & CVAE & DeepSMOTE & DDHS  & SMOTE-CLS \\
    \midrule
    \texttt{ecoli} & $0.966_{\pm0.004}$ & $0.961_{\pm0.003}~(7)$ & $0.951_{\pm 0.004}~(9)$ & $0.966_{\pm0.004}~(5)$ & $0.967_{\pm0.003}~(3)$ & $0.953_{\pm 0.003}~(8)$ & $0.969_{\pm 0.003}~(2)$ & $\mathbf{0.970_{\pm 0.004}~(1)}$ & $0.966_{\pm 0.004}~(5)$ & $0.966_{\pm 0.003}~(4)$ \\
    \texttt{libras\_move} & $0.830_{\pm0.024}$ & $0.777_{\pm 0.025}~(9)$ & $0.793_{\pm 0.016}~(8)$ & $0.800_{\pm0.025}~(7)$ & $0.827_{\pm 0.013}~(5)$ & $0.849_{\pm 0.029}~(4)$ & $0.867_{\pm 0.028}~(2)$ & $\mathbf{0.887_{\pm 0.020}~(1)}$ & $0.850_{\pm 0.017}~(3)$ & $0.811_{\pm 0.025}~(6)$ \\
    \texttt{spectrometer} & $0.858_{\pm0.008}$ & $0.886_{\pm 0.009}~(5)$ & $0.864_{\pm 0.006}~(8)$ & $0.887_{\pm 0.008}~(4)$ & $0.867_{\pm 0.002}~(7)$ & $0.869_{\pm 0.006}~(6)$ & $0.908_{\pm 0.008}~(2)$ & $0.908_{\pm 0.013}~(3)$ & $0.845_{\pm 0.008}~(9)$ & $\mathbf{0.909_{\pm 0.012}~(1)}$ \\
    \texttt{oil} & $0.968_{\pm0.006}$ & $0.972_{\pm0.002}~(4)$ & $0.973_{\pm 0.005}~(3)$ & $0.972_{\pm0.003}~(5)$ & $0.942_{\pm 0.012}~(9)$ & $\mathbf{0.978_{\pm 0.003}~(1)}$ & $0.952_{\pm 0.015}~(8)$ & $0.969_{\pm 0.004}~(6)$ & $0.964_{\pm 0.012}~(7)$ & $0.976_{\pm 0.003}~(2)$ \\
    \texttt{yeast} & $0.801_{\pm 0.017}$ & $0.933_{\pm 0.006}~(2)$ & $0.908_{\pm 0.006}~(3)$ & $\mathbf{0.941_{\pm 0.004}~(1)}$ & $0.900_{\pm 0.016}~(5)$ & $0.812_{\pm 0.017}~(7)$ & $0.881_{\pm 0.007}~(6)$ & $0.809_{\pm 0.012}~(8)$ & $0.809_{\pm 0.016}~(9)$ & $0.900_{\pm 0.010}~(4)$ \\
    \texttt{car\_eval} & $0.994_{\pm 0.001}$ & $0.995_{\pm 0.001}~(2)$ & $0.995_{\pm 0.001}~(2)$ & $\mathbf{0.995_{\pm 0.000}~(1)}$ & $0.994_{\pm 0.000}~(4)$ & $0.994_{\pm 0.001}~(6)$ & $0.992_{\pm 0.002}~(8)$ & $0.993_{\pm 0.001}~(7)$ & $0.990_{\pm 0.001}~(9)$ & $0.994_{\pm 0.000}~(4)$ \\
    \texttt{us\_crime} & $0.884_{\pm 0.005}$ & $0.894_{\pm 0.004}~(6)$ & $\mathbf{0.908_{\pm 0.003}~(1)}$ & $0.904_{\pm 0.003}~(2)$ & $0.898_{\pm 0.004}~(4)$ & $0.900_{\pm 0.004}~(3)$ & $0.890_{\pm 0.004}~(7)$ & $0.883_{\pm 0.003}~(8)$  & $0.868_{\pm 0.004}~(9)$ & $0.896_{\pm 0.004}~(5)$\\
    \texttt{scene} & $0.728_{\pm 0.010}$ & $0.770_{\pm 0.007}~(6)$ & $\mathbf{0.780_{\pm 0.007}~(1)}$ & $0.773_{\pm 0.004}~(3)$ & $0.775_{\pm 0.005}~(2)$ & $0.770_{\pm 0.004}~(4)$ & $0.703_{\pm 0.009}~(8)$ & $0.691_{\pm 0.008}~(9)$ & $0.715_{\pm 0.009}~(7)$ & $0.770_{\pm 0.004}~(4)$ \\
    \texttt{abalone} & $0.836_{\pm 0.002}$ & $0.846_{\pm 0.002}~(2)$ & $0.846_{\pm 0.002}~(2)$ & $0.843_{\pm 0.002}~(4)$ & $0.813_{\pm 0.003}~(9)$ & $0.835_{\pm 0.002}~(6)$ & $0.838_{\pm 0.003}~(7)$ & $0.842_{\pm 0.002}~(5)$ & $0.826_{\pm 0.003}~(8)$ & $\mathbf{0.849_{\pm 0.003}~(1)}$\\
    \texttt{optical\_digits} & $0.982_{\pm 0.001}$ & $0.986_{\pm 0.001}~(2)$ & $0.978_{\pm 0.002}~(5)$ & $0.984_{\pm 0.001}~(3)$ & $0.981_{\pm 0.002}~(4)$ & $0.972_{\pm 0.001}~(7)$ & $0.971_{\pm 0.002}~(8)$ & $0.968_{\pm 0.002}~(9)$ & $0.977_{\pm 0.001}~(6)$ & $\mathbf{0.987_{\pm 0.001}~ (1)}$ \\
    \texttt{satimage} & $0.934_{\pm 0.002}$ & $0.942_{\pm 0.002}~(3)$ & $0.930_{\pm 0.002}~(8)$ & $0.930_{\pm 0.002}~(8)$ & $0.931_{\pm 0.001}~(7)$ & $0.935_{\pm 0.001}~(6)$ & $0.941_{\pm 0.002}~(4)$ & $\mathbf{0.944_{\pm 0.001}~(1)}$ & $0.937_{\pm 0.002}~(5)$ & $0.943_{\pm 0.001}~(2)$ \\
    \texttt{isolet} & $0.973_{\pm 0.001}$ & $\mathbf{0.979_{\pm 0.001}~(1)}$ & $0.961_{\pm 0.002}~(8)$ & $0.976_{\pm 0.001}~(4)$ & $0.976_{\pm 0.001}~(3)$ & $0.970_{\pm 0.001}~(6)$ & $0.971_{\pm 0.002}~(5)$ & $0.969_{\pm 0.002}~(7)$ & $0.964_{\pm 0.001}~(9)$ & $0.977_{\pm 0.001}~(2)$ \\
    Average rank & - & 4.08 & 4.83 & 3.92 & 5.17 & 5.33 & 5.58 & 5.42 & 7.17 & \textbf{3.00} \\ 
    \bottomrule
  \end{tabular}
  }
\label{tab:auc}
\end{sidewaystable}

Table \ref{tab:auprc} summarizes the average AUPRC and standard error of 10 repeated experiments. Regarding imbalanced classification performance, SMOTE-CLS exhibits consistently better performance than other methods, regardless of whether the datasets have low or high imbalance ratios. SMOTE-CLS is the top-performing method in 7 out of 12 datasets, with an average ranking of 2.08. It's noteworthy that SMOTE-CLS consistently outperforms SMOTE across all datasets, confirming its ability to address the weaknesses of SMOTE and enhance its performance. Additionally, our proposed model demonstrates its effectiveness, particularly in high-dimensional datasets like \texttt{scene} and \texttt{isolet}, outperforming deep learning approaches. 

Table \ref{tab:auc} summarizes the average AUC and standard error of 10 repeated experiments. When assessing overall classification performance, SMOTE-CLS consistently ranks among the top-performing models, with an average ranking of 3.00. It is worth noting that the SMOTE-based oversampling methods, including our proposed SMOTE-CLS, outperform the deep learning methods in terms of AUC. This observation verifies that the samples generated by SMOTE-based methods are more effective than those generated by deep learning methods in the benchmark dataset. 

Our real data analysis demonstrates that SMOTE-CLS significantly enhances minor labels (or positive samples) while maintaining a solid overall classification performance. Additionally, the comparison with SMOTE and BSMOTE on average performance highlights that simple filtering algorithms, like those used in BSMOTE, may not be as effective as our data-adaptive filtering in handling imbalanced classification tasks. These results justify the importance of our approach, which takes into account the data's characteristics to elaborate decisions about filtering and oversampling. For additional performance metrics, please refer to Appendix \ref{app:metrics}.

\subsubsection{Visualizations}

\begin{figure}[!ht]
    \centering
    \subfigure[DFBS]{\includegraphics[width=0.24\linewidth]{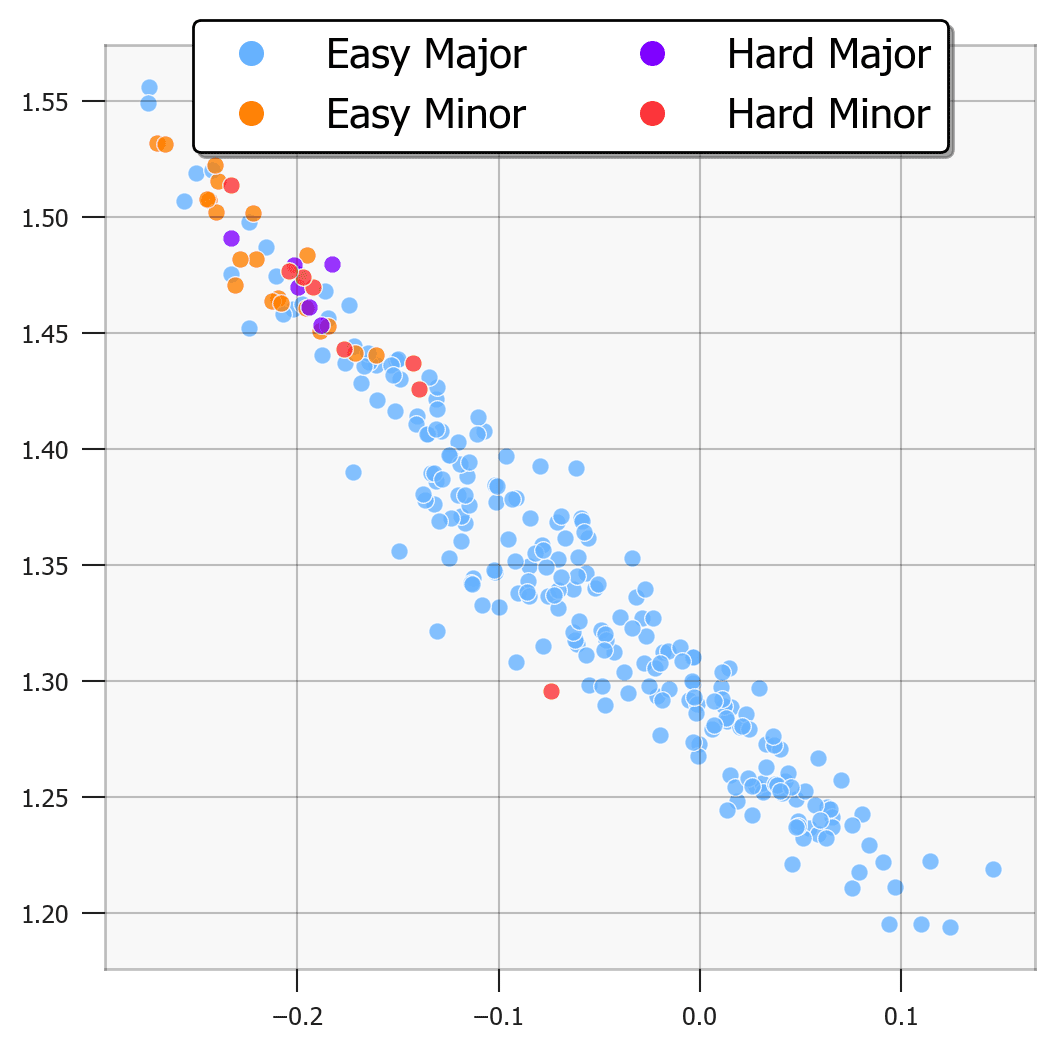}}
    \subfigure[DeepSMOTE]{\includegraphics[width=0.24\linewidth]{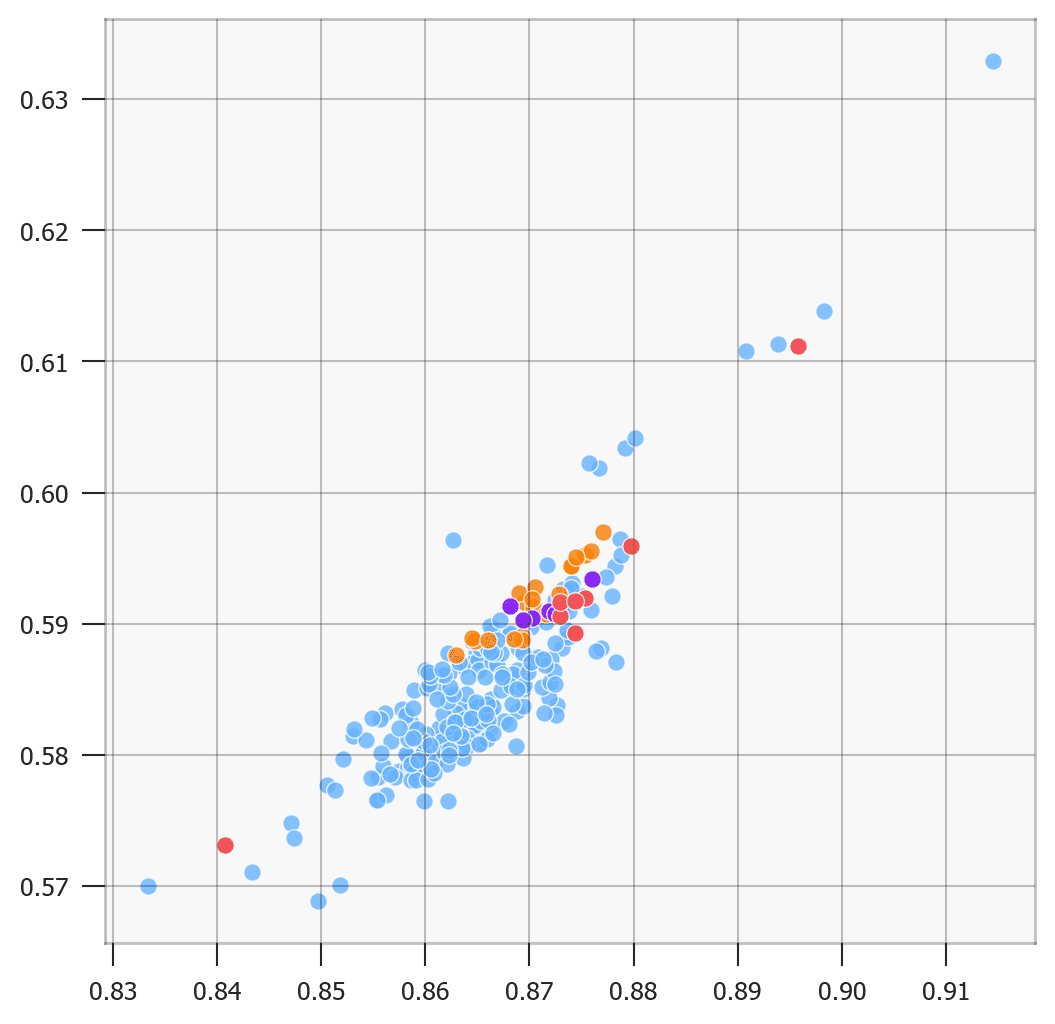}}
    \subfigure[DDHS]{\includegraphics[width=0.24\linewidth]{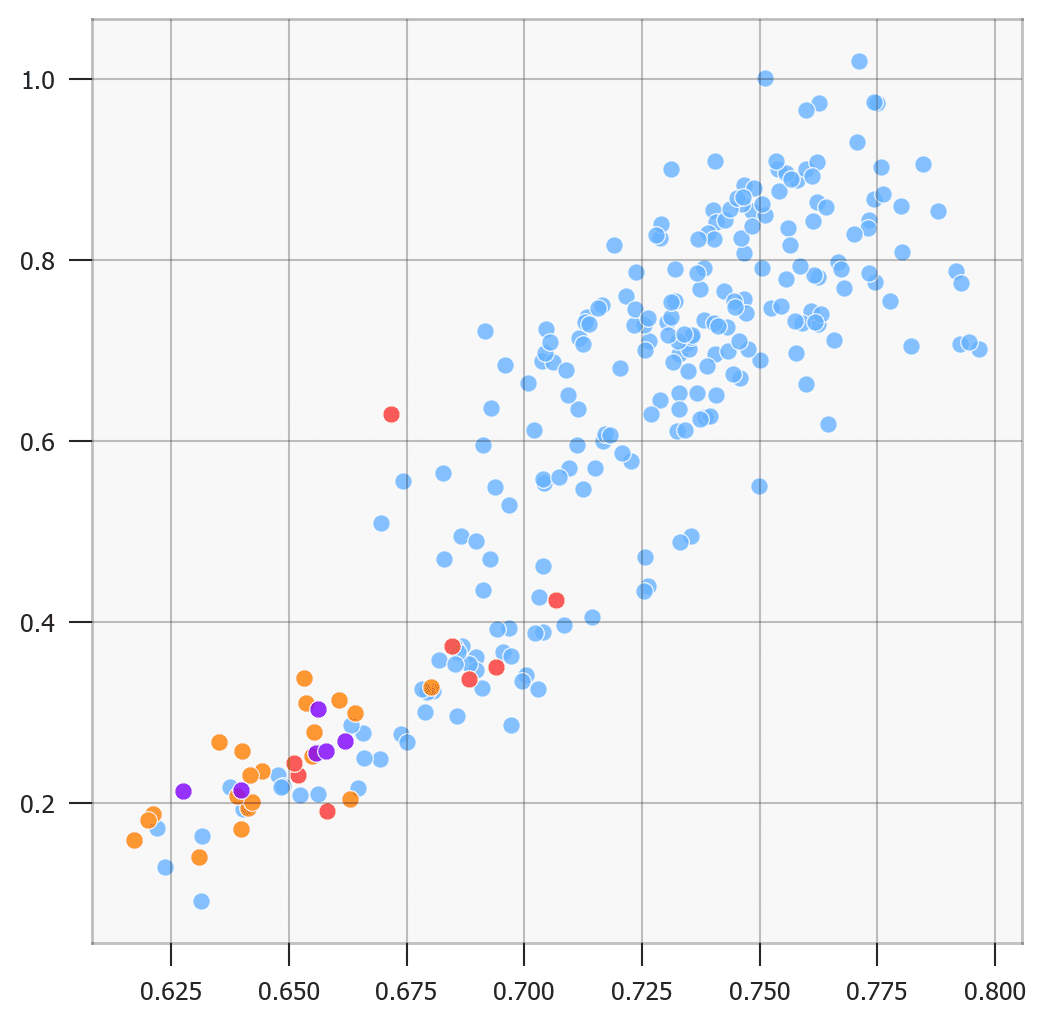}}
    \subfigure[SMOTE-CLS]{\includegraphics[width=0.24\linewidth]{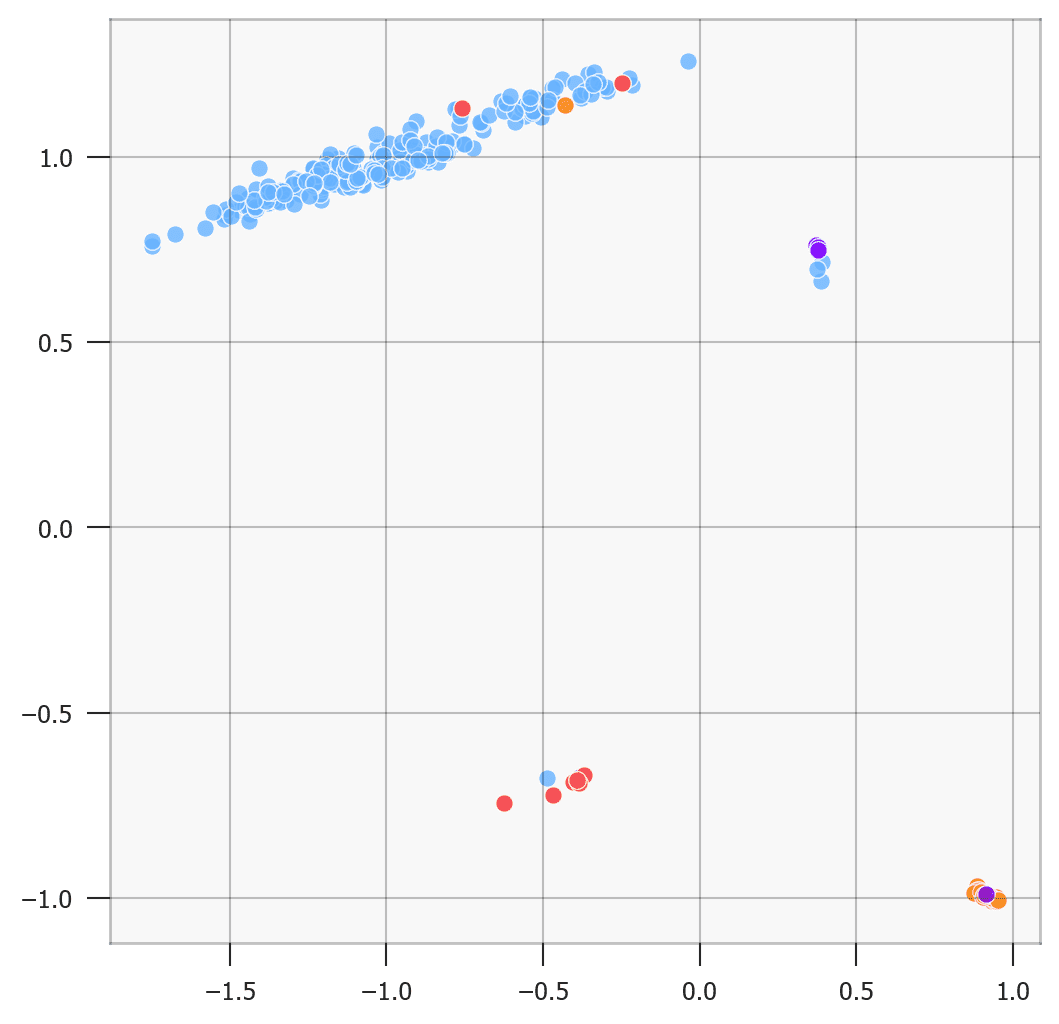}}
    \subfigure[DFBS]{\includegraphics[width=0.24\linewidth]{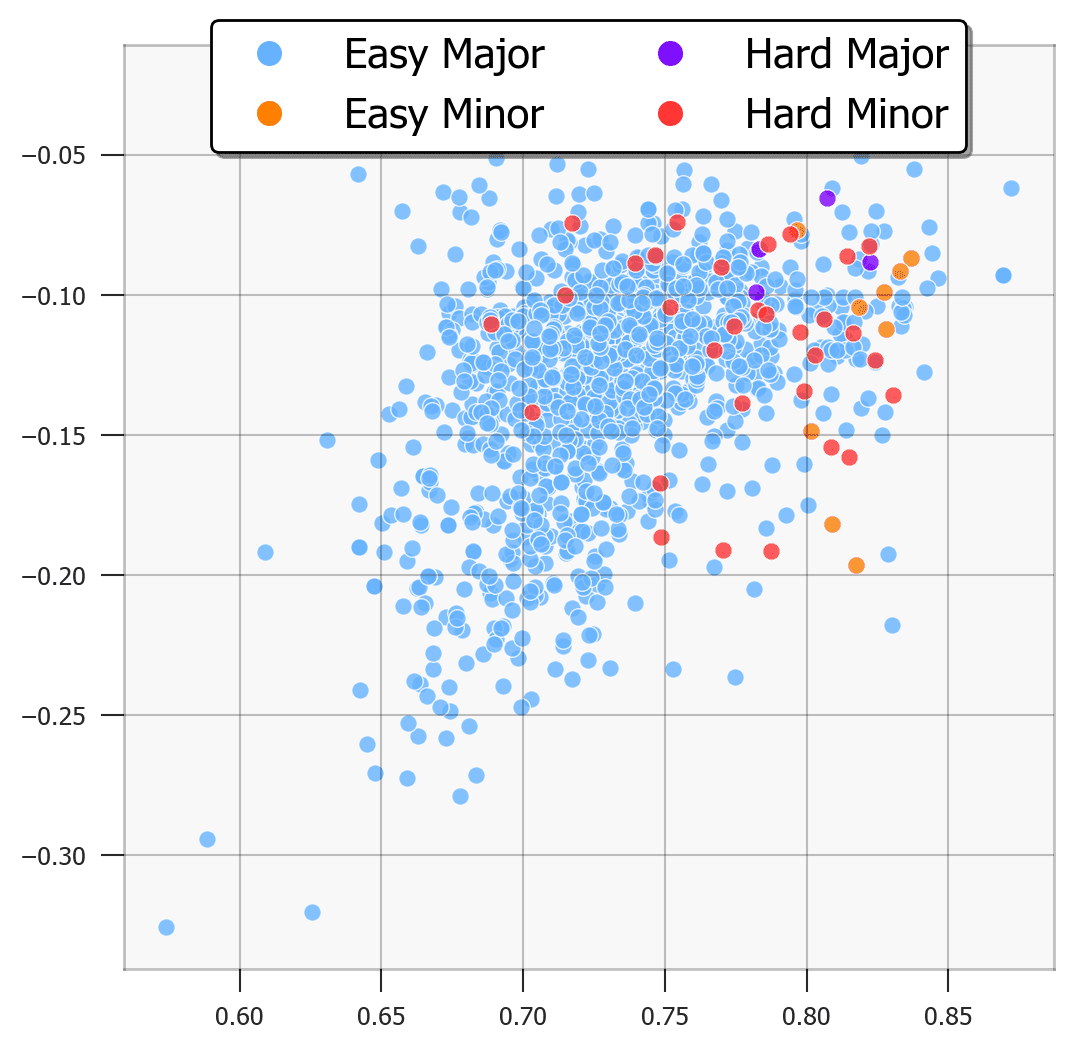}}
    \subfigure[DeepSMOTE]{\includegraphics[width=0.24\linewidth]{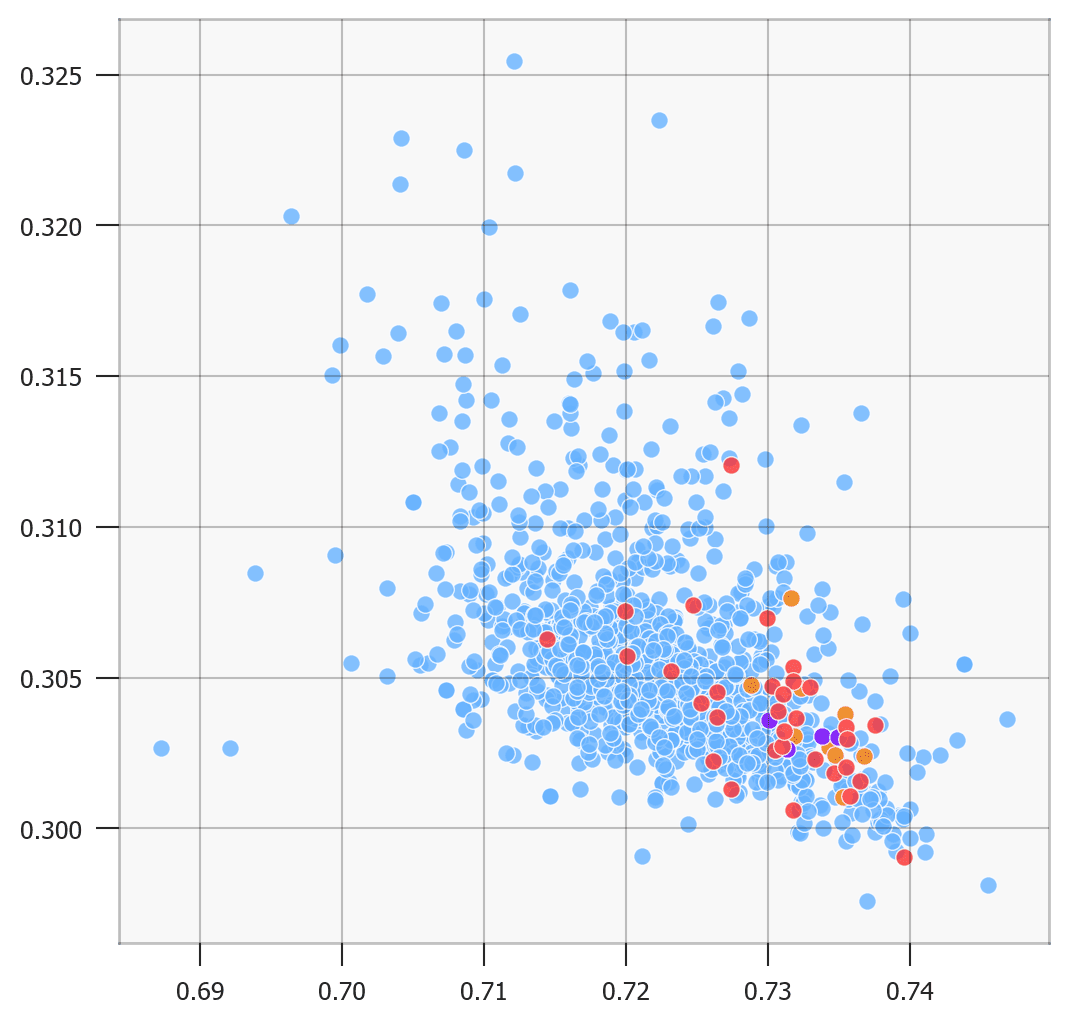}}
    \subfigure[DDHS]{\includegraphics[width=0.24\linewidth]{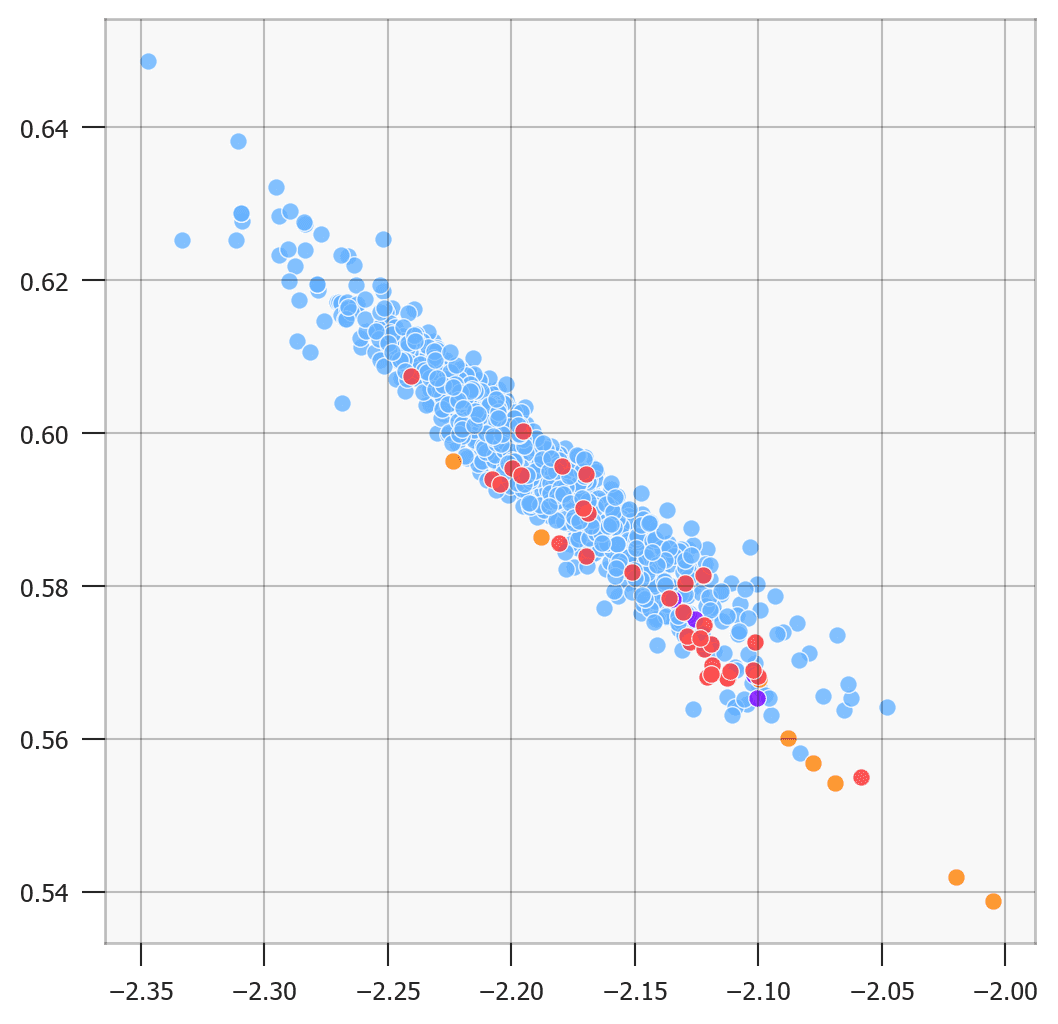}}
    \subfigure[SMOTE-CLS]{\includegraphics[width=0.24\linewidth]{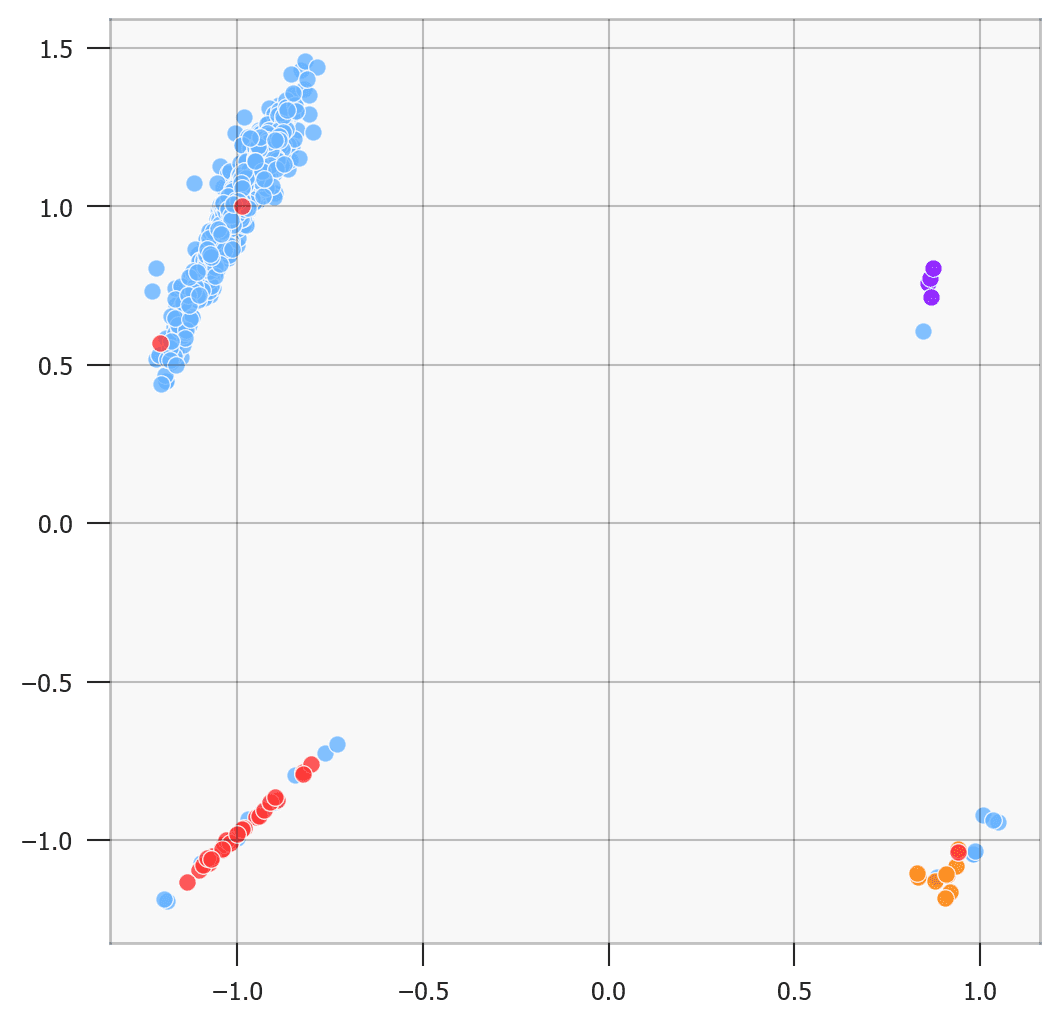}}
    \subfigure[DFBS]{\includegraphics[width=0.24\linewidth]{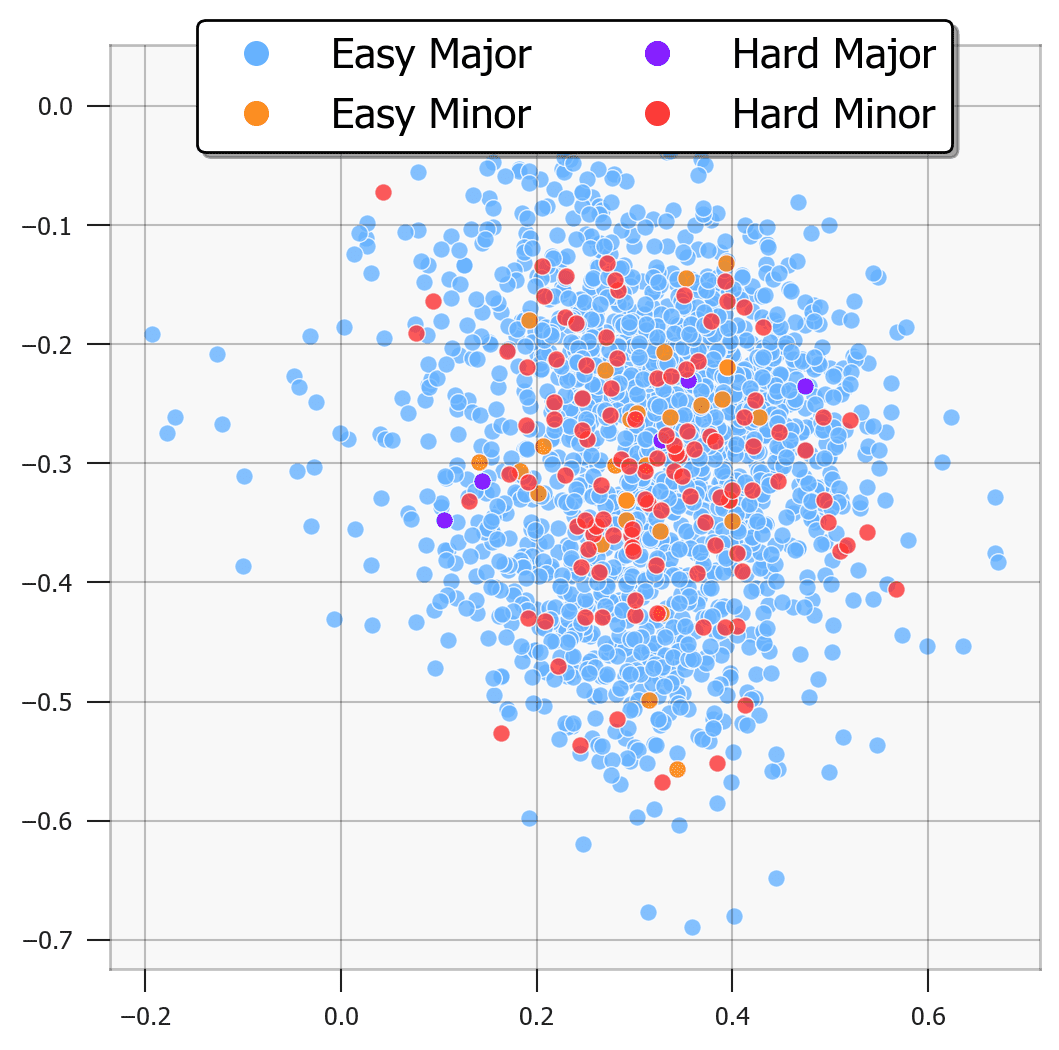}}
    \subfigure[DeepSMOTE]{\includegraphics[width=0.24\linewidth]{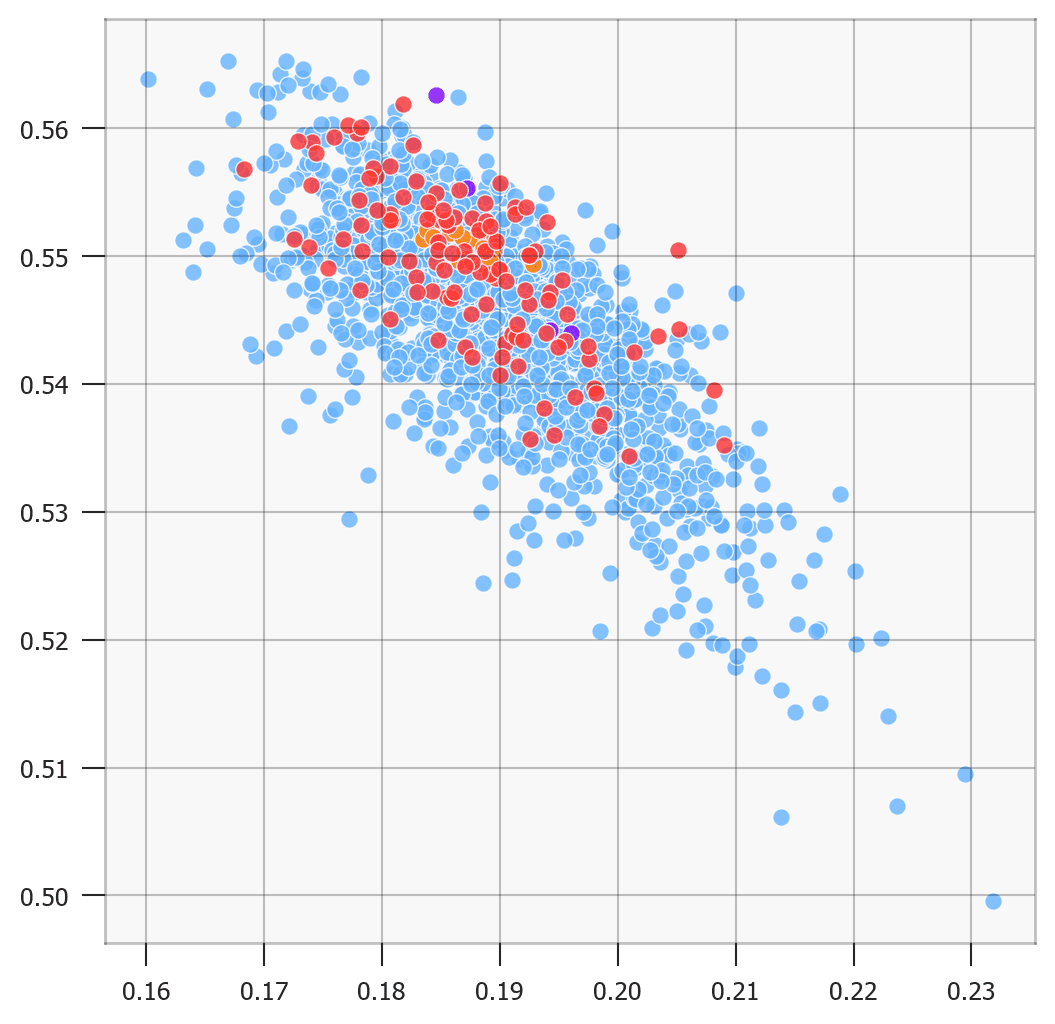}}
    \subfigure[DDHS]{\includegraphics[width=0.24\linewidth]{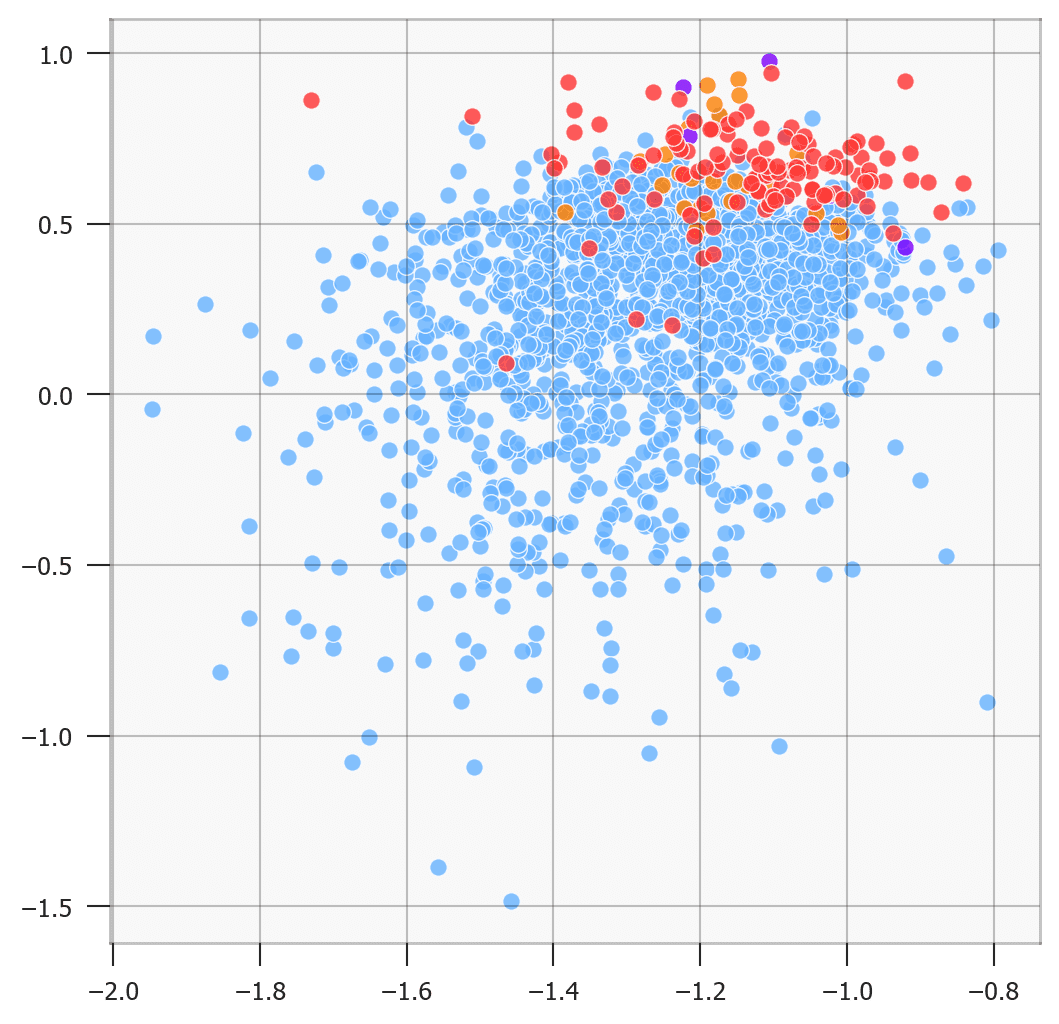}}
    \subfigure[SMOTE-CLS]{\includegraphics[width=0.24\linewidth]{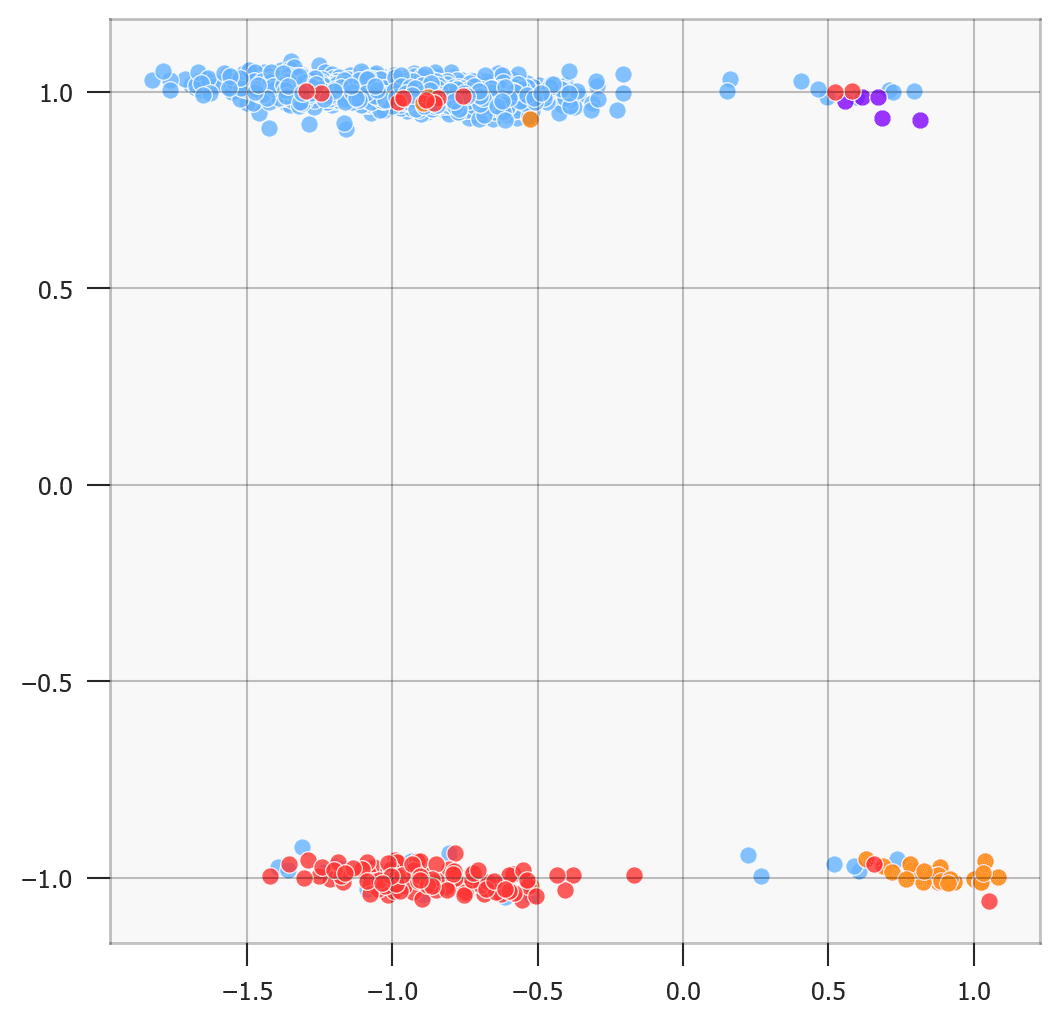}}
    \subfigure[DFBS]{\includegraphics[width=0.24\linewidth]{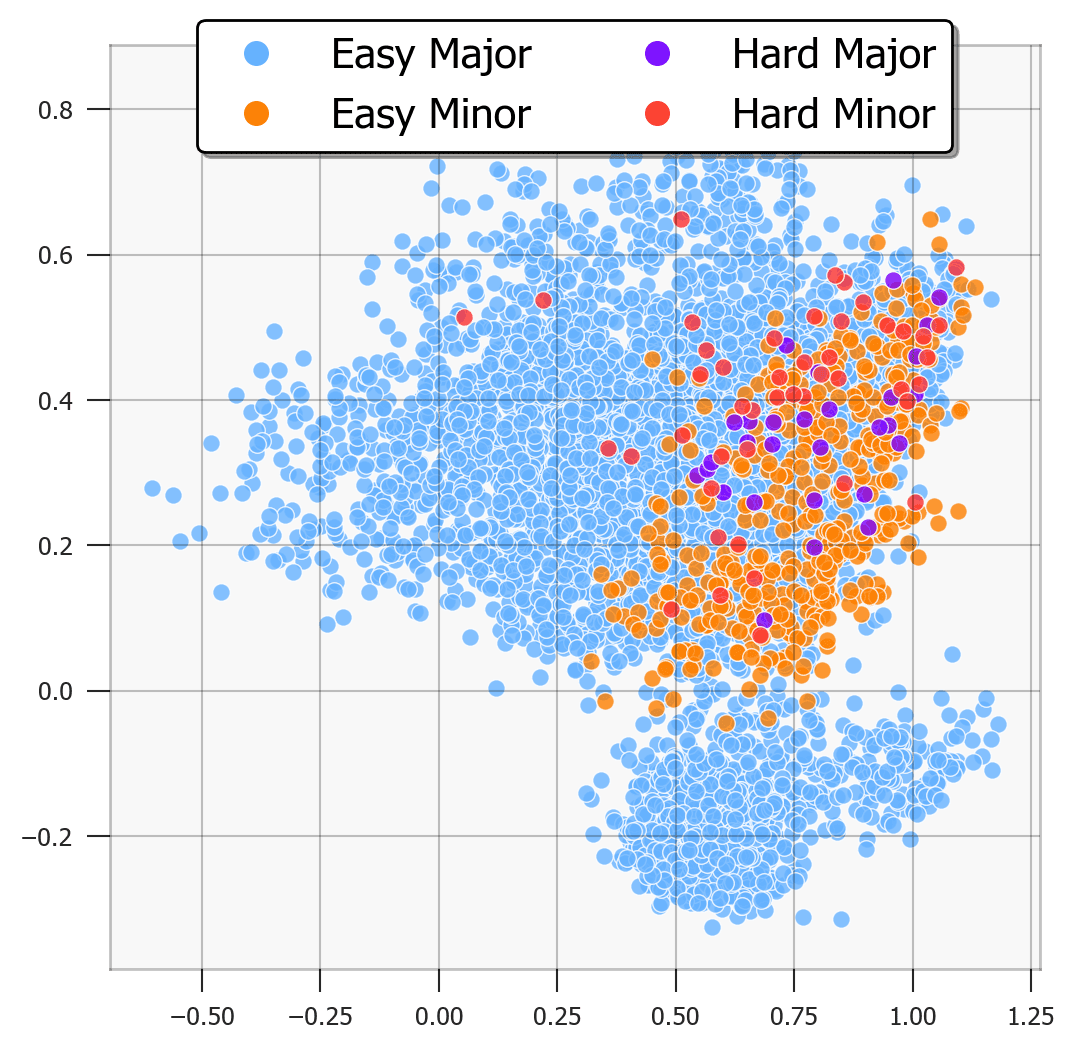}}
    \subfigure[DeepSMOTE]{\includegraphics[width=0.24\linewidth]{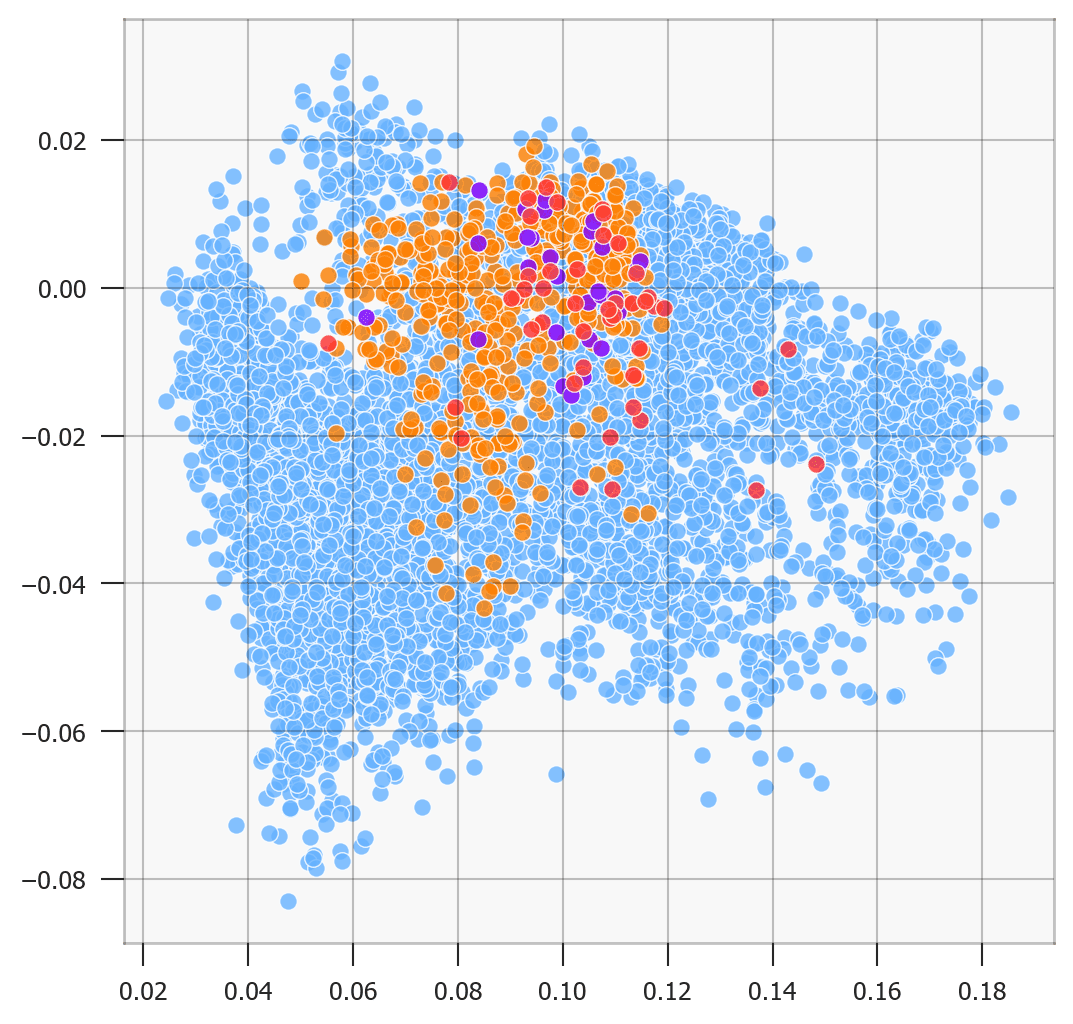}}
    \subfigure[DDHS]{\includegraphics[width=0.24\linewidth]{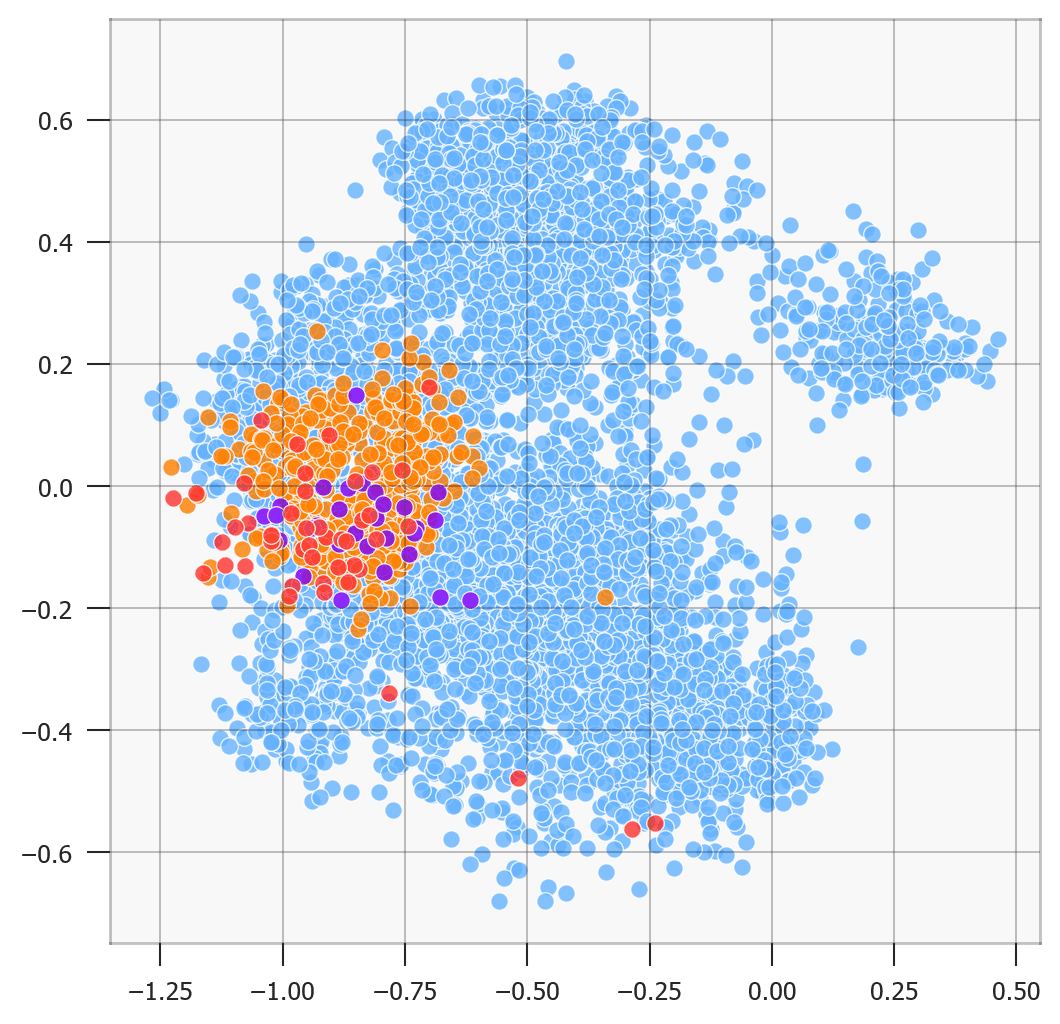}}
    \subfigure[SMOTE-CLS]{\includegraphics[width=0.24\linewidth]{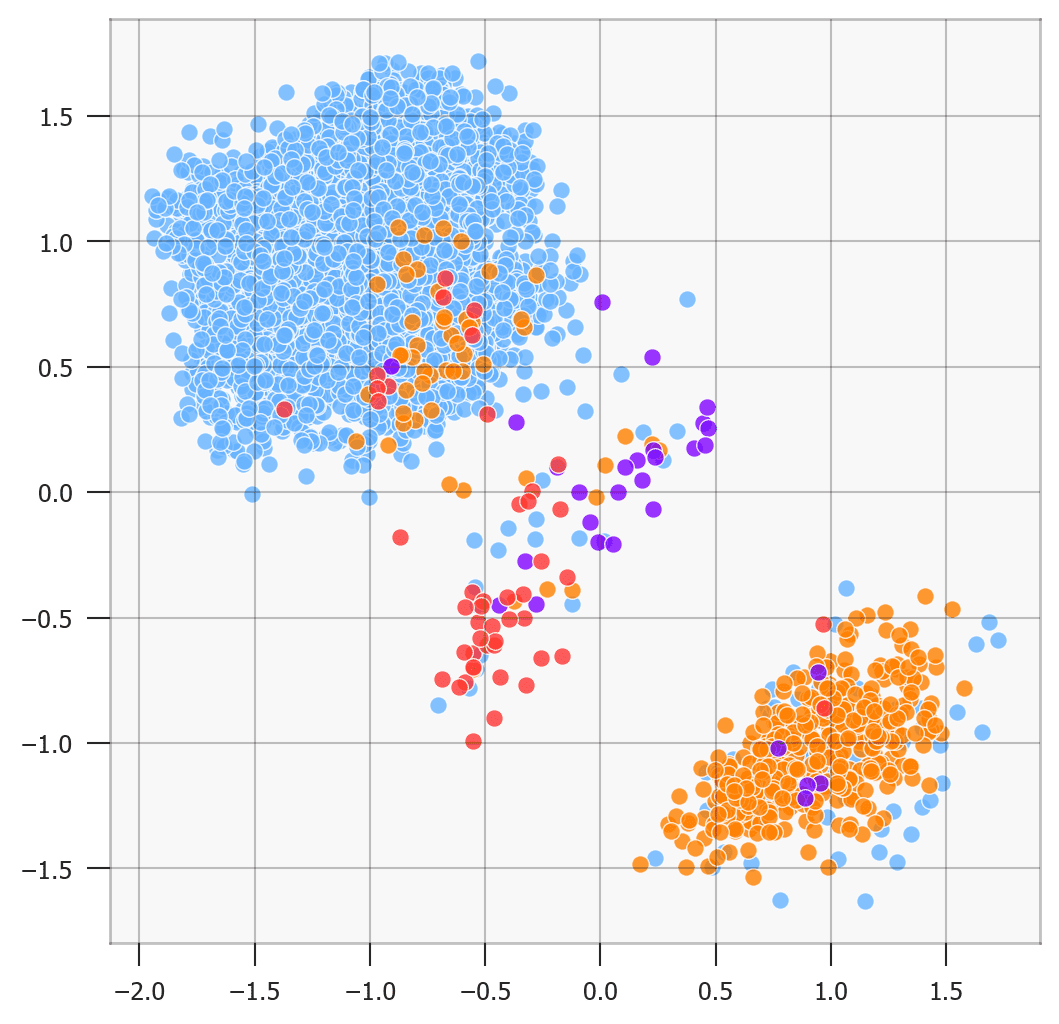}}
    \caption{Visualization of latent spaces for 4 datasets: (a) - (d): \texttt{ecoli} dataset, (e) - (h): \texttt{yeast} dataset, (i) - (l): \texttt{scene} dataset, (m) - (p): \texttt{isolet} dataset. The assignment of labels is determined by the KNN classifier $f_\kappa$. In the case of SMOTE-CLS, the noise samples are identified using our filtering algorithm and subsequently removed from the training set.}
    \label{fig:real_results}
\end{figure}

\begin{figure}[!ht]
    \centering
    \subfigure[SMOTE]{\includegraphics[width=0.3\linewidth]{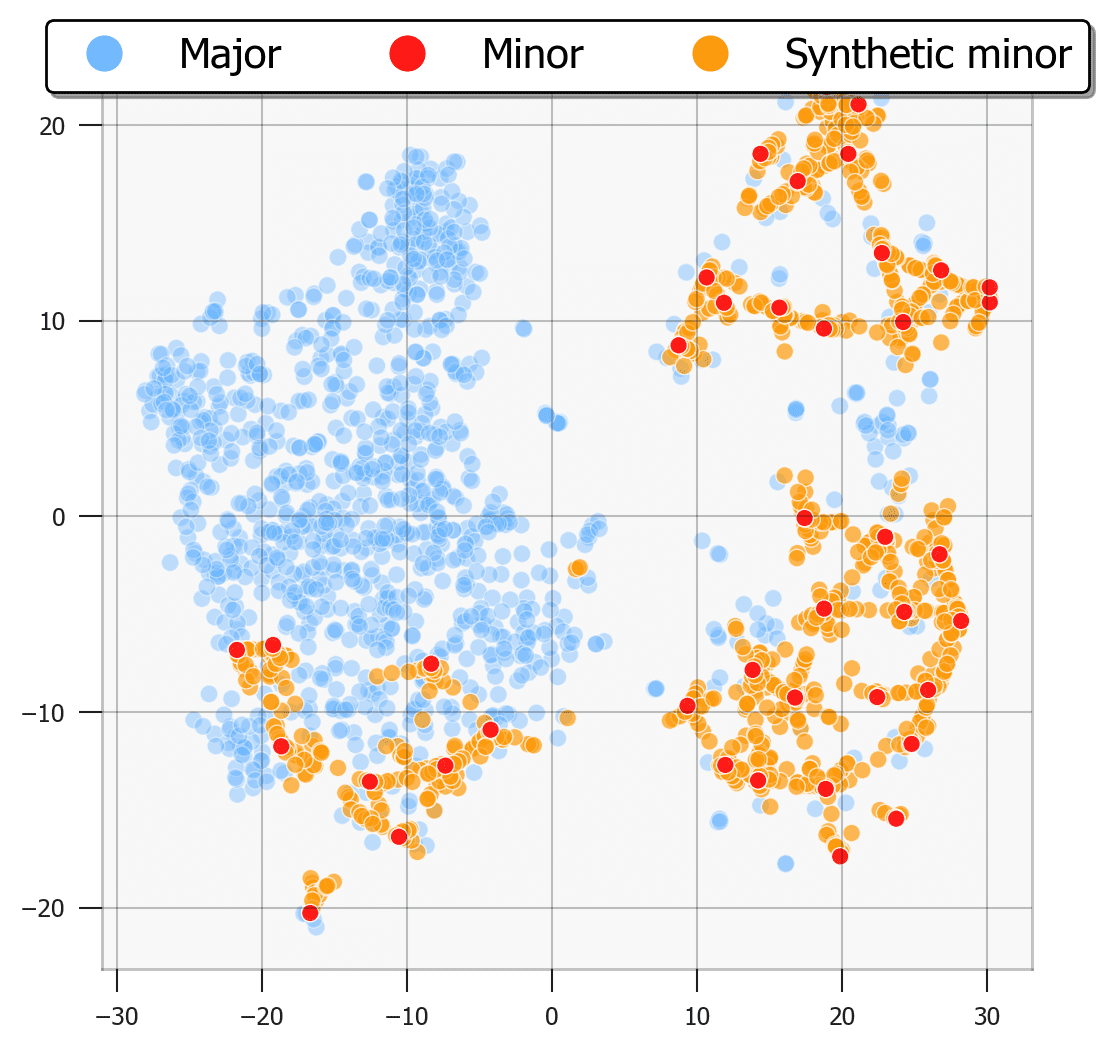}}
    \subfigure[BSMOTE]{\includegraphics[width=0.3\linewidth]{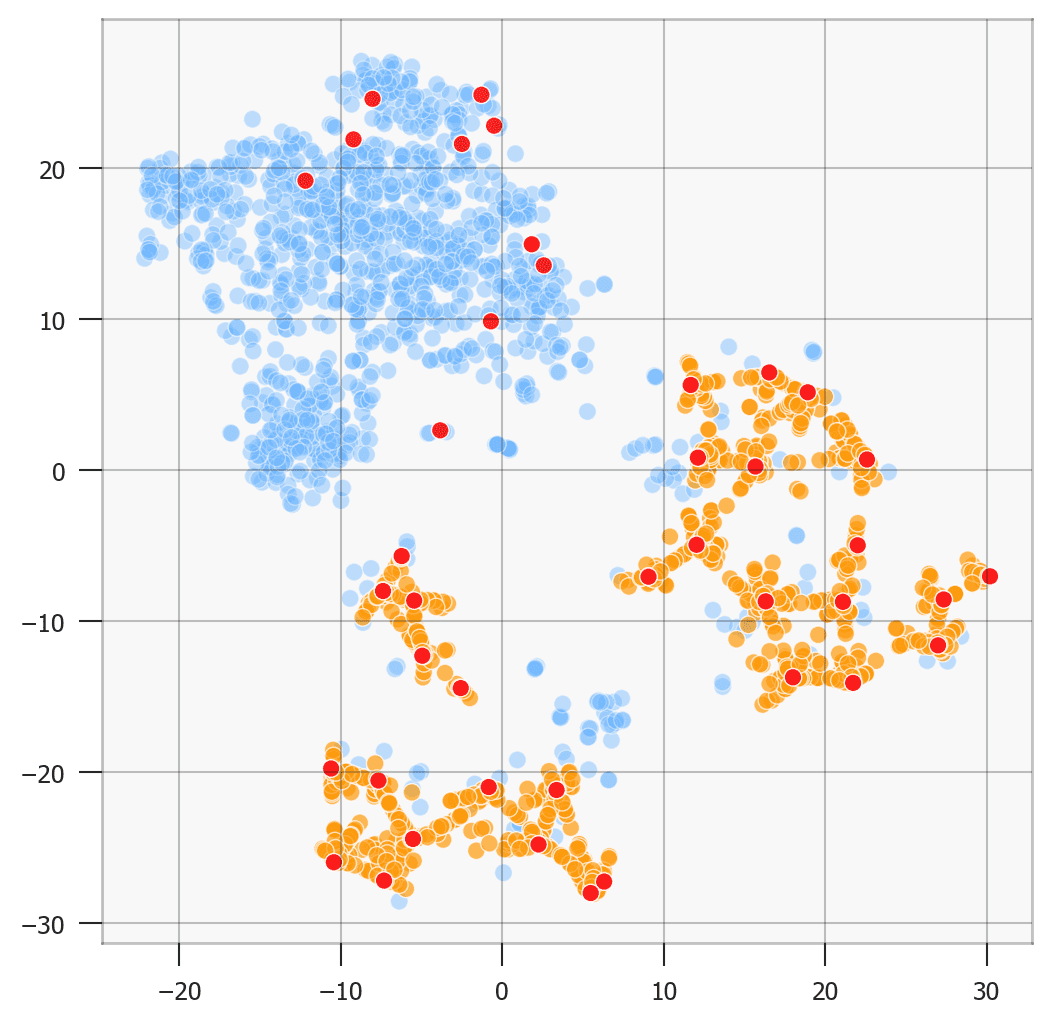}}
    \subfigure[SMOTE-ENN]{\includegraphics[width=0.3\linewidth]{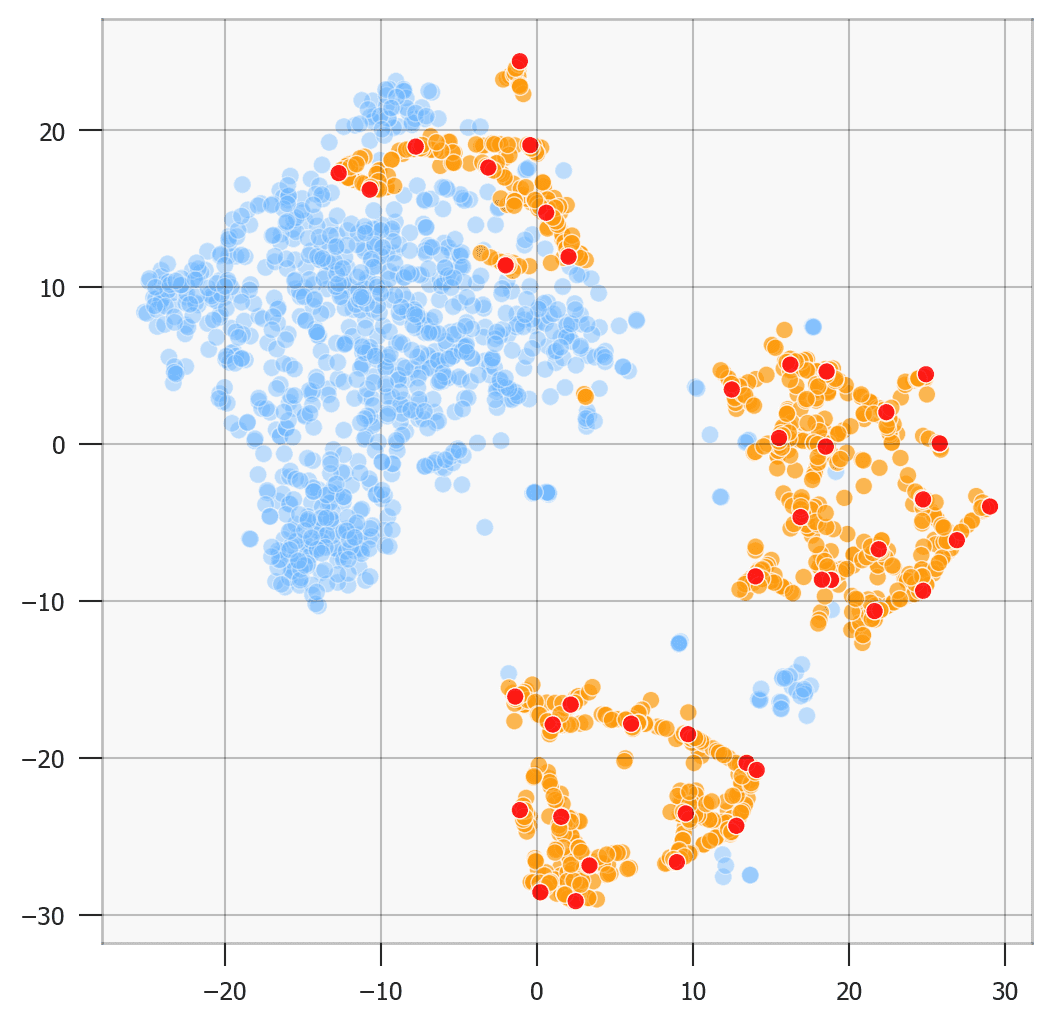}}
    \subfigure[KMSMOTE]{\includegraphics[width=0.3\linewidth]{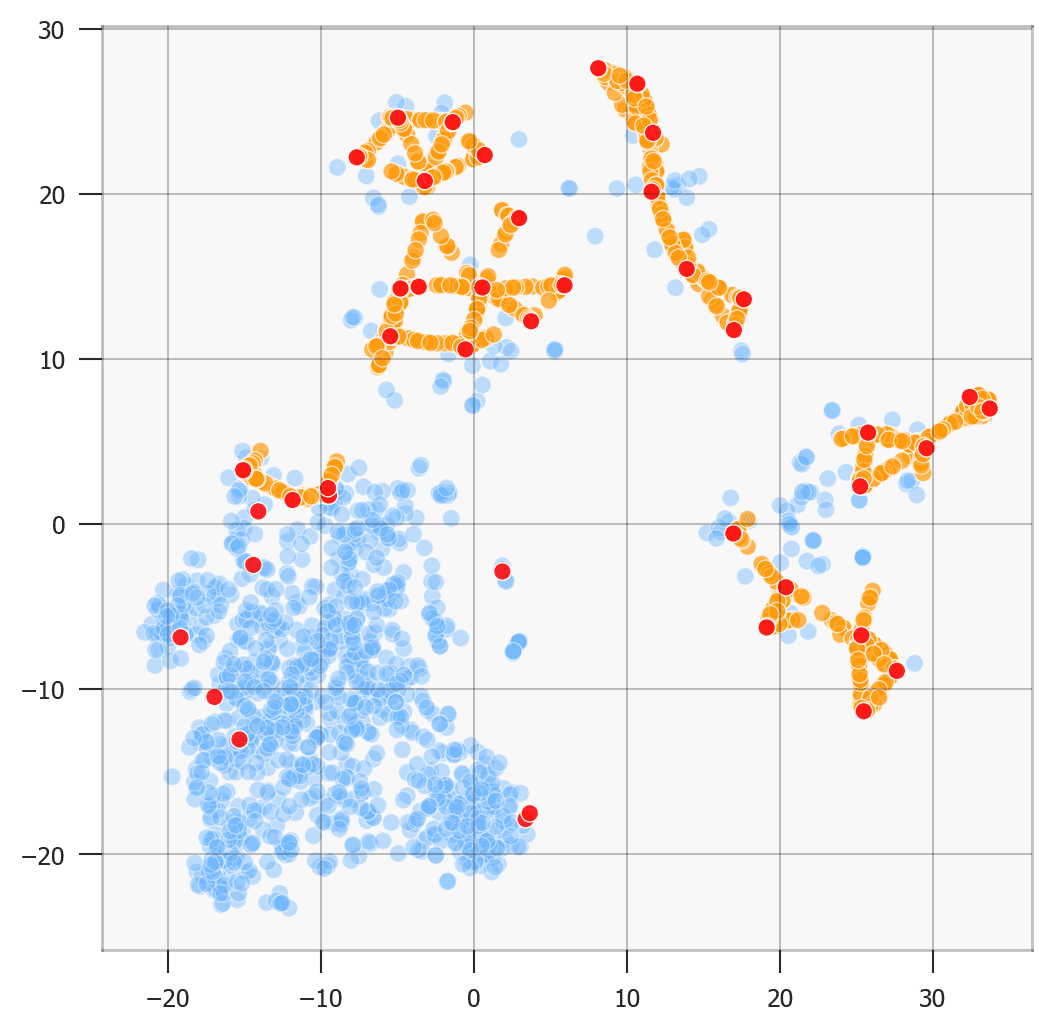}}
    \subfigure[DFBS]{\includegraphics[width=0.3\linewidth]{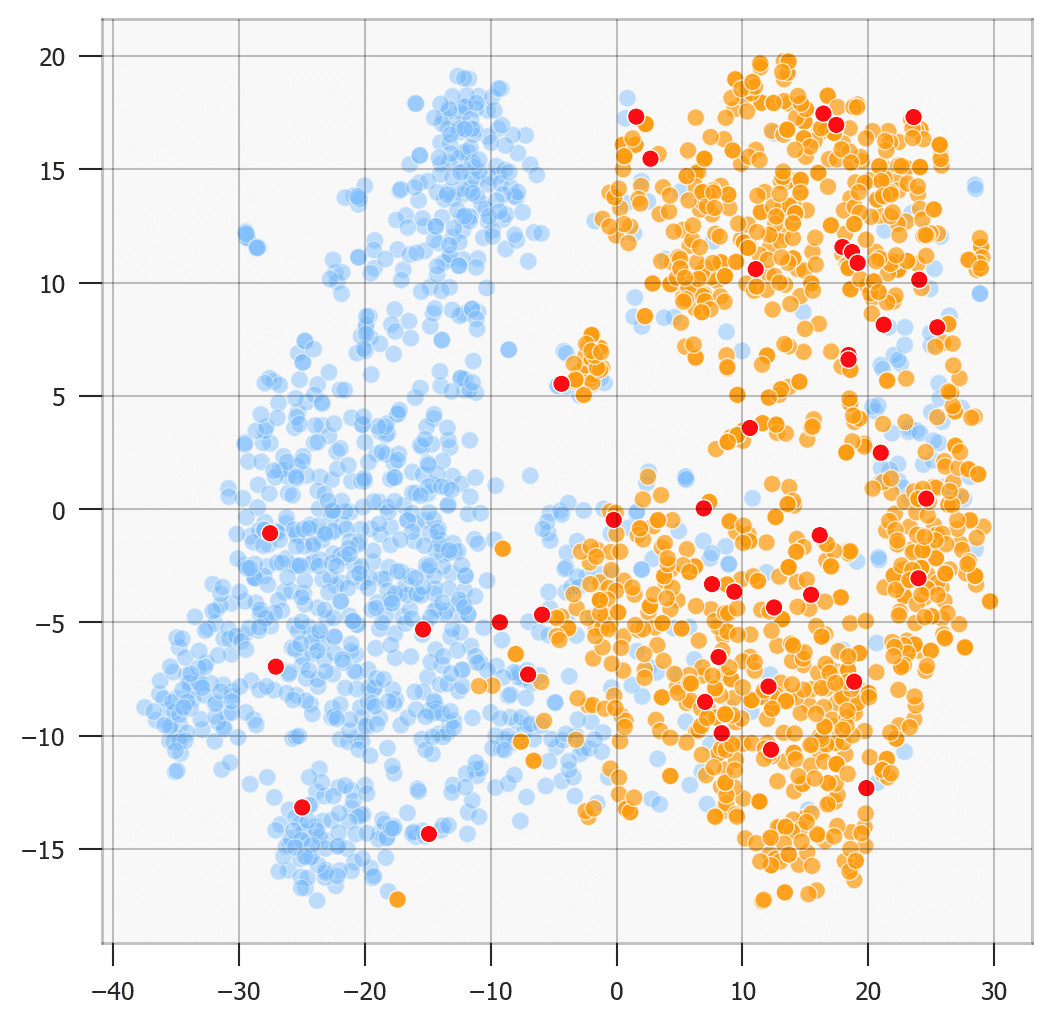}}
    \subfigure[CVAE]{\includegraphics[width=0.3\linewidth]{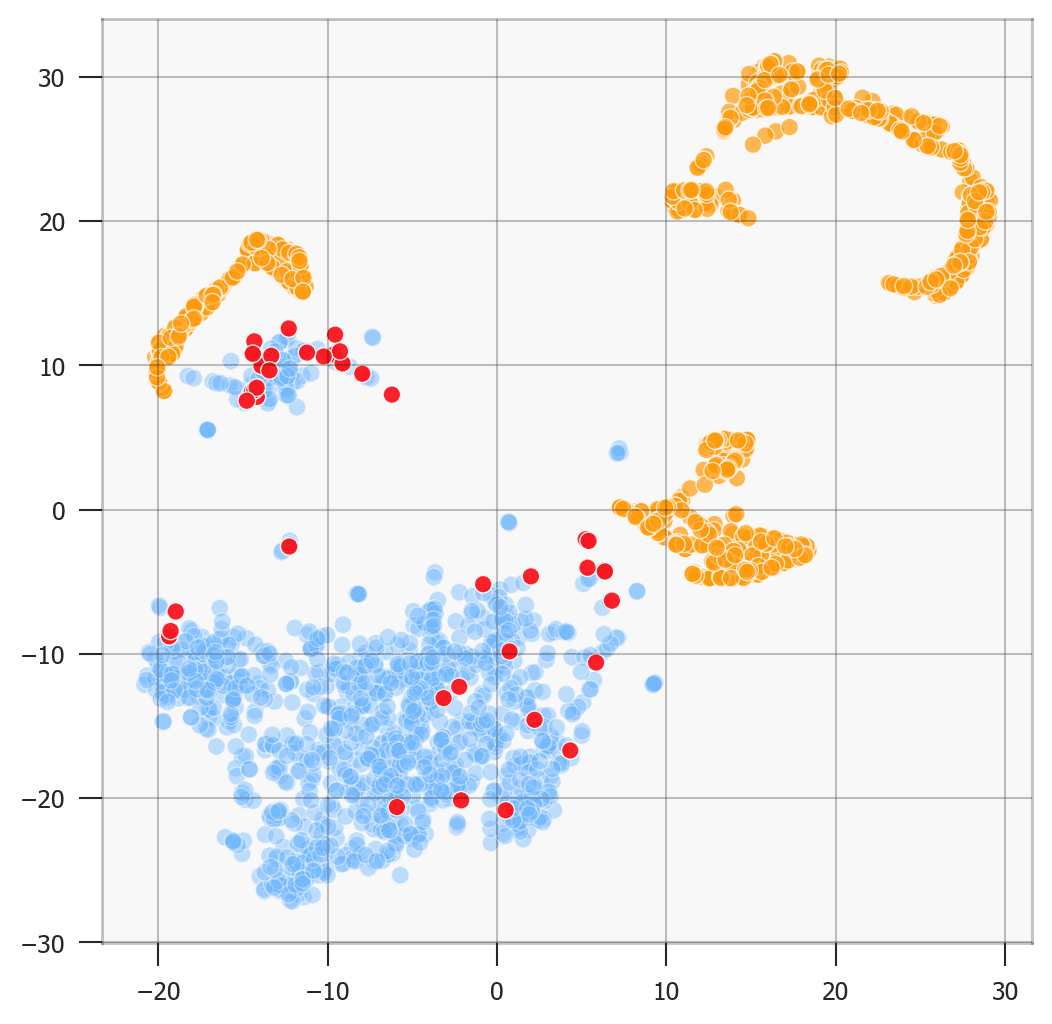}}
    \subfigure[DeepSMOTE]{\includegraphics[width=0.3\linewidth]{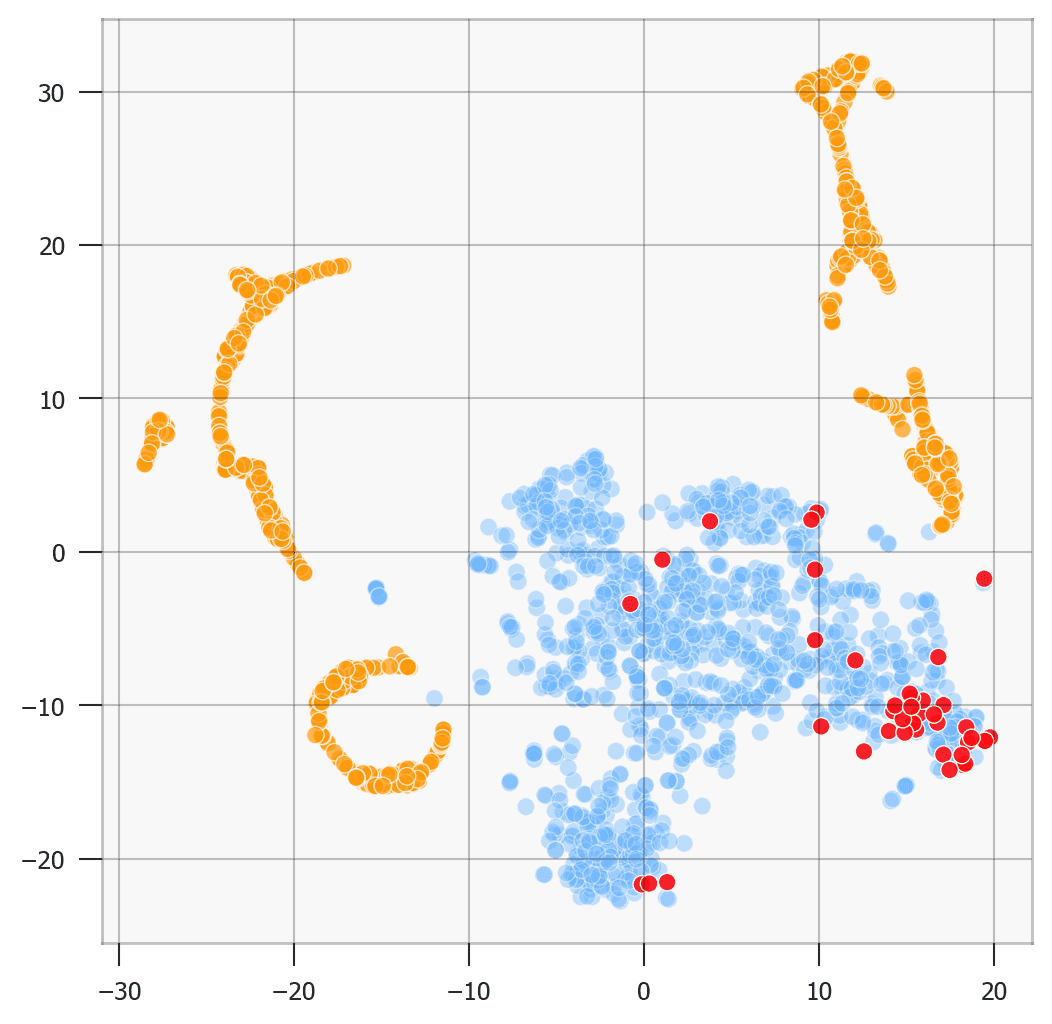}}
    \subfigure[DDHS]{\includegraphics[width=0.3\linewidth]{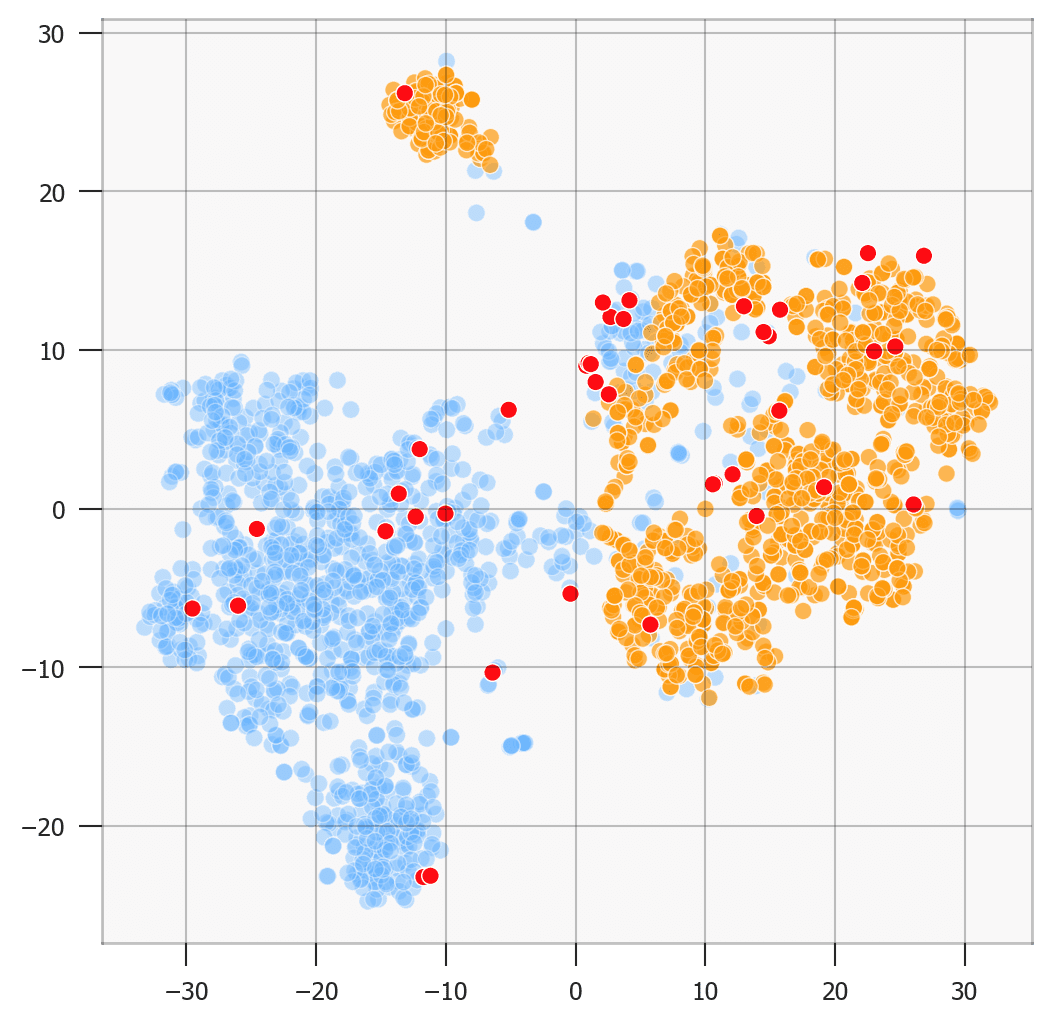}}
    \subfigure[SMOTE-CLS]{\includegraphics[width=0.3\linewidth]{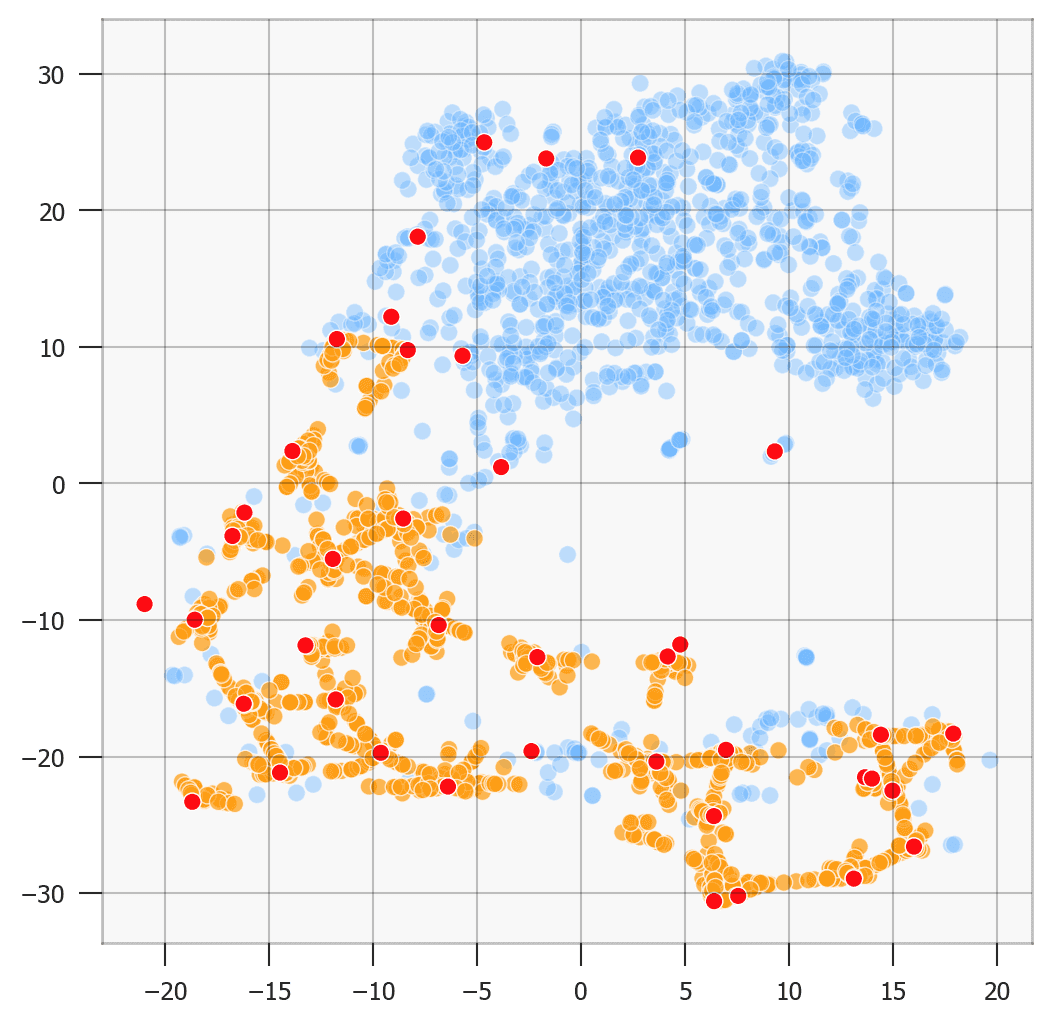}}
    \caption{t-SNE plots of the augmented $\texttt{yeast}$ dataset by oversampling methods with $\rho = 1.0$.}
    \label{fig:tsne}
\end{figure}

Figure \ref{fig:real_results} shows the latent spaces of DFBS, DeepSMOTE, DDHS, and SMOTE-CLS for four datasets: the first row for \texttt{ecoli}, the second row for \texttt{yeast}, the third row for \texttt{scene}, and the last row for \texttt{isolet}. The colors of the points represent the sample difficulty as determined by the KNN classifier $f_\kappa$. In low-dimensional datasets like \texttt{ecoli}, the latent space of all models shows distinctive features for each class. However, in the high-dimensional datasets (\texttt{scene} and \texttt{isolet}), latent variables of DFBS, DeepSMOTE, and DDHS are overlapped among the four classes. On the other hand,
SMOTE-CLS exhibits a relatively more distinct latent space with the augmented labels. It is worth noting that the latent visualizations provided by SMOTE-CLS effectively aid in noise identification, allowing us to determine the appropriate threshold $\tau$ for filtering it out.

Figure \ref{fig:tsne} displays the t-SNE plots \citep{van2008visualizing} of the augmented \texttt{yeast} dataset. Figure \ref{fig:tsne} (a) shows that SMOTE is prone to noise samples, which is undesirable. 
 Figure \ref{fig:tsne} (c) SMOTE-ENN seems to mitigate the problem of generating minority samples around noise samples. However, when a high oversampling ratio is used, such as $\rho = 1.0$ in our setup, a significant number of generated samples forms a cluster, and SMOTE-ENN fails to filter out noise samples effectively.  Consequently, SMOTE-ENN yields similar results to those of SMOTE. 
However, both BSMOTE (Figure \ref{fig:tsne} (b)), KMSMOTE (Figure \ref{fig:tsne} (d)), and SMOTE-CLS (Figure \ref{fig:tsne} (i)) filter them out in the candidates for oversampling. In addition, BSMOTE, while conservative, tends to overlook small disjuncts of the minor class within the major class. In contrast, our approach can identify these small disjuncts without clustering. Furthermore, KMSMOTE, which generates synthetic minority samples within clusters using SMOTE, has the potential drawback of reducing sample diversity and being sensitive to the results of the clustering process. 
In the context of deep learning approaches, specifically CVAE (Figure \ref{fig:tsne} (f)) and DeepSMOTE (Figure \ref{fig:tsne} (g)), which directly employ the decoder to generate minority samples, there is a tendency to produce low-quality samples due to their limited reconstruction ability. Regarding DFBS (Figure \ref{fig:tsne} (e)) and DDHS (Figure \ref{fig:tsne} (h)), which employ their generation techniques, they generate various minority samples. However, these samples often overlap with the major class regions, negatively impacting classification performance.

\begin{figure}[t!]
    \centering
    \subfigure[DFBS]{\includegraphics[width=0.32\linewidth]{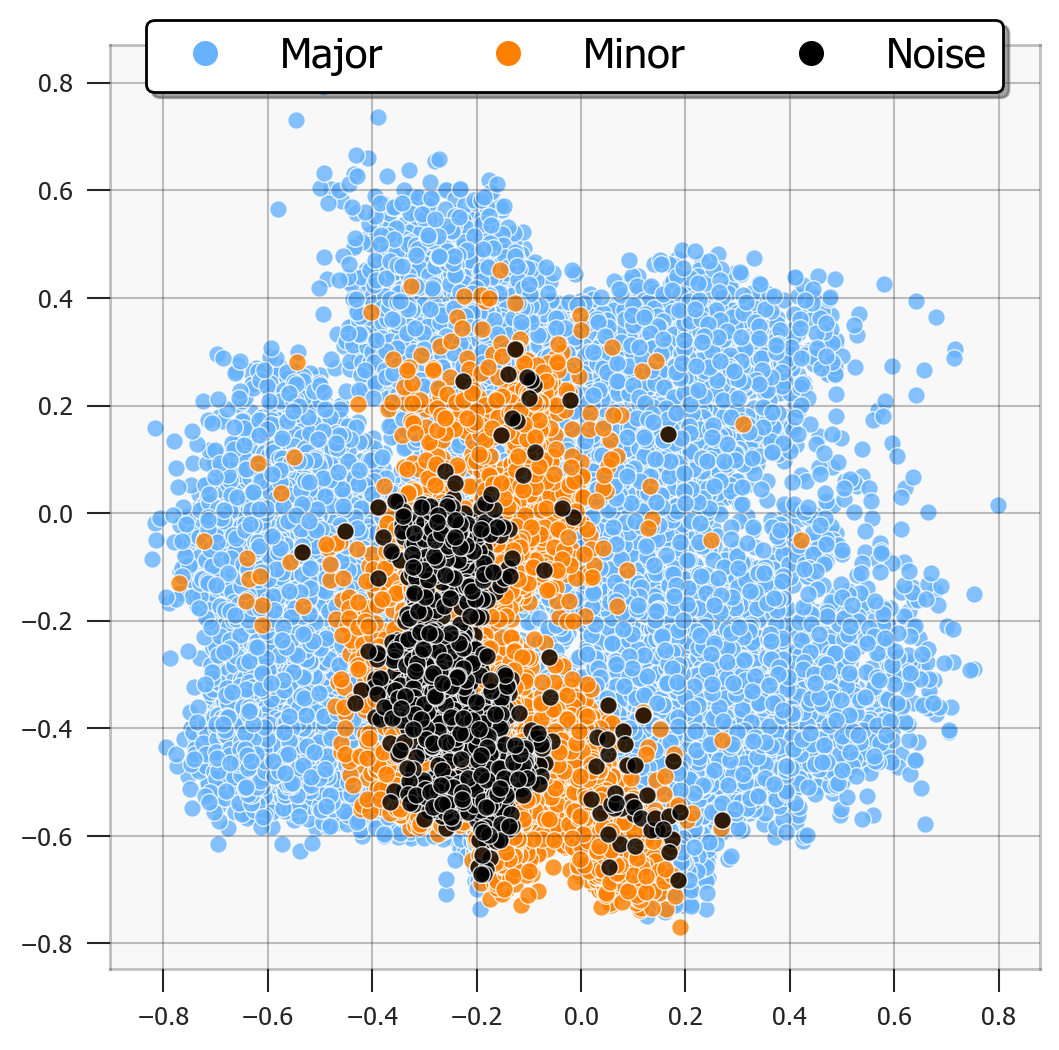}}
    \subfigure[DDHS]{\includegraphics[width=0.32\linewidth]{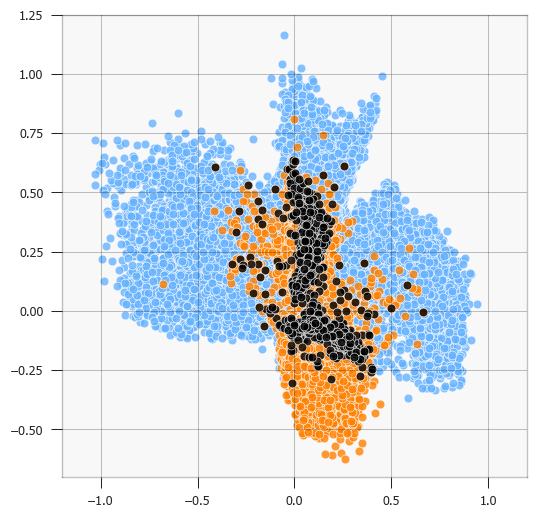}}
    \subfigure[SMOTE-CLS]{\includegraphics[width=0.32\linewidth]{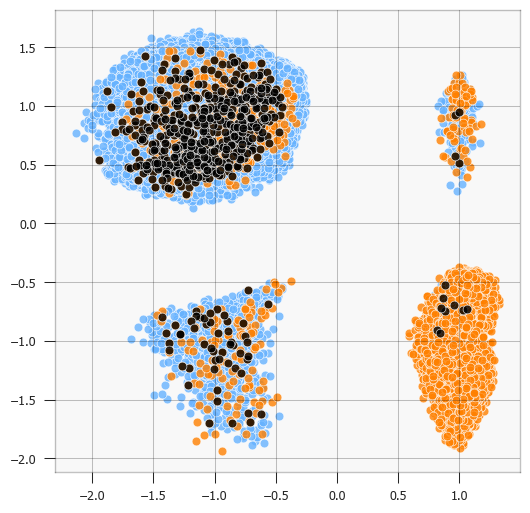}}
    \subfigure[DFBS]{\includegraphics[width=0.32\linewidth]{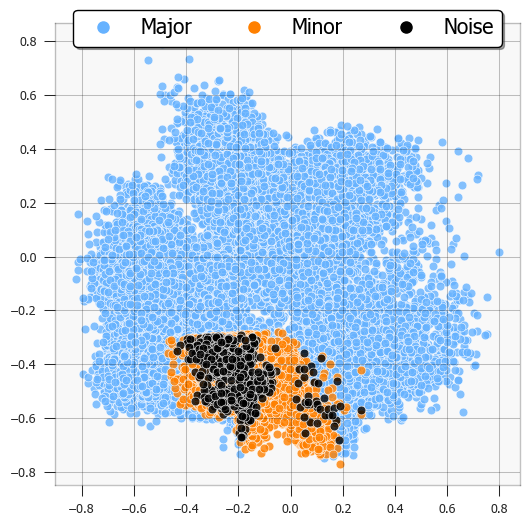}}
    \subfigure[DDHS]{\includegraphics[width=0.32\linewidth]{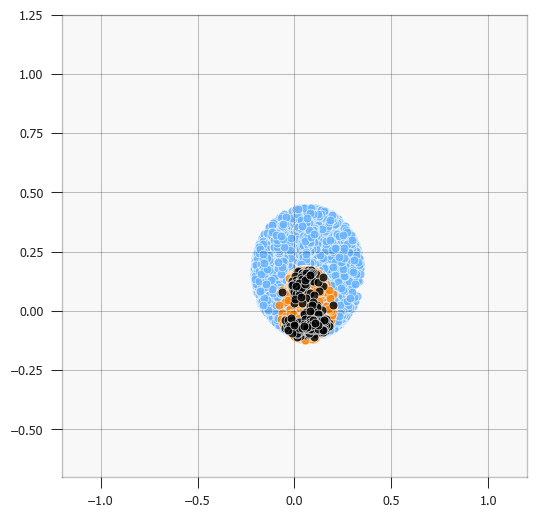}}
    \subfigure[SMOTE-CLS]{\includegraphics[width=0.32\linewidth]{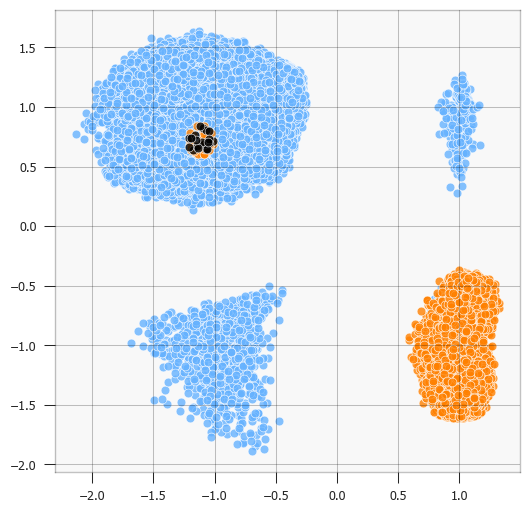}}
    \caption{Visualization of latent spaces for \texttt{MNIST} dataset with $\varrho = 0.1$. (a) - (c): the trained latent variables. (d) - (f): the selected latent variables by each filtering method.}
    \label{fig:filtering_results}
\end{figure}

\subsection{Robustness to Label Noise} \label{sec:noise}
In this section, we evaluate the filtering performance of our proposed model using the image dataset, \texttt{MNIST} \citep{lecun1998mnist}. 
Moreover, we explore the process of selecting optimal thresholds in KDE through K-fold cross-validation during the training phase. 
Our evaluation focuses on the effectiveness of our filtering algorithm in comparison with benchmark models that also include filtering mechanisms.
This analysis allows us to determine how well our model can identify and correct mislabeled data within the context of complex image datasets. 

\subsubsection{Experimental Design}
As distinguished from the benchmark datasets in Section \ref{sec: ra}, \texttt{MNIST} is characterized by multiple labels, e.g., $\mathcal{Y} = \{0, 1, \dots, 9 \}$. Therefore, to frame a problem with multiple labels as a binary classification problem, we divide these labels into two groups.
Inspired by prior work \citep{fajardo2021oversampling}, we define minor set $\{(\bx, m) | (\bx, y) \in \mathcal{X} \times \{3, 4\} \}$. 
We specifically select noisy samples originating from images originally labeled as `8' or `9', considering the visual similarities between digits such as `3' and `8' or `4' and `9', which make them more prone to mislabeling. 
The ratio of noisy samples to generate is $\varrho$, i.e., the number of noisy samples is $\varrho * |D_m|$.
The remaining samples are defined as the major set with the label $M$. This manipulation is validated based on the visual similarity between the chosen digits, which are more likely to be confused with each other than with other digits. This setup allows us to rigorously test the effectiveness of noise filtering in a controlled yet realistically challenging scenario. We conduct experiments across various noise ratios $\varrho \in \{0.05, 0.1, 0.2, 0.3\}$, counting the number of noisy samples identified by the filtering mechanism.

For more complex image datasets, the dimension of the hidden layers in the autoencoder architecture is specified in order as $(16h, 8h, 2h, h, 2h, 8h, 16h)$.
Additionally, our model incorporates a convolutional neural network (CNN), which is a standard baseline model for image classification, as the classifier $ f_\eta$. This seamless classifier integration allows our model to handle a variety of datasets across different domains. The experiment is conducted ten times with different seed numbers, and the mean and standard errors of the evaluation metrics are reported. This experiment employs a simple MLP classifier with one hidden layer of 64 dimensions as a downstream classifier, as in \cite{fajardo2021oversampling}. 

\subsubsection{Selecting the threshold in the filtering process via K-fold cross-validation}
In our framework, a key aspect involves the selection of optimal thresholds ($\tau$s) in the filtering algorithm. These thresholds play a crucial role in determining the effectiveness of the noise filtering process. During the training phase, we can determine the optimal thresholds by employing K-fold cross-validation \citep{rodriguez2009sensitivity}. 
The process begins by applying our filtering algorithms to the training dataset to filter out noise samples. Following this, we calculate the 5-fold cross-validation AUPRC and AUC metrics to evaluate performance. We specifically consider the thresholds $\tau_1$ and $\tau_2$ as quantiles, setting them at levels within the sets $\{ 0.1, 0.2, 0.3 \}$ and $\{ 0.65, 0.8, 0.95 \}$, respectively, to fine-tune the filtering process. For instance, setting $\tau_2$ a quantile at the level $0.8$ implies that only the top 20\% of samples in the hard-minor class, based on their density scores, are selected as candidates for oversampling.

Table \ref{tab:taus} presents the 5-fold cross-validation performances across various threshold pairs. As shown in Table \ref{tab:taus}, selecting samples with higher densities in the hard-minority class -- such as setting $\tau_2$ as quantile at level $0.95$ -- results in more accurate classification performance in the 5-fold cross-validation.
These outcomes align with our initial conjectures that the hard-minority class contains a significant number of noisy samples, which contribute to poor oversampling performance. As $\rho$ increases, the noisy samples begin to include the easy-minor class, prompting an increase in the optimal $\tau_1$ threshold. We identify the optimal threshold pairs as those yielding the highest AUPRC during cross-validation. In the next section, we will use these optimal thresholds to compare our method with other techniques on the test set.

\begin{table}[t]
\caption{5-fold cross-validation performance of SMOTE-CLS with various $\tau_1$ and $\tau_2$ on the \texttt{MNIST} datasets. The most favorable value is bolded in each $\rho$.}
\centering
    \begin{tabular}{llccccccc}
    \toprule
     &  & & \multicolumn{3}{c}{AUPRC} & \multicolumn{3}{c}{AUC} \\ 
    \cmidrule(lr){4-6} \cmidrule(lr){7-9}
    $\rho$ & $\tau_1$ & $\tau_2$ & 0.65 & 0.8 & 0.95 & 0.65 & 0.8 & 0.95 \\ 
    \midrule
    \multirow{3}{*}{0.05} & 0.1 & & $0.0.819_{\pm 0.013}$ & $0.831_{\pm 0.018}$ & $\mathbf{0.841_{\pm 0.020}}$ & $0.852_{\pm 0.008}$ & $0.859_{\pm 0.014}$ & $\mathbf{0.875_{\pm 0.011}}$ \\
    & 0.2 & & $0.802_{\pm 0.049}$ & $0.816_{\pm 0.042}$ & $0.831_{\pm 0.022}$ & $0.836_{\pm 0.034}$ & $0.844_{\pm 0.029}$ & $0.858_{\pm 0.014}$\\ 
    & 0.3 & & $0.796_{\pm 0.027}$ & $0.804_{\pm 0.030}$ & $0.770_{\pm 0.053}$ & $0.835_{\pm 0.019}$ & $0.844_{\pm 0.020}$ & $0.815_{\pm 0.038}$\\
    \midrule
    \multirow{3}{*}{0.1} & 0.1 & & $0.790_{\pm 0.016}$ & $0.784_{\pm 0.035}$ & $0.817_{\pm 0.028}$ & $0.829_{\pm 0.007}$ & $0.830_{\pm 0.022}$ & $0.848_{\pm 0.019}$ \\
    & 0.2 & & $0.749_{\pm 0.038}$ & $0.716_{\pm 0.040}$ & $0.742_{\pm 0.036}$ & $0.798_{\pm 0.027}$ & $0.778_{\pm 0.030}$ & $0.791_{\pm 0.027}$ \\
    & 0.3 & & $0.817_{\pm 0.013}$ & $0.816_{\pm 0.017}$ & $\mathbf{0.844_{\pm 0.011}}$ & $0.844_{\pm 0.009}$ & $0.849_{\pm 0.013}$ & $\mathbf{0.867_{\pm 0.007}}$ \\
    \midrule
    \multirow{3}{*}{0.2} & 0.1 & & $0.786_{\pm 0.026}$ & $0.831_{\pm 0.019}$ & $\mathbf{0.840_{\pm 0.013}}$ & $0.824_{\pm 0.018}$ & $0.856_{\pm 0.015}$ & $0.858_{\pm 0.012}$ \\
    & 0.2 & & $0.717_{\pm 0.020}$ & $0.764_{\pm 0.024}$ & $0.777_{\pm 0.025}$ & $0.780_{\pm 0.015}$ & $0.812_{\pm 0.015}$ & $0.825_{\pm 0.016}$ \\
    & 0.3 & & $0.794_{\pm 0.035}$ & $0.780_{\pm 0.022}$ & $0.831_{\pm 0.028}$ & $0.827_{\pm 0.028}$ & $0.824_{\pm 0.016}$ & $\mathbf{0.864_{\pm 0.017}}$ \\
    \midrule
    \multirow{3}{*}{0.3} & 0.1 & & $0.768_{\pm 0.041}$ & $0.759_{\pm 0.027}$ & $0.716_{\pm 0.059}$ & $0.817_{\pm 0.024}$ & $0.803_{\pm 0.016}$ & $0.775_{\pm 0.040}$ \\
    & 0.2 & & $0.640_{\pm 0.035}$ & $0.690_{\pm 0.041}$ & $\mathbf{0.777_{\pm 0.045}}$ & $0.722_{\pm 0.026}$ & $0.764_{\pm 0.028}$ & $\mathbf{0.823_{\pm 0.031}}$ \\
    & 0.3 & & $0.715_{\pm 0.045}$ & $0.720_{\pm 0.050}$ & $0.745_{\pm 0.051}$ & $0.776_{\pm 0.029}$ & $0.784_{\pm 0.034}$ & $0.800_{\pm 0.034}$ \\
    \bottomrule
    \end{tabular}
\label{tab:taus}
\end{table}

\subsubsection{Filtering Results}
To assess the effectiveness of noise filtering and its impact on downstream task performance, we calculate classification metrics: recall and F1-score (the harmonic mean of precision and recall) for noise detection and AUPRC for the downstream task.

\begin{sidewaystable}[ph!]
\caption{Noise filtering and downstream classification performance of DDFS, DDHS, and SMOTE-CLS using the MLP classifier on the \texttt{MNIST} datasets. The most favorable value is bolded.}
  \centering
  \begin{tabular}{lccccccccc}
    \toprule
    Task & \multicolumn{6}{c}{Noise filtering} & \multicolumn{3}{c}{Downstream task} \\ 
    Metric & \multicolumn{3}{c}{Recall} & \multicolumn{3}{c}{F1} & \multicolumn{3}{c}{AUPRC} \\
     \cmidrule(lr){2-4} \cmidrule(lr){5-7} \cmidrule(lr){8-10}
    $\varrho$ & DFBS & DDHS & SMOTE-CLS & DFBS & DDHS & SMOTE-CLS & DFBS & DDHS & SMOTE-CLS \\
    \midrule
    0.05 & $0.623_{\pm 0.035}$ & $0.683_{\pm 0.024}$ & $\mathbf{0.927_{\pm 0.002}}$ & $0.116_{\pm 0.004}$ & $0.080_{\pm 0.002}$ & $\mathbf{0.427_{\pm 0.001}}$ & $0.991_{\pm0.000}$ & $\mathbf{0.992_{\pm0.000}}$ & $0.991_{\pm0.000}$ \\
    0.1 & $0.590_{\pm 0.035}$ & $0.686_{\pm 0.024}$ & $\mathbf{0.933_{\pm 0.002}}$ & $0.194_{\pm 0.008}$ & $0.146_{\pm 0.004}$ & $\mathbf{0.603_{\pm0.001}}$ & $0.989_{\pm0.000}$ & $\mathbf{0.990_{\pm0.000}}$ & $0.989_{\pm0.000}$ \\
    0.2 & $0.606_{\pm 0.027}$ & $0.697_{\pm 0.019}$ & $\mathbf{0.930_{\pm 0.003}}$ & $0.336_{\pm 0.011}$ & $0.250_{\pm 0.005}$ & $\mathbf{0.783_{\pm0.002}}$ & $0.987_{\pm0.000}$ & $0.988_{\pm0.000}$ & $\mathbf{0.990_{\pm0.000}}$ \\
    0.3 & $0.530_{\pm 0.032}$ & $0.719_{\pm 0.016}$ & $\mathbf{0.868_{\pm 0.004}}$ & $0.358_{\pm 0.017}$ & $0.336_{\pm 0.005}$ & $\mathbf{0.850_{\pm 0.002}}$ & $0.985_{\pm 0.000}$ & $0.985_{\pm 0.001}$ & $\mathbf{0.988_{\pm 0.000}}$ \\
    \bottomrule
  \end{tabular}
\label{tab:noise}
\end{sidewaystable}

Figure \ref{fig:filtering_results} showcases the latent spaces generated by DFBS, DDHS, and SMOTE-CLS for the \texttt{MNIST} dataset, specifically adapted for this experiment with a noise ratio $\varrho = 0.1$. The first row of Figure \ref{fig:filtering_results} visualizes the latent variables ($\bz$) produced by the three oversampling methods. As shown in Figures \ref{fig:filtering_results} (a) and (b), the latent variables of the minor and noise groups are not clearly separated in the cases of DFBS and DDHS, indicating a lack of effective disentanglement. In contrast, SMOTE-CLS demonstrates a more distinct separation between noise and minor groups, as seen in Figure \ref{fig:filtering_results} (c). After each method learns the latent space, the second row of Figure \ref{fig:filtering_results} shows the latent variables after each filtering algorithm. Figures \ref{fig:filtering_results} (d) and (e) reveal that both DDHS and DFBS are overly conservative in selecting minor samples and fail to adequately detect and filter out the noise. On the other hand, Figure \ref{fig:filtering_results} (f) shows that SMOTE-CLS not only selects minor samples effectively but also successfully filters out noise, demonstrating its robustness in noise management and enhancement of data quality for downstream tasks.

Table \ref{tab:noise} presents the noise filtering and downstream task performance of DFBS, DDHS, and our proposed method, SMOTE-CLS. The filtering algorithms of DFBS and DDHS are elaborated in equations \eqref{eq:dfbs_filtering} and \eqref{eq:ddhs_filtering}, respectively (for more detailed descriptions, refer to Section \ref{sec:simul}). Through Table \ref{tab:noise}, it is clear that SMOTE-CLS significantly outperforms the other methods, achieving higher mean values and lower standard errors in noise filtering tasks. Notably, in the recall metric, SMOTE-CLS consistently detects noise with high accuracy across different noise ratios, $\varrho$. This indicates that the latent space learned by our proposed method has a superior ability to distinguish true noise compared to other distance or density-based approaches. 
In terms of the F1 metric, which balances recall and precision, SMOTE-CLS demonstrates an average improvement of 290\% over DFBS and 340\% over DDHS. Additionally, in downstream tasks, the MLP classifier utilizing synthetic samples generated by SMOTE-CLS outperforms the others as the noise ratio increases, highlighting its robustness and versatility in handling real-world data challenges. This suggests that our difficulty-adaptive filtering algorithm is more effective at selectively identifying and removing noisy samples than other approaches.

\section{Conclusions and Limitations} \label{sec:cl}
In this study, we proposed a novel oversampling algorithm, SMOTE-CLS, which exploits the strengths of  SMOTE and VAE. Our algorithm comprises four key steps: classifying sample difficulty, training VAE with a tailored latent space, applying density-based filtering, and oversampling using SMOTE. Through numerical studies with the synthetic data, we demonstrated the necessity of each step in improving the SMOTE algorithm and superiority over the existing methods. In real data analysis, SMOTE-CLS significantly improved classification performance for the minor class, as measured by the AUPRC. Moreover, the visualization results provided empirical evidence of SMOTE-CLS's ability to generate high-quality minority samples and its robustness to noise. Additionally, we anticipate that our filtering algorithm can seamlessly integrate with other state-of-the-art oversampling methods, further enhancing its versatility and applicability.

One intriguing finding was elucidated through the current investigation: the substantial impact of data generation capability via neural network models on enhancing the predictive performance of minor class augmentation. Specifically, the augmented minority samples, while effectively distinguished from majority samples, show distinct characteristics compared to the original minor class data and occasionally degrade the classification models' predictive efficacy. Conversely, by augmenting samples within the data space, SMOTE ensured that the augmented samples have features more akin to the observed minority samples. In essence, for minor class data augmentation, the quantity of data pertinent to the minor class, rather than that of the whole dataset, proved crucial. Especially for sparse data, augmenting directly within the data space, as facilitated by techniques like SMOTE, which define and augment neighbors, exhibited superior augmentation performance over neural network-based generative models that struggle with data reproducibility.

Even though SMOTE-CLS showed superior performance compared to other oversampling methods, it still has some limitations. One notable limitation is its applicability to multi-class tasks, as customizing the latent space based on the confusion matrix can be challenging in such scenarios. While one potential solution could involve utilizing a higher-dimensional latent space, it is essential to consider that this approach may present challenges to the effectiveness of the density-based filtering algorithm, particularly in higher-dimensional settings.
Some possible solutions extending the applicability of our approach include leveraging prior knowledge of label relationships and developing methods to map such relationships to the latent space. One approach could involve learning the prior distribution rather than fixing it to maximize the distance among classes, as proposed by \cite{hajimiri2021semi}. We leave them as our future works.

\bibliography{ref}

\begin{thebibliography}{49}
\expandafter\ifx\csname natexlab\endcsname\relax\def\natexlab#1{#1}\fi
\expandafter\ifx\csname url\endcsname\relax
  \def\url#1{\texttt{#1}}\fi
\expandafter\ifx\csname urlprefix\endcsname\relax\def\urlprefix{}\fi

\bibitem[{Achlioptas et~al.(2018)Achlioptas, Diamanti, Mitliagkas and Guibas}]{pmlr-v80-achlioptas18a}
\textsc{Achlioptas, P.}, \textsc{Diamanti, O.}, \textsc{Mitliagkas, I.} and \textsc{Guibas, L.} (2018).
\newblock Learning representations and generative models for 3d point clouds.
\newblock In \textit{International Conference on Machine Learning}.

\bibitem[{An and Jeon(2024)}]{an2024customization}
\textsc{An, S.} and \textsc{Jeon, J.-J.} (2024).
\newblock Customization of latent space in semi-supervised variational autoencoder.
\newblock \textit{Pattern Recognition Letters}, \textbf{177} 54--60.

\bibitem[{Azhar et~al.(2022)Azhar, Pozi, Din and Jatowt}]{azhar2022investigation}
\textsc{Azhar, N.~A.}, \textsc{Pozi, M. S.~M.}, \textsc{Din, A.~M.} and \textsc{Jatowt, A.} (2022).
\newblock An investigation of smote based methods for imbalanced datasets with data complexity analysis.
\newblock \textit{IEEE Transactions on Knowledge and Data Engineering}.

\bibitem[{Batista et~al.(2004)Batista, Prati and Monard}]{batista2004study}
\textsc{Batista, G.~E.}, \textsc{Prati, R.~C.} and \textsc{Monard, M.~C.} (2004).
\newblock A study of the behavior of several methods for balancing machine learning training data.
\newblock \textit{ACM SIGKDD explorations newsletter}, \textbf{6} 20--29.

\bibitem[{Breiman(2001)}]{breiman2001random}
\textsc{Breiman, L.} (2001).
\newblock Random forests.
\newblock \textit{Machine learning}, \textbf{45} 5--32.

\bibitem[{Camino and Hammerschmidt(2020)}]{Camino2020OversamplingTD}
\textsc{Camino, R.~D.} and \textsc{Hammerschmidt, C.~A.} (2020).
\newblock Oversampling tabular data with deep generative models: Is it worth the effort?
\newblock In \textit{ICBINB@NeurIPS}.

\bibitem[{Chang and Lin(2011)}]{chang2011libsvm}
\textsc{Chang, C.-C.} and \textsc{Lin, C.-J.} (2011).
\newblock Libsvm: a library for support vector machines.
\newblock \textit{ACM transactions on intelligent systems and technology (TIST)}, \textbf{2} 1--27.

\bibitem[{Chawla et~al.(2002)Chawla, Bowyer, Hall and Kegelmeyer}]{chawla2002smote}
\textsc{Chawla, N.~V.}, \textsc{Bowyer, K.~W.}, \textsc{Hall, L.~O.} and \textsc{Kegelmeyer, W.~P.} (2002).
\newblock Smote: synthetic minority over-sampling technique.
\newblock \textit{Journal of artificial intelligence research}, \textbf{16} 321--357.

\bibitem[{Chen and Guestrin(2016)}]{Chen2016XGBoostAS}
\textsc{Chen, T.} and \textsc{Guestrin, C.} (2016).
\newblock Xgboost: A scalable tree boosting system.
\newblock \textit{Proceedings of the 22nd ACM SIGKDD International Conference on Knowledge Discovery and Data Mining}.

\bibitem[{Dablain et~al.(2023{\natexlab{a}})Dablain, Krawczyk and Chawla}]{Dablain2021DeepSMOTEFD}
\textsc{Dablain, D.}, \textsc{Krawczyk, B.} and \textsc{Chawla, N.~V.} (2023{\natexlab{a}}).
\newblock Deepsmote: Fusing deep learning and smote for imbalanced data.
\newblock \textit{IEEE Transactions on Neural Networks and Learning Systems}, \textbf{34} 6390--6404.

\bibitem[{Dablain et~al.(2023{\natexlab{b}})Dablain, Bellinger, Krawczyk and Chawla}]{dablain2023efficient}
\textsc{Dablain, D.~A.}, \textsc{Bellinger, C.}, \textsc{Krawczyk, B.} and \textsc{Chawla, N.~V.} (2023{\natexlab{b}}).
\newblock Efficient augmentation for imbalanced deep learning.
\newblock In \textit{2023 IEEE 39th International Conference on Data Engineering (ICDE)}. IEEE.

\bibitem[{Dai and Wipf(2019)}]{dai2018diagnosing}
\textsc{Dai, B.} and \textsc{Wipf, D.} (2019).
\newblock Diagnosing and enhancing {VAE} models.
\newblock In \textit{International Conference on Learning Representations}.

\bibitem[{Dai et~al.(2019)Dai, Ng, Severson, Huang, Anderson and Stultz}]{Dai2019GenerativeOW}
\textsc{Dai, W.}, \textsc{Ng, K.}, \textsc{Severson, K.~A.}, \textsc{Huang, W.}, \textsc{Anderson, F.} and \textsc{Stultz, C.~M.} (2019).
\newblock Generative oversampling with a contrastive variational autoencoder.
\newblock \textit{2019 IEEE International Conference on Data Mining (ICDM)} 101--109.

\bibitem[{Davis and Goadrich(2006)}]{davis2006relationship}
\textsc{Davis, J.} and \textsc{Goadrich, M.} (2006).
\newblock The relationship between precision-recall and roc curves.
\newblock In \textit{Proceedings of the 23rd international conference on Machine learning}.

\bibitem[{Douzas et~al.(2018)Douzas, Bacao and Last}]{douzas2018improving}
\textsc{Douzas, G.}, \textsc{Bacao, F.} and \textsc{Last, F.} (2018).
\newblock Improving imbalanced learning through a heuristic oversampling method based on k-means and smote.
\newblock \textit{Information Sciences}, \textbf{465} 1--20.

\bibitem[{Elreedy et~al.(2023)Elreedy, Atiya and Kamalov}]{elreedy2023theoretical}
\textsc{Elreedy, D.}, \textsc{Atiya, A.~F.} and \textsc{Kamalov, F.} (2023).
\newblock A theoretical distribution analysis of synthetic minority oversampling technique (smote) for imbalanced learning.
\newblock \textit{Machine Learning} 1--21.

\bibitem[{Fajardo et~al.(2021)Fajardo, Findlay, Jaiswal, Yin, Houmanfar, Xie, Liang, She and Emerson}]{fajardo2021oversampling}
\textsc{Fajardo, V.~A.}, \textsc{Findlay, D.}, \textsc{Jaiswal, C.}, \textsc{Yin, X.}, \textsc{Houmanfar, R.}, \textsc{Xie, H.}, \textsc{Liang, J.}, \textsc{She, X.} and \textsc{Emerson, D.} (2021).
\newblock On oversampling imbalanced data with deep conditional generative models.
\newblock \textit{Expert Systems with Applications}, \textbf{169} 114463.

\bibitem[{Fern{\'a}ndez et~al.(2018)Fern{\'a}ndez, Garcia, Herrera and Chawla}]{fernandez2018smote}
\textsc{Fern{\'a}ndez, A.}, \textsc{Garcia, S.}, \textsc{Herrera, F.} and \textsc{Chawla, N.~V.} (2018).
\newblock Smote for learning from imbalanced data: progress and challenges, marking the 15-year anniversary.
\newblock \textit{Journal of artificial intelligence research}, \textbf{61} 863--905.

\bibitem[{Ghosh and Chaudhuri(2004)}]{ghosh2004optimal}
\textsc{Ghosh, A.~K.} and \textsc{Chaudhuri, P.} (2004).
\newblock Optimal smoothing in kernel discriminant analysis.
\newblock \textit{Statistica Sinica} 457--483.

\bibitem[{Goodfellow et~al.(2014)Goodfellow, Pouget-Abadie, Mirza, Xu, Warde-Farley, Ozair, Courville and Bengio}]{goodfellow2014generative}
\textsc{Goodfellow, I.}, \textsc{Pouget-Abadie, J.}, \textsc{Mirza, M.}, \textsc{Xu, B.}, \textsc{Warde-Farley, D.}, \textsc{Ozair, S.}, \textsc{Courville, A.} and \textsc{Bengio, Y.} (2014).
\newblock Generative adversarial nets.
\newblock \textit{Advances in neural information processing systems}, \textbf{27}.

\bibitem[{Hajimiri et~al.(2021)Hajimiri, Lotfi and Baghshah}]{hajimiri2021semi}
\textsc{Hajimiri, S.}, \textsc{Lotfi, A.} and \textsc{Baghshah, M.~S.} (2021).
\newblock Semi-supervised disentanglement of class-related and class-independent factors in vae.
\newblock \textit{arXiv preprint arXiv:2102.00892}.

\bibitem[{Han et~al.(2005)Han, Wang and Mao}]{Han2005BorderlineSMOTEAN}
\textsc{Han, H.}, \textsc{Wang, W.} and \textsc{Mao, B.} (2005).
\newblock Borderline-smote: A new over-sampling method in imbalanced data sets learning.
\newblock In \textit{International Conference on Intelligent Computing}.

\bibitem[{Hancock et~al.(2023)Hancock, Khoshgoftaar and Johnson}]{hancock2023evaluating}
\textsc{Hancock, J.~T.}, \textsc{Khoshgoftaar, T.~M.} and \textsc{Johnson, J.~M.} (2023).
\newblock Evaluating classifier performance with highly imbalanced big data.
\newblock \textit{Journal of Big Data}, \textbf{10} 42.

\bibitem[{He et~al.(2008)He, Bai, Garcia and Li}]{he2008adasyn}
\textsc{He, H.}, \textsc{Bai, Y.}, \textsc{Garcia, E.~A.} and \textsc{Li, S.} (2008).
\newblock Adasyn: Adaptive synthetic sampling approach for imbalanced learning.
\newblock In \textit{2008 IEEE international joint conference on neural networks (IEEE world congress on computational intelligence)}. IEEE.

\bibitem[{He and Garcia(2009)}]{he2009learning}
\textsc{He, H.} and \textsc{Garcia, E.~A.} (2009).
\newblock Learning from imbalanced data.
\newblock \textit{IEEE Transactions on knowledge and data engineering}, \textbf{21} 1263--1284.

\bibitem[{Idakwo et~al.(2020)Idakwo, Thangapandian, Luttrell, Li, Wang, Zhou, Hong, Yang, Zhang and Gong}]{idakwo2020structure}
\textsc{Idakwo, G.}, \textsc{Thangapandian, S.}, \textsc{Luttrell, J.}, \textsc{Li, Y.}, \textsc{Wang, N.}, \textsc{Zhou, Z.}, \textsc{Hong, H.}, \textsc{Yang, B.}, \textsc{Zhang, C.} and \textsc{Gong, P.} (2020).
\newblock Structure--activity relationship-based chemical classification of highly imbalanced tox21 datasets.
\newblock \textit{Journal of cheminformatics}, \textbf{12} 1--19.

\bibitem[{Kamalov(2020)}]{kamalov2020kernel}
\textsc{Kamalov, F.} (2020).
\newblock Kernel density estimation based sampling for imbalanced class distribution.
\newblock \textit{Information Sciences}, \textbf{512} 1192--1201.

\bibitem[{Kingma and Welling(2014)}]{kingma2013auto}
\textsc{Kingma, D.~P.} and \textsc{Welling, M.} (2014).
\newblock Auto-encoding variational bayes.
\newblock In \textit{International Conference on Learning Representations}.

\bibitem[{LeCun(1998)}]{lecun1998mnist}
\textsc{LeCun, Y.} (1998).
\newblock The mnist database of handwritten digits.
\newblock {URL}: http://yann.lecun.com/exdb/mnist.

\bibitem[{Lema{\^\i}tre et~al.(2017)Lema{\^\i}tre, Nogueira and Aridas}]{lemaitre2017imbalanced}
\textsc{Lema{\^\i}tre, G.}, \textsc{Nogueira, F.} and \textsc{Aridas, C.~K.} (2017).
\newblock Imbalanced-learn: A python toolbox to tackle the curse of imbalanced datasets in machine learning.
\newblock \textit{The Journal of Machine Learning Research}, \textbf{18} 559--563.

\bibitem[{Liu and Chang(2022)}]{Liu2022LearningFI}
\textsc{Liu, C.-L.} and \textsc{Chang, Y.} (2022).
\newblock Learning from imbalanced data with deep density hybrid sampling.
\newblock \textit{IEEE Transactions on Systems, Man, and Cybernetics: Systems}, \textbf{52} 7065--7077.

\bibitem[{Liu(2023)}]{liu2023novel}
\textsc{Liu, R.} (2023).
\newblock A novel synthetic minority oversampling technique based on relative and absolute densities for imbalanced classification.
\newblock \textit{Applied Intelligence}, \textbf{53} 786--803.

\bibitem[{Liu et~al.(2023)Liu, Qian, Berrevoets and van~der Schaar}]{liu2023goggle}
\textsc{Liu, T.}, \textsc{Qian, Z.}, \textsc{Berrevoets, J.} and \textsc{van~der Schaar, M.} (2023).
\newblock {GOGGLE}: Generative modelling for tabular data by learning relational structure.
\newblock In \textit{The Eleventh International Conference on Learning Representations}.

\bibitem[{Liu et~al.(2018)Liu, Liu and Tseng}]{Liu2018DeepDF}
\textsc{Liu, Y.}, \textsc{Liu, C.-L.} and \textsc{Tseng, V.~S.} (2018).
\newblock Deep discriminative features learning and sampling for imbalanced data problem.
\newblock \textit{2018 IEEE International Conference on Data Mining (ICDM)} 1146--1151.

\bibitem[{Malhotra and Kamal(2019)}]{malhotra2019empirical}
\textsc{Malhotra, R.} and \textsc{Kamal, S.} (2019).
\newblock An empirical study to investigate oversampling methods for improving software defect prediction using imbalanced data.
\newblock \textit{Neurocomputing}, \textbf{343} 120--140.

\bibitem[{Markelle~Kelly(2023)}]{uci}
\textsc{Markelle~Kelly, K.~N., Rachel~Longjohn} (2023).
\newblock The uci machine learning repository.

\bibitem[{Mazurowski et~al.(2008)Mazurowski, Habas, Zurada, Lo, Baker and Tourassi}]{mazurowski2008training}
\textsc{Mazurowski, M.~A.}, \textsc{Habas, P.~A.}, \textsc{Zurada, J.~M.}, \textsc{Lo, J.~Y.}, \textsc{Baker, J.~A.} and \textsc{Tourassi, G.~D.} (2008).
\newblock Training neural network classifiers for medical decision making: The effects of imbalanced datasets on classification performance.
\newblock \textit{Neural networks}, \textbf{21} 427--436.

\bibitem[{Nakai(1996)}]{misc_ecoli_39}
\textsc{Nakai, K.} (1996).
\newblock {Ecoli}.
\newblock UCI Machine Learning Repository.
\newblock {URL}: https://doi.org/10.24432/C5388M.

\bibitem[{Rodriguez et~al.(2009)Rodriguez, Perez and Lozano}]{rodriguez2009sensitivity}
\textsc{Rodriguez, J.~D.}, \textsc{Perez, A.} and \textsc{Lozano, J.~A.} (2009).
\newblock Sensitivity analysis of k-fold cross validation in prediction error estimation.
\newblock \textit{IEEE transactions on pattern analysis and machine intelligence}, \textbf{32} 569--575.

\bibitem[{Scott(1979)}]{Scott1979OnOA}
\textsc{Scott, D.~W.} (1979).
\newblock On optimal and data based histograms.
\newblock \textit{Biometrika}, \textbf{66} 605--610.

\bibitem[{Shin and Kang(2022)}]{shin2022adanoise}
\textsc{Shin, K.} and \textsc{Kang, S.} (2022).
\newblock Adanoise: Training neural networks with adaptive noise for imbalanced data classification.
\newblock \textit{Expert Systems with Applications}, \textbf{192} 116364.

\bibitem[{Sohn et~al.(2015)Sohn, Lee and Yan}]{sohn2015learning}
\textsc{Sohn, K.}, \textsc{Lee, H.} and \textsc{Yan, X.} (2015).
\newblock Learning structured output representation using deep conditional generative models.
\newblock \textit{Advances in neural information processing systems}, \textbf{28}.

\bibitem[{Solomon et~al.(2022)Solomon, Jayavelu, Ferdaus and Kumar}]{solomon2022data}
\textsc{Solomon, I.}, \textsc{Jayavelu, S.}, \textsc{Ferdaus, M.~M.} and \textsc{Kumar, U.} (2022).
\newblock Data oversampling with structure preserving variational learning.
\newblock In \textit{Proceedings of the 31st ACM International Conference on Information \& Knowledge Management}.

\bibitem[{Sorscher et~al.(2022)Sorscher, Geirhos, Shekhar, Ganguli and Morcos}]{sorscher2022beyond}
\textsc{Sorscher, B.}, \textsc{Geirhos, R.}, \textsc{Shekhar, S.}, \textsc{Ganguli, S.} and \textsc{Morcos, A.} (2022).
\newblock Beyond neural scaling laws: beating power law scaling via data pruning.
\newblock \textit{Advances in Neural Information Processing Systems}, \textbf{35} 19523--19536.

\bibitem[{Tian et~al.(2021)Tian, Chen, Zhang, Feng, Xiong, Wu and Dou}]{tian2021re}
\textsc{Tian, J.}, \textsc{Chen, S.}, \textsc{Zhang, X.}, \textsc{Feng, Z.}, \textsc{Xiong, D.}, \textsc{Wu, S.} and \textsc{Dou, C.} (2021).
\newblock Re-embedding difficult samples via mutual information constrained semantically oversampling for imbalanced text classification.
\newblock In \textit{Proceedings of the 2021 Conference on Empirical Methods in Natural Language Processing}.

\bibitem[{Van~der Maaten and Hinton(2008)}]{van2008visualizing}
\textsc{Van~der Maaten, L.} and \textsc{Hinton, G.} (2008).
\newblock Visualizing data using t-sne.
\newblock \textit{Journal of machine learning research}, \textbf{9}.

\bibitem[{Van~Kerm(2003)}]{van2003adaptive}
\textsc{Van~Kerm, P.} (2003).
\newblock Adaptive kernel density estimation.
\newblock \textit{The Stata Journal}, \textbf{3} 148--156.

\bibitem[{Wen et~al.(2016)Wen, Zhang, Li and Qiao}]{wen2016discriminative}
\textsc{Wen, Y.}, \textsc{Zhang, K.}, \textsc{Li, Z.} and \textsc{Qiao, Y.} (2016).
\newblock A discriminative feature learning approach for deep face recognition.
\newblock In \textit{Computer Vision--ECCV 2016: 14th European Conference, Amsterdam, The Netherlands, October 11--14, 2016, Proceedings, Part VII 14}. Springer.

\bibitem[{Xie et~al.(2020)Xie, Qiu, Zhang, Peng and Chen}]{xie2020gaussian}
\textsc{Xie, Y.}, \textsc{Qiu, M.}, \textsc{Zhang, H.}, \textsc{Peng, L.} and \textsc{Chen, Z.} (2020).
\newblock Gaussian distribution based oversampling for imbalanced data classification.
\newblock \textit{IEEE Transactions on Knowledge and Data Engineering}, \textbf{34} 667--679.

\end{thebibliography}
\bibliographystyle{ims}

\appendix 
\section*{Effect of the classifier $f_\eta$ on customizing latent space} \label{app:cls}
In Section \ref{sec:pr}, we emphasize the importance of exploiting the superior performance of the tree-based classifier. Figure \ref{fig:appendix2} presents the visual representation of the filtering processes using a multilayer perceptron and XGBoost for $f_\eta$. 
`MLP' employs the multilayer perceptron for the $f_\eta$ function, and `XGB' denotes the proposed SMOTE-CLS with XGBoost for $f_\eta$. Comparing Figures \ref{fig:appendix2} (a) and (d), it demonstrates that utilizing the XGBoost classifier significantly enhances the disentanglement of the latent space, particularly concerning augmented labels, $\mathcal{Y}^*$. Furthermore, it's important to note that poor disentanglement can lead to lower interpretability. These findings underscore the pivotal role of our approach, which leverages a tree-based classifier to utilize its classification capabilities, aligning with our goals.

\setcounter{figure}{10}
\begin{figure}[ht]
    \centering
    \subfigure[MLP]{\includegraphics[width=0.26\linewidth]{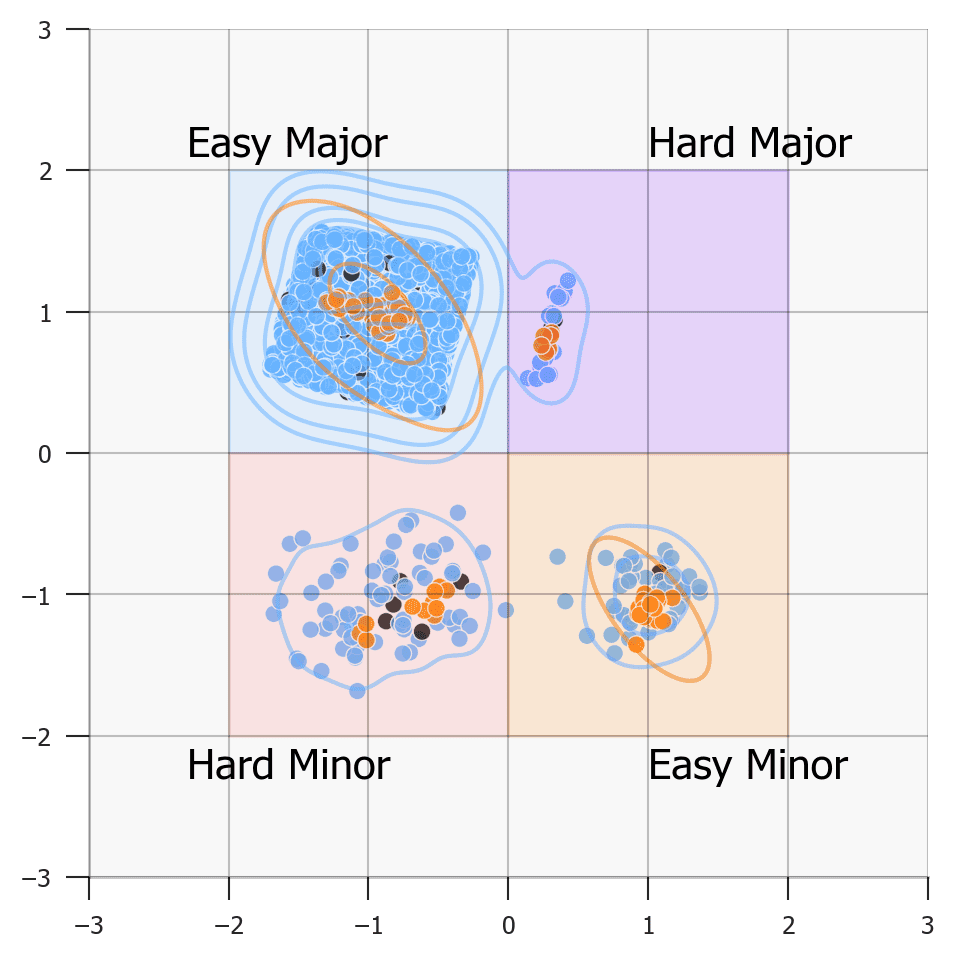}}
    \subfigure[MLP]{\includegraphics[width=0.26\linewidth]{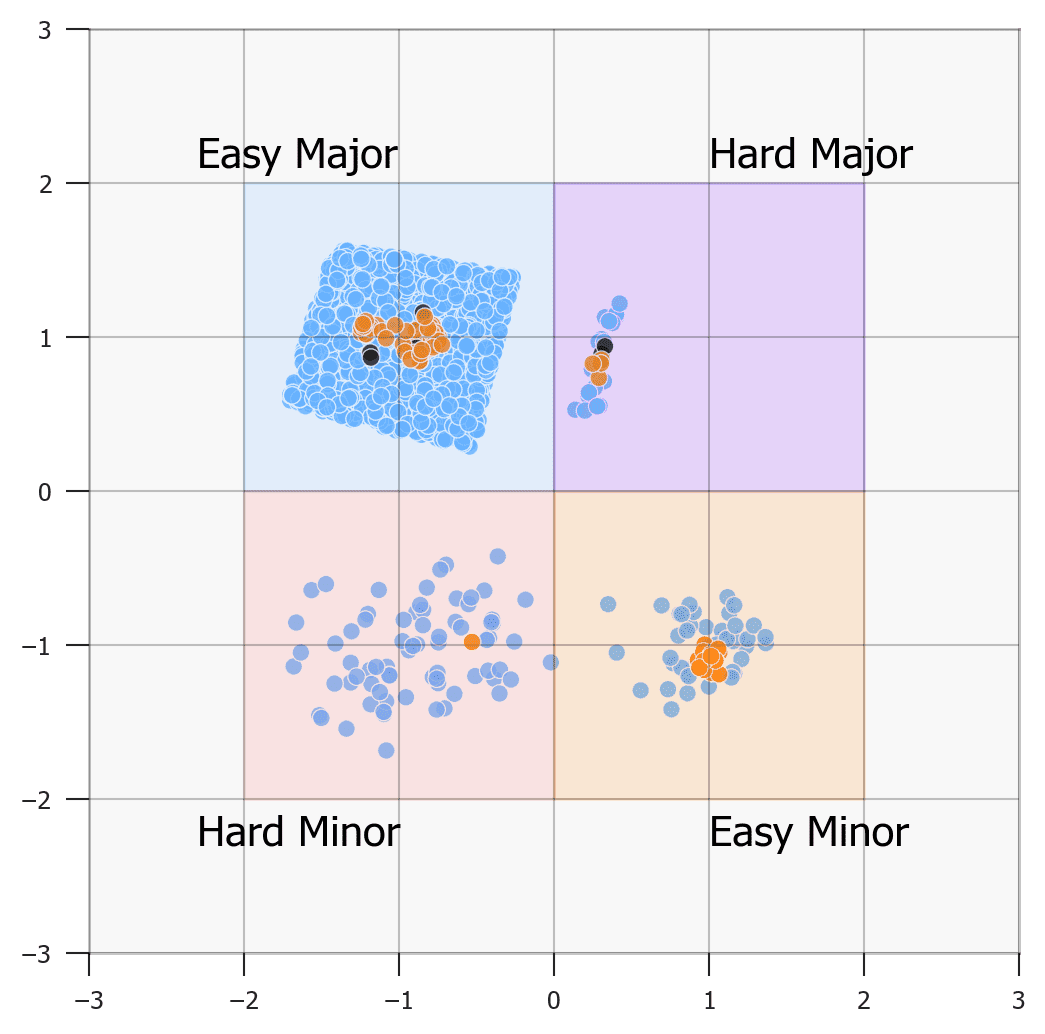}}
    \subfigure[MLP]{\includegraphics[width=0.27\linewidth]{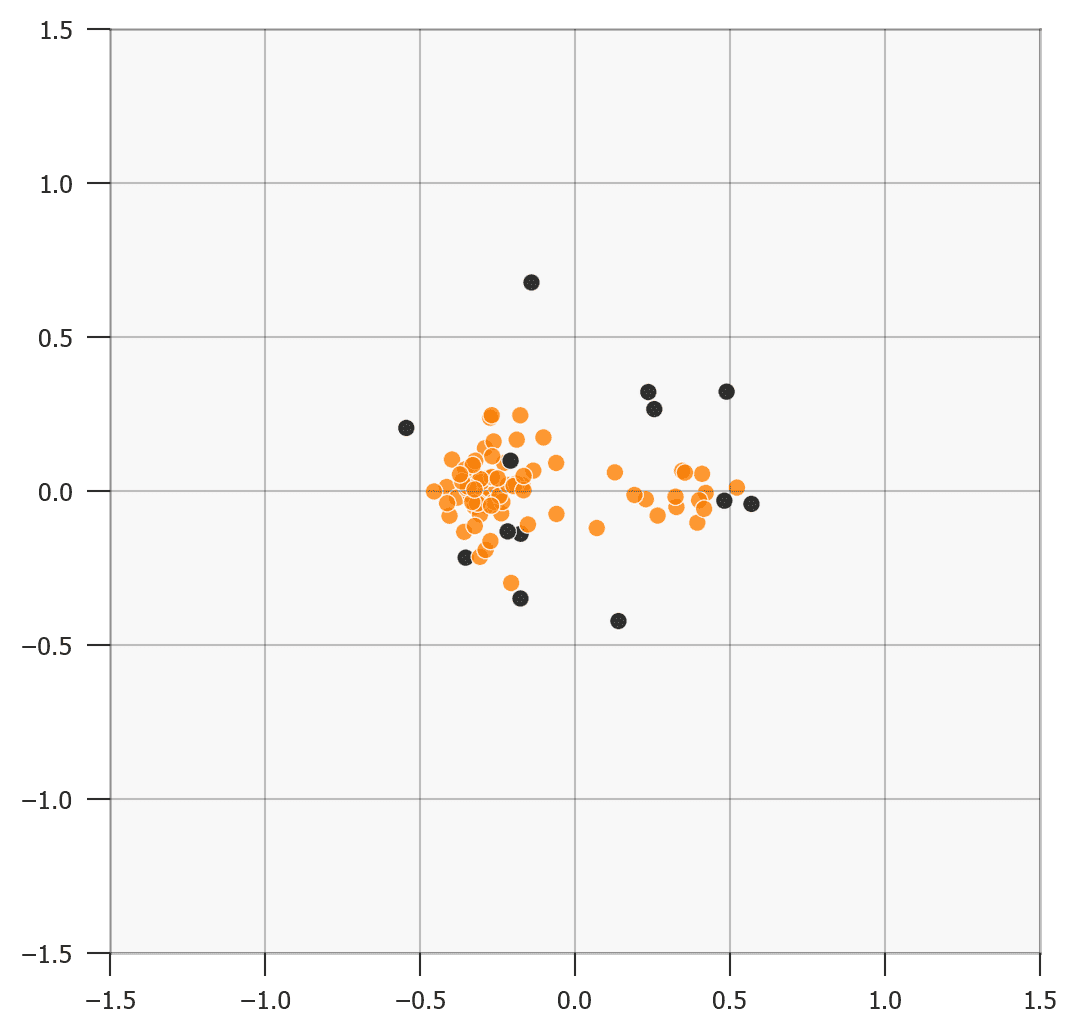}}
    \subfigure[XGB]{\includegraphics[width=0.26\linewidth]{fig/simul/simul_exon.png}}
    \subfigure[XGB]{\includegraphics[width=0.26\linewidth]{fig/simul/simul_exon_selected_latent.png}}
    \subfigure[XGB]{\includegraphics[width=0.27\linewidth]{fig/simul/simul_exon_selected_x.png}}
    \caption{Visualization of the effect of classifier on the latent space. (a) and (d): the trained latent variables. (b) and (e): the selected latent variables by filtering method. (c) and (f): the selected minority samples for oversampling.}
    \label{fig:appendix2}
\end{figure}

\newpage
\section*{Effects of prior distribution} \label{app:prior}
\begin{figure}[ht]
    \centering
    \subfigure[SMOTE-CLS1]{\includegraphics[width=0.24\linewidth]{fig/simul/simul_exon.png}}
    \subfigure[SMOTE-CLS2]{\includegraphics[width=0.24\linewidth]{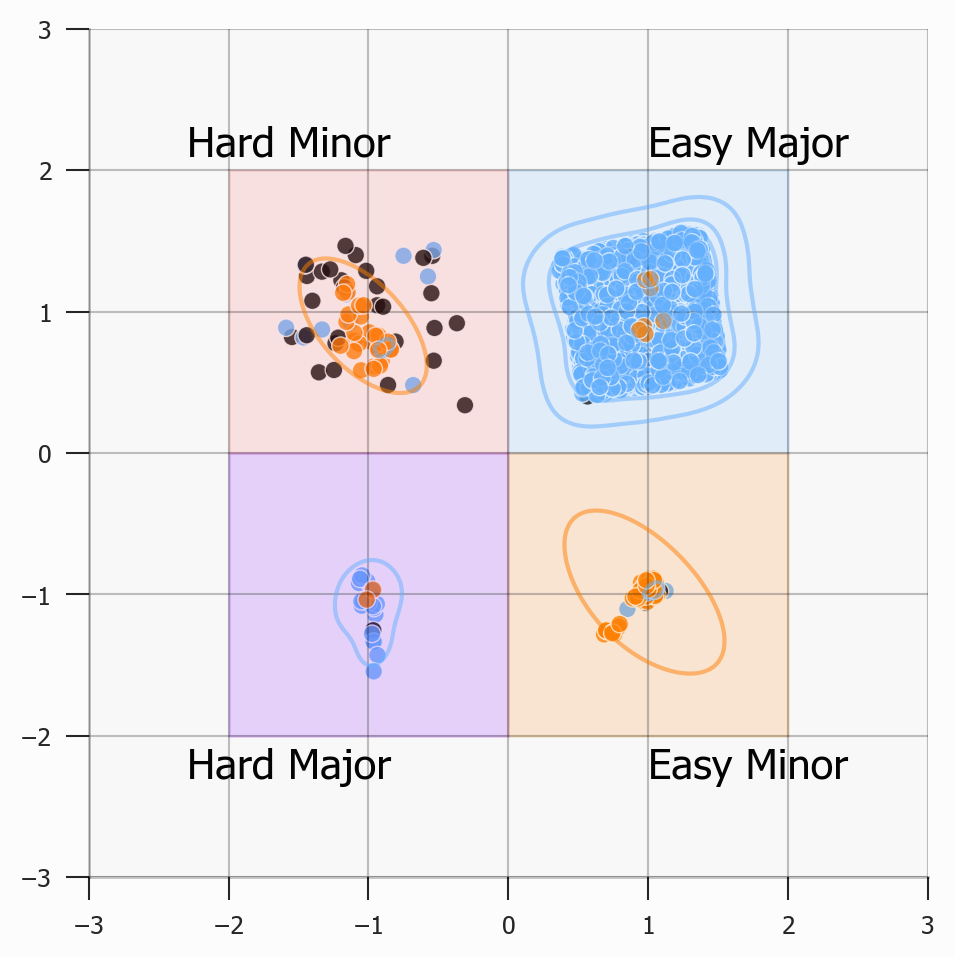}}
    \subfigure[SMOTE-CLS3]{\includegraphics[width=0.24\linewidth]{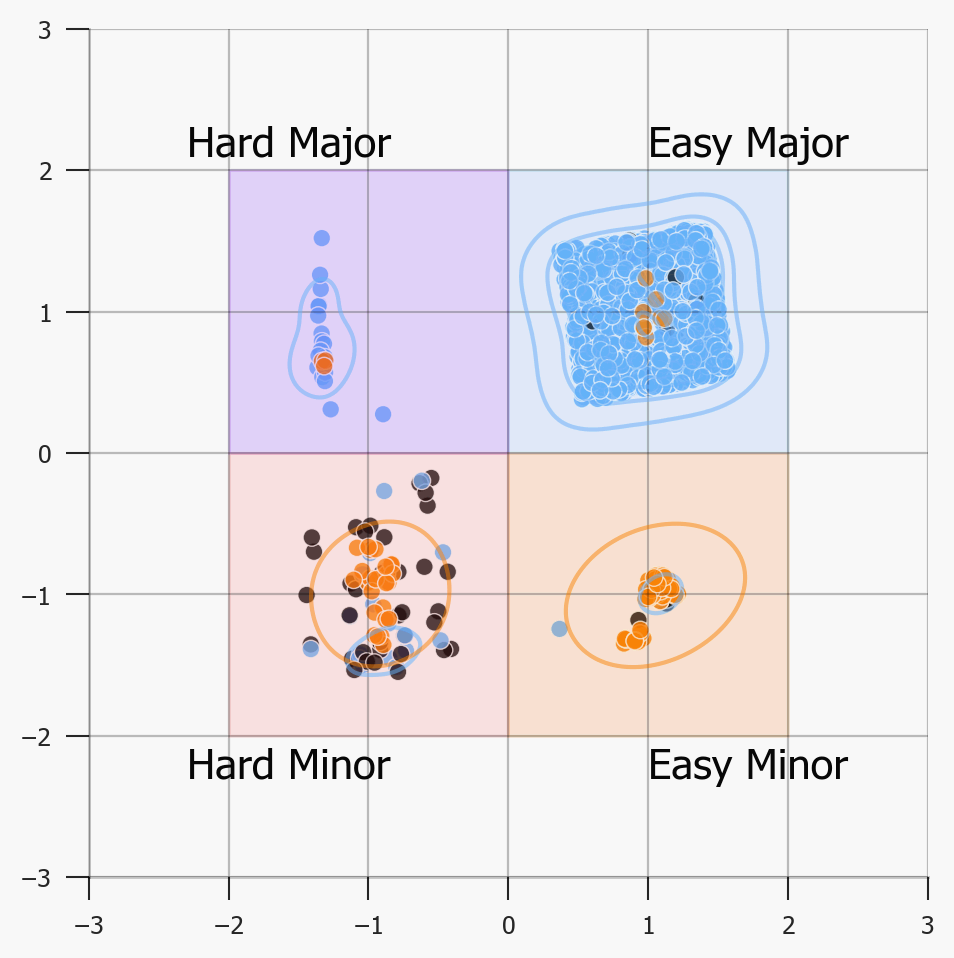}}
    \subfigure[SMOTE-CLS4]{\includegraphics[width=0.24\linewidth]{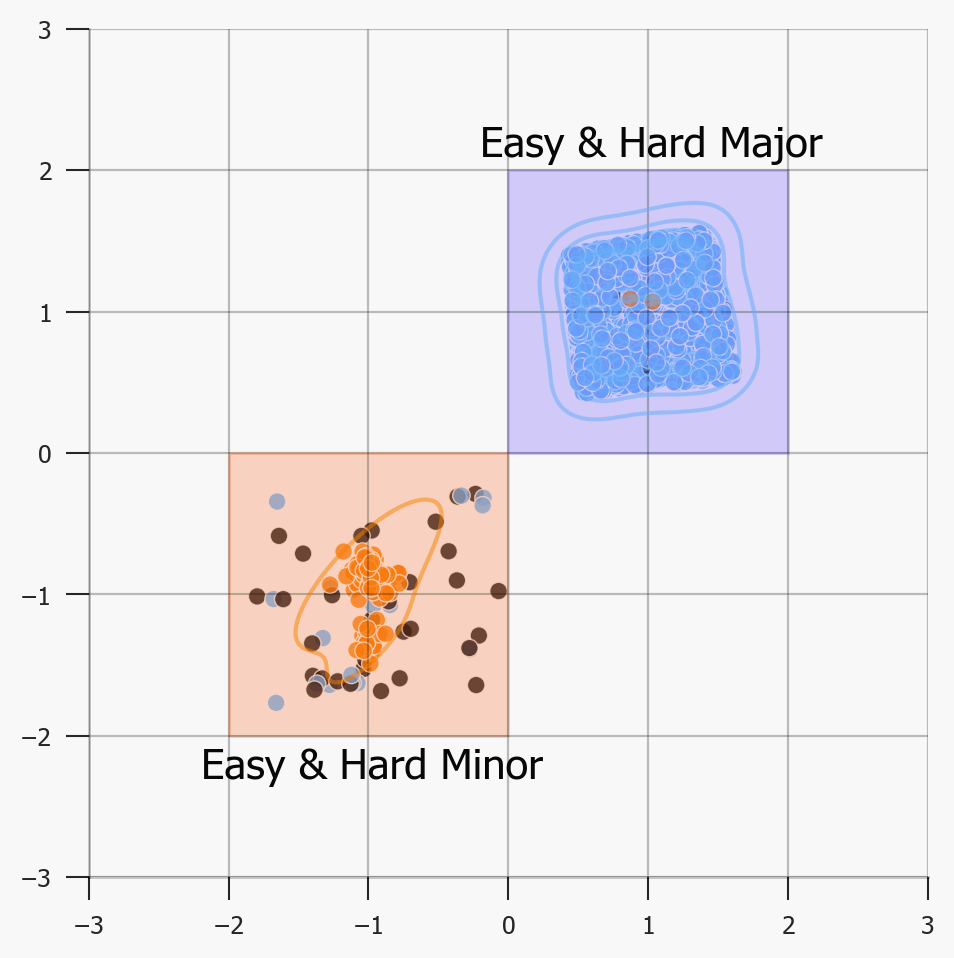}}
    \subfigure[SMOTE-CLS1]{\includegraphics[width=0.24\linewidth]{fig/simul/simul_exon_selected_latent.png}}
    \subfigure[SMOTE-CLS2]{\includegraphics[width=0.24\linewidth]{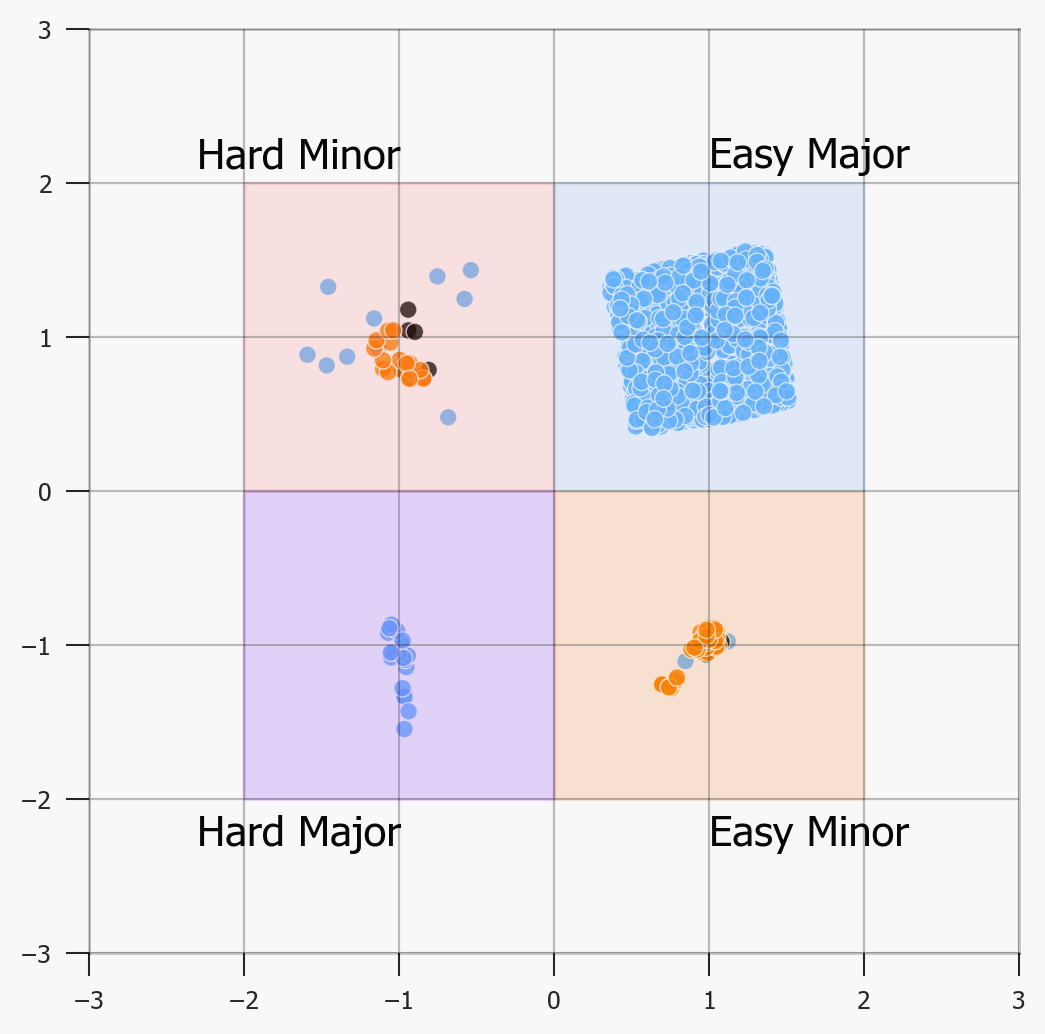}}
    \subfigure[SMOTE-CLS3]{\includegraphics[width=0.24\linewidth]{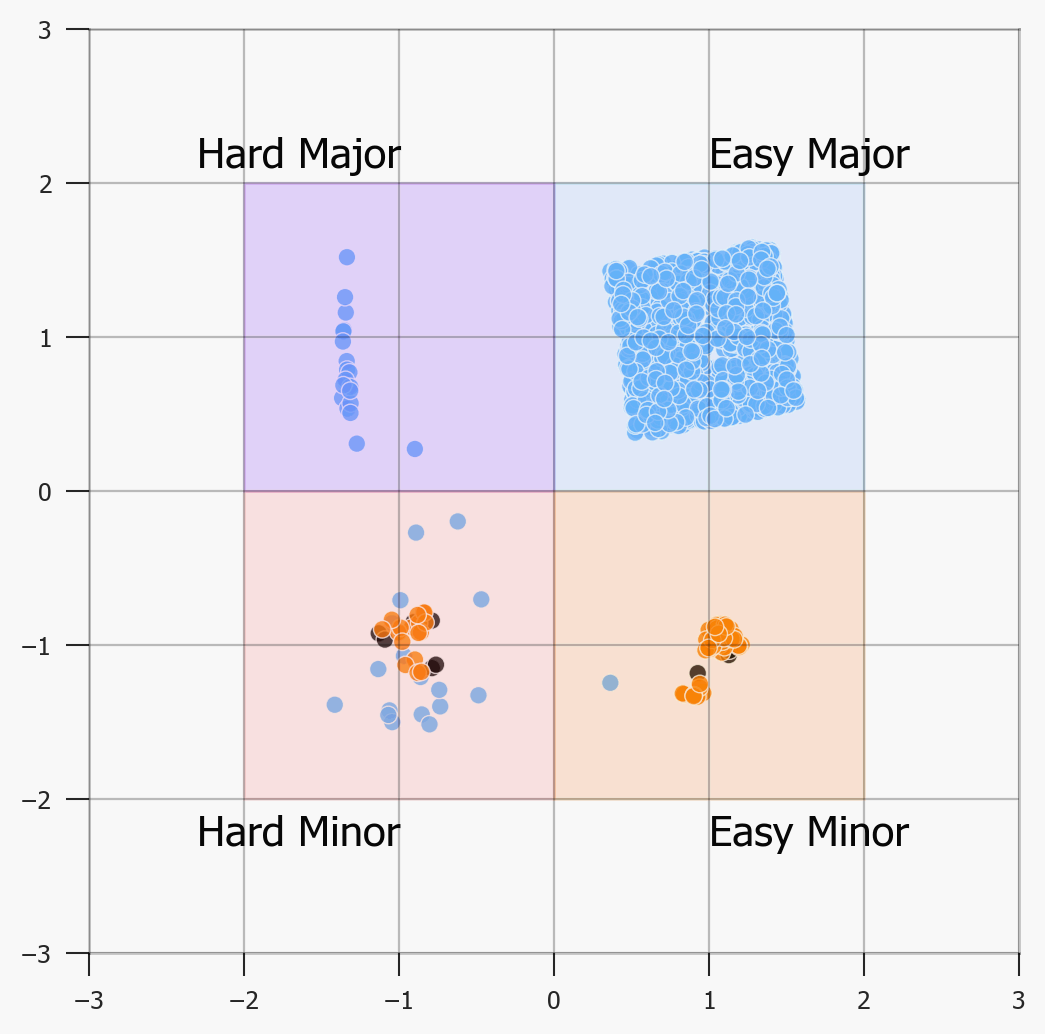}}
    \subfigure[SMOTE-CLS4]{\includegraphics[width=0.24\linewidth]{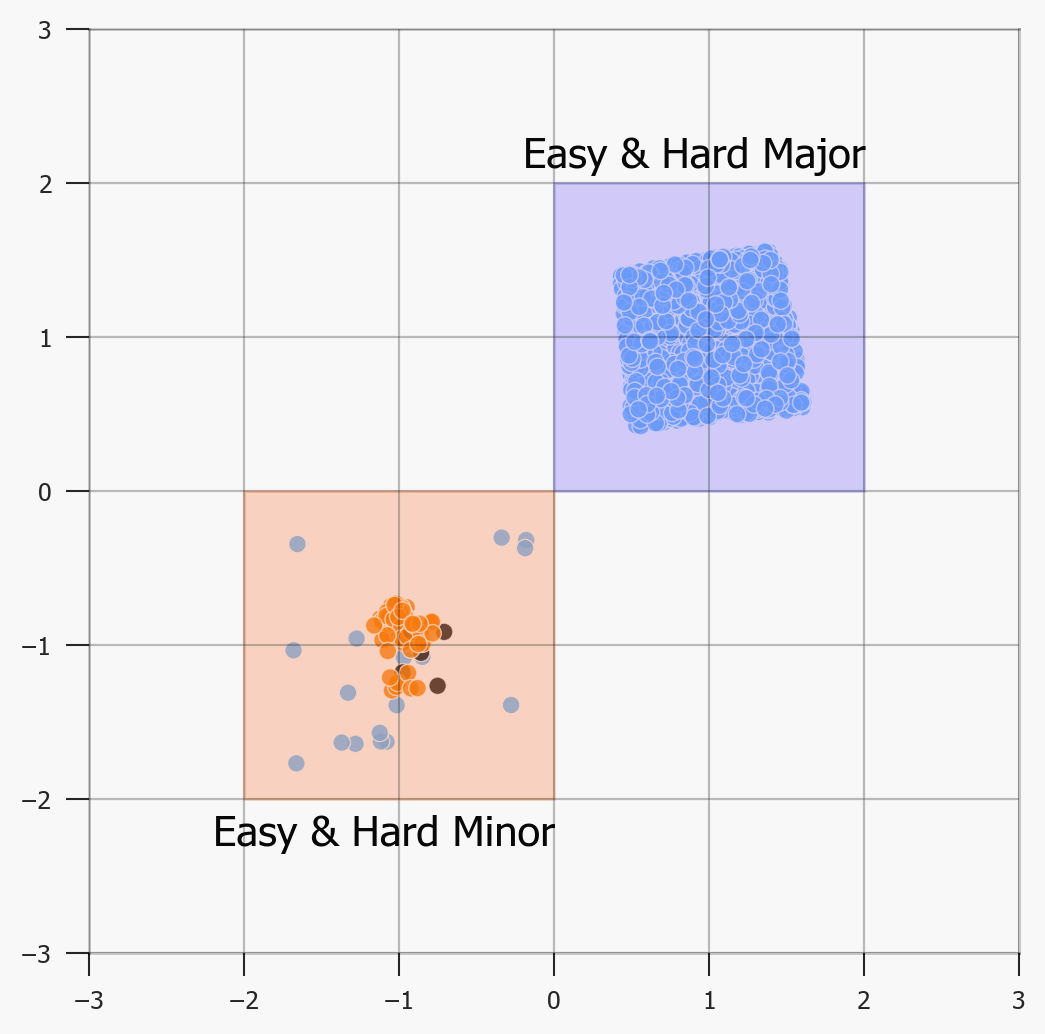}}
    \subfigure[SMOTE-CLS1]{\includegraphics[width=0.24\linewidth]{fig/simul/simul_exon_selected_x.png}}
    \subfigure[SMOTE-CLS2]{\includegraphics[width=0.24\linewidth]{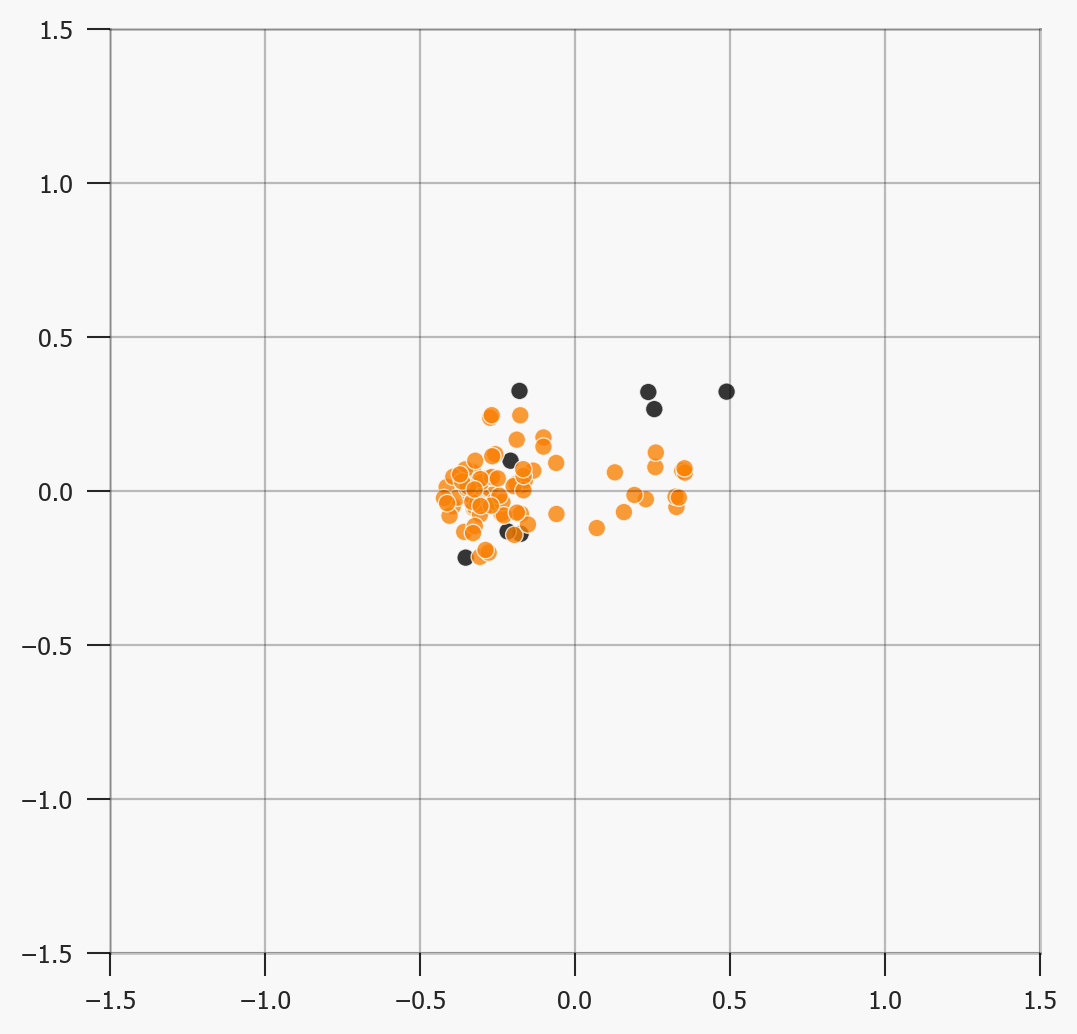}}
    \subfigure[SMOTE-CLS3]{\includegraphics[width=0.24\linewidth]{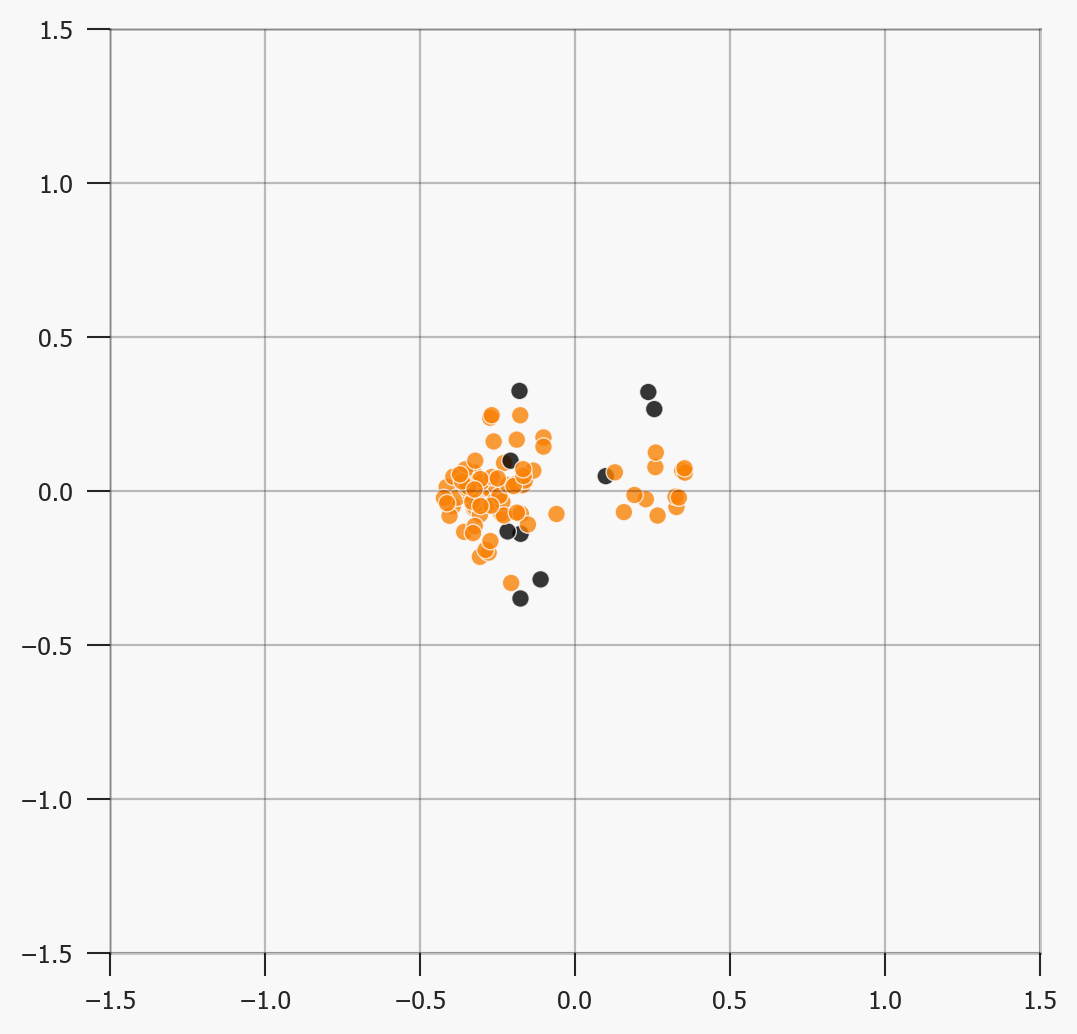}}
    \subfigure[SMOTE-CLS4]{\includegraphics[width=0.24\linewidth]{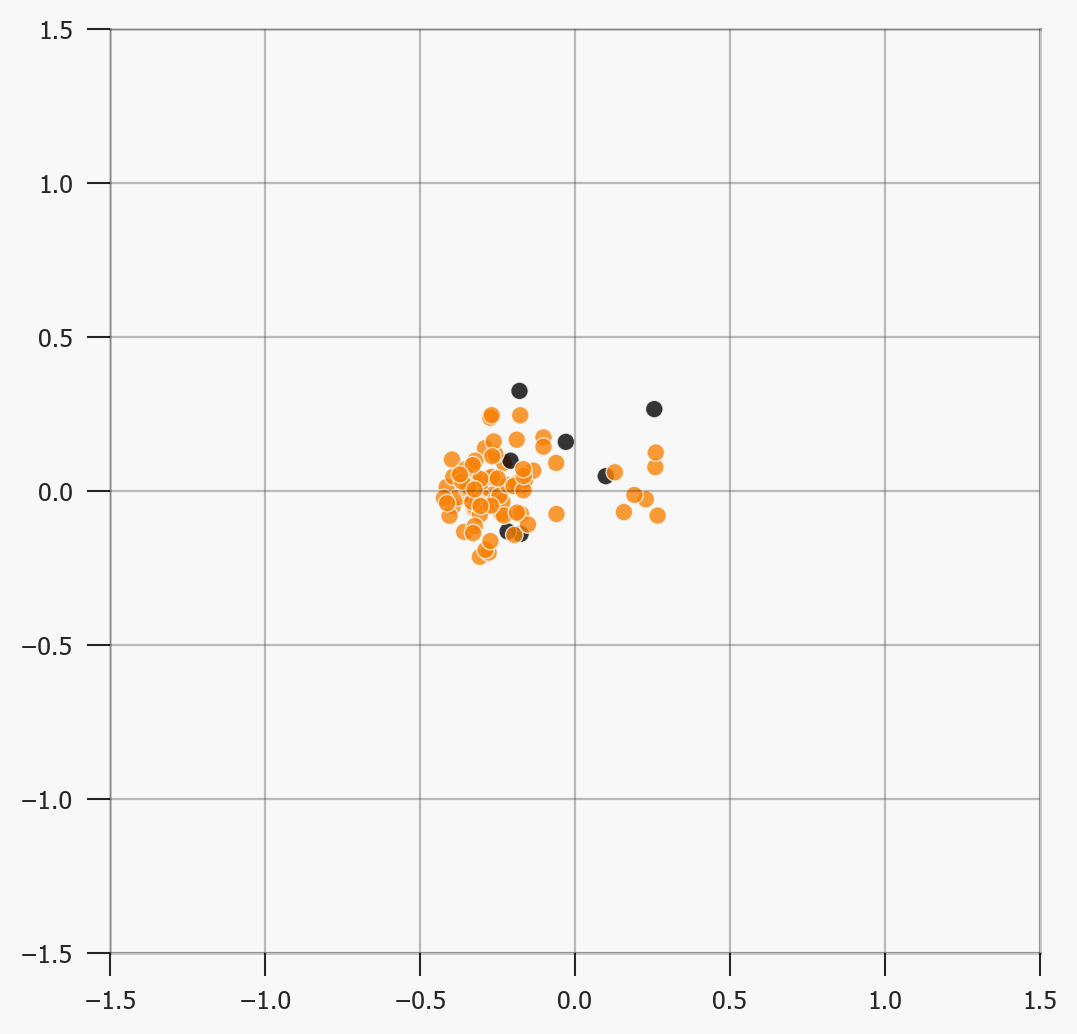}}
    \caption{Visualization of the ablation study. (a) - (d): the trained latent variables. (e) - (h): the selected latent variables by each filtering method. (i) - (l): the selected minority samples for oversampling.}
    \label{fig:appendix}
\end{figure}

We investigate the effects of location parameters of the prior. Figure \ref{fig:appendix} visually represents the filtering process using SMOTE-CLS with different prior settings. Each colored square represents the predefined area for each group.
`SMOTE-CSL1' presents SMOTE-CLS used in Section \ref{sec:simul} and \ref{sec:ablation}. In `SMOTE-CSL2', we adjust the distances within each class to be greater than one between classes. For 'SMOTE-CSL3', we allocate the x-axis to represent difficulty and the y-axis to denote class. In `SMOTE-CLS4', we set the same location parameters for hard and easy samples of each class. As illustrated in Figures \ref{fig:appendix} (i) - (l), our methods are robust to the selection of the prior. 

\newpage
\section*{Effect of the threshold $\tau$s} \label{app:thr}
\begin{figure}[ht]
    \centering
    \subfigure[70\%]{\includegraphics[width=0.26\linewidth]{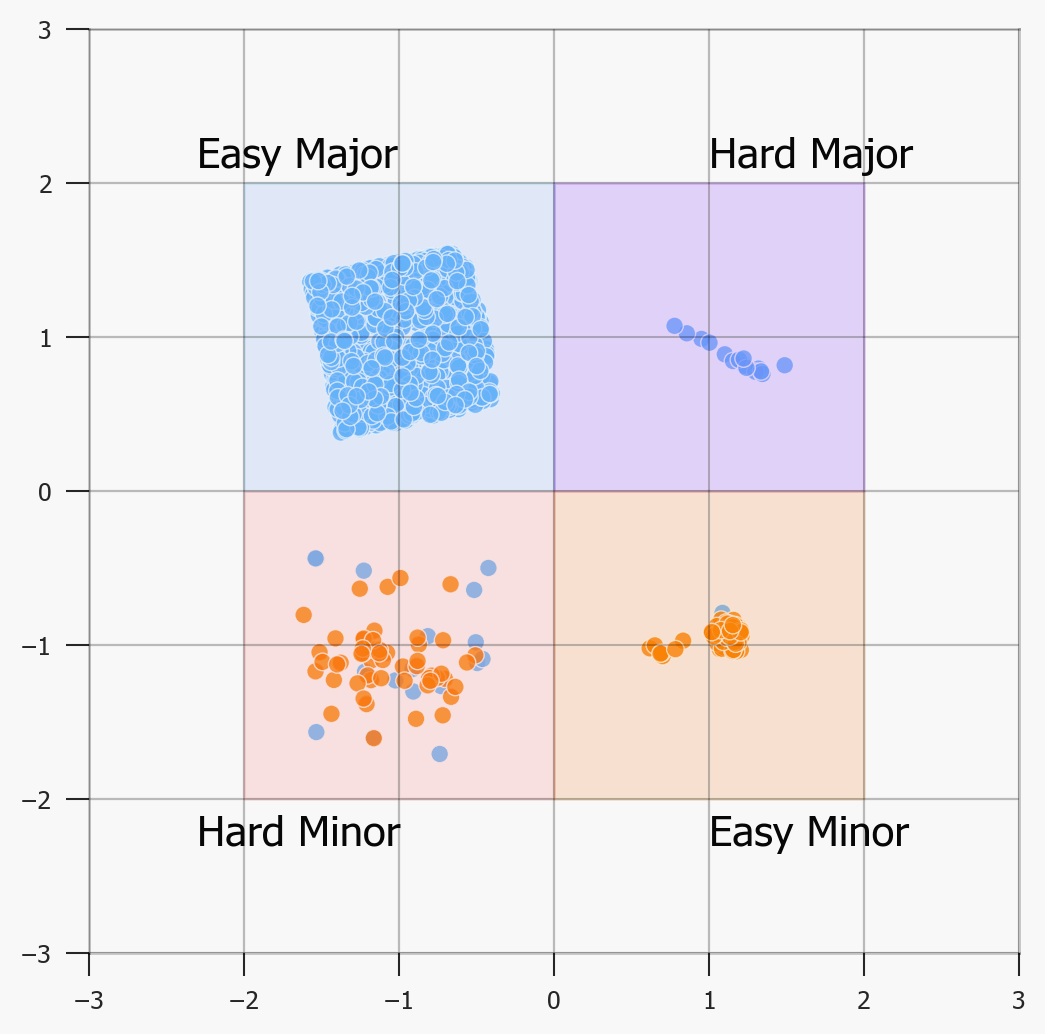}}
    \subfigure[50\%]{\includegraphics[width=0.26\linewidth]{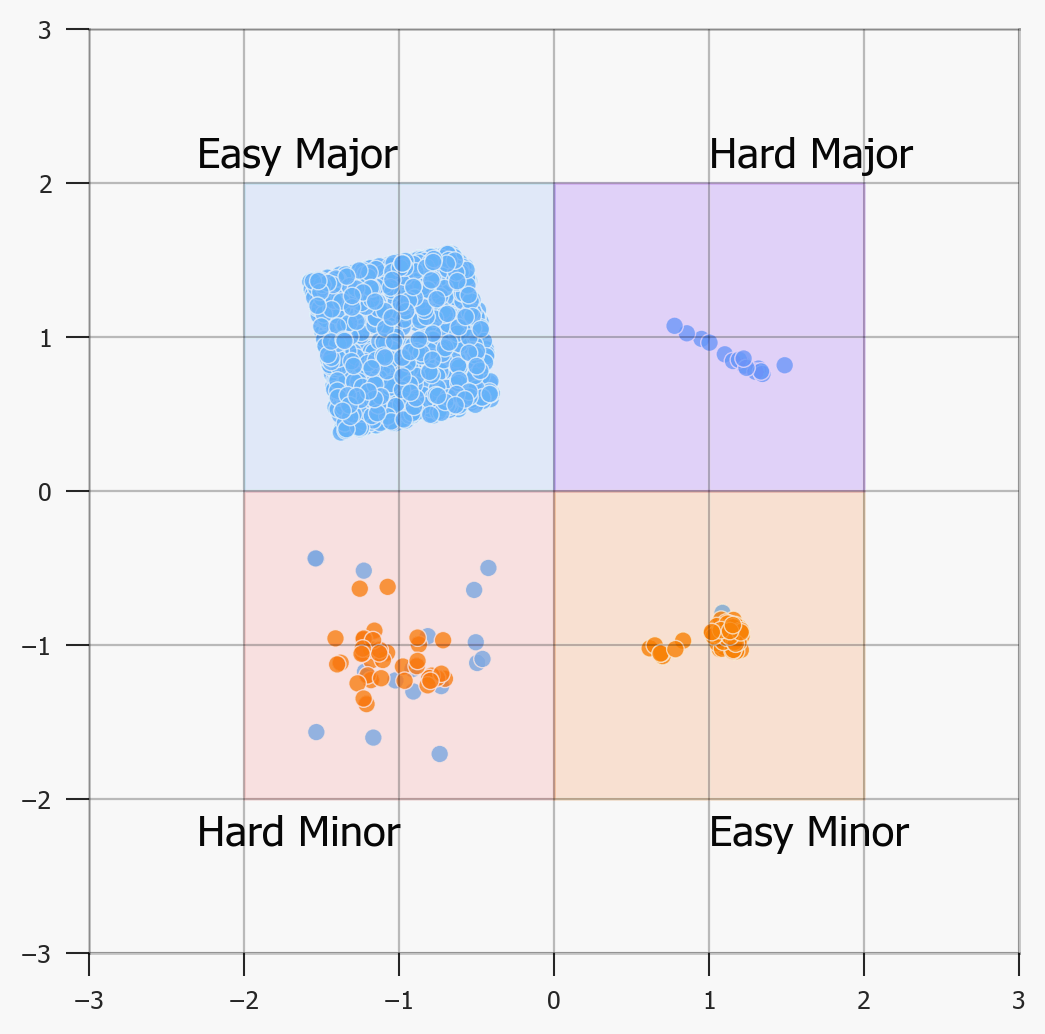}}
    \subfigure[30\%]{\includegraphics[width=0.26\linewidth]{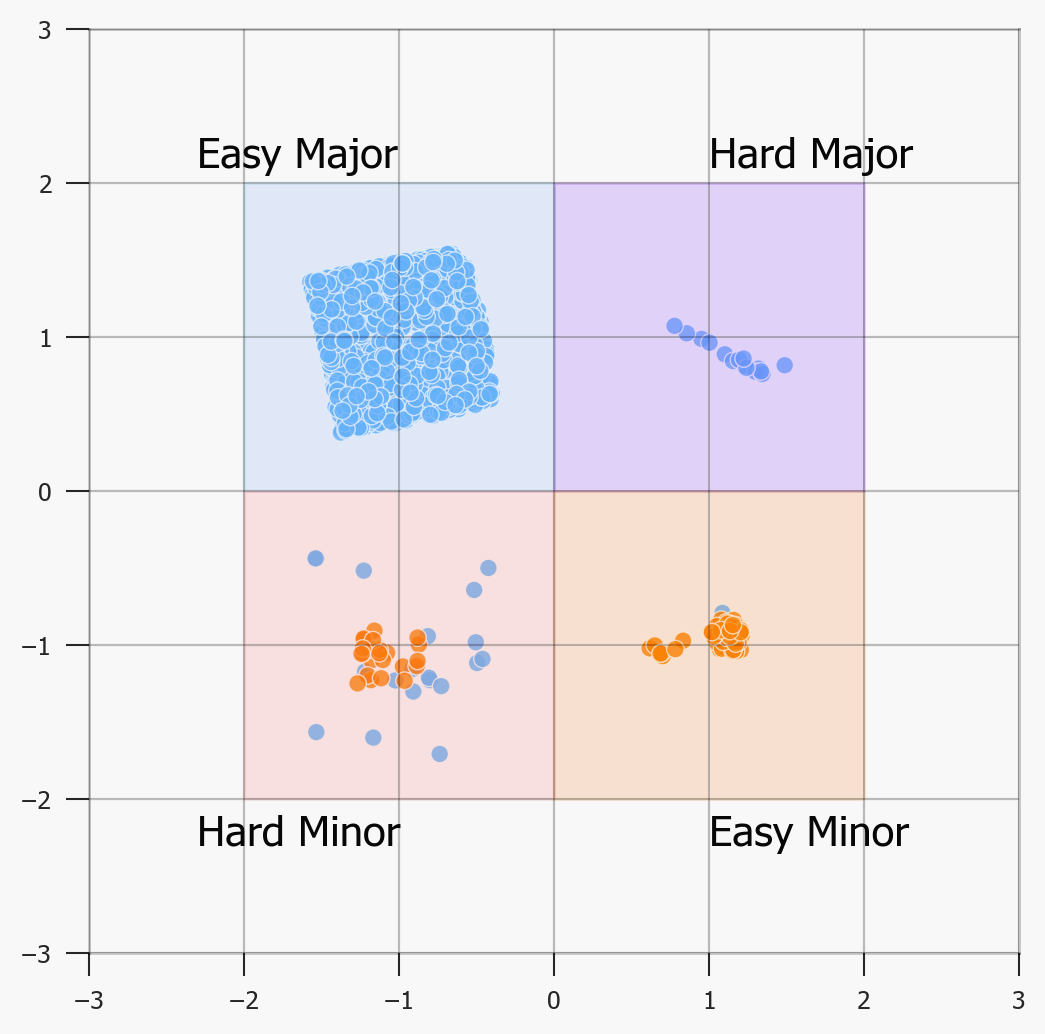}}
    \subfigure[70\%]{\includegraphics[width=0.27\linewidth]{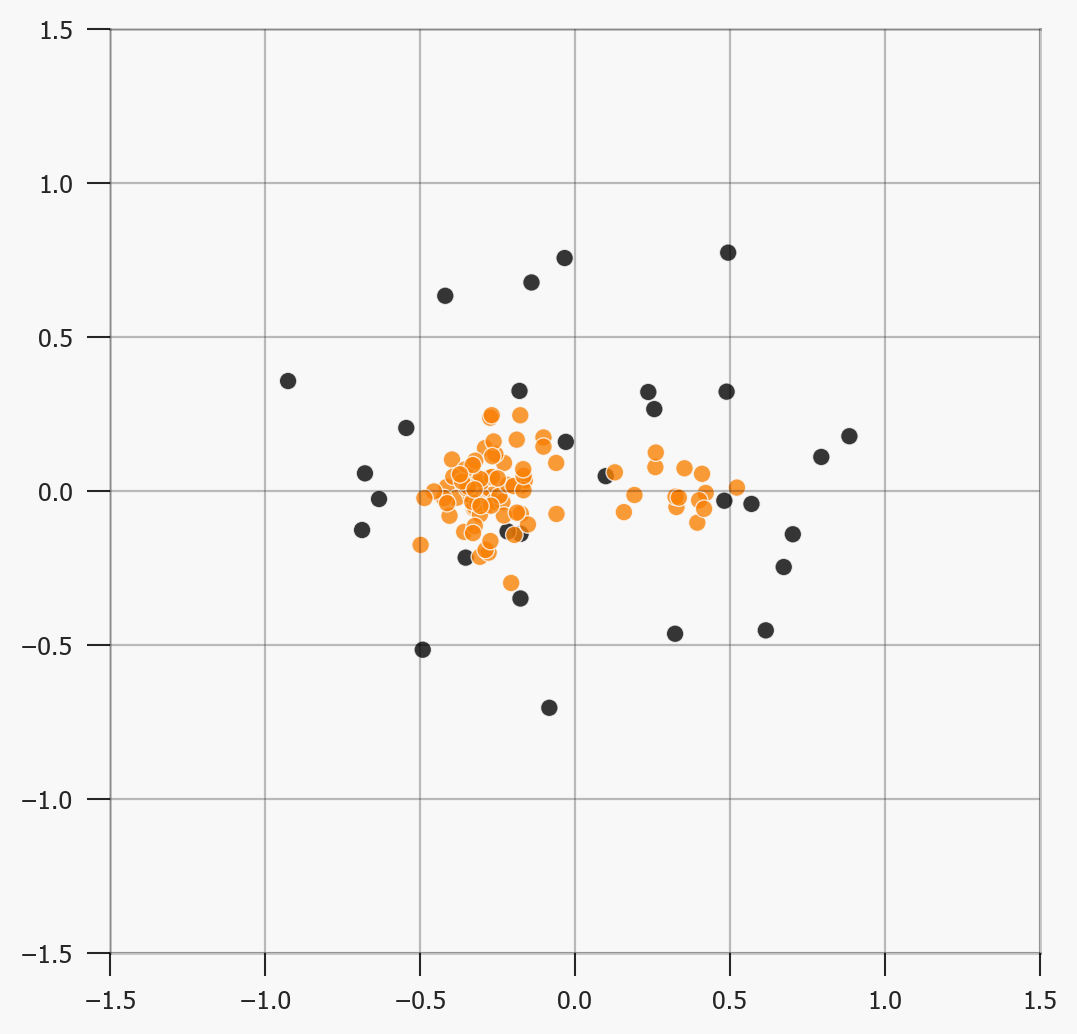}}
    \subfigure[50\%]{\includegraphics[width=0.27\linewidth]{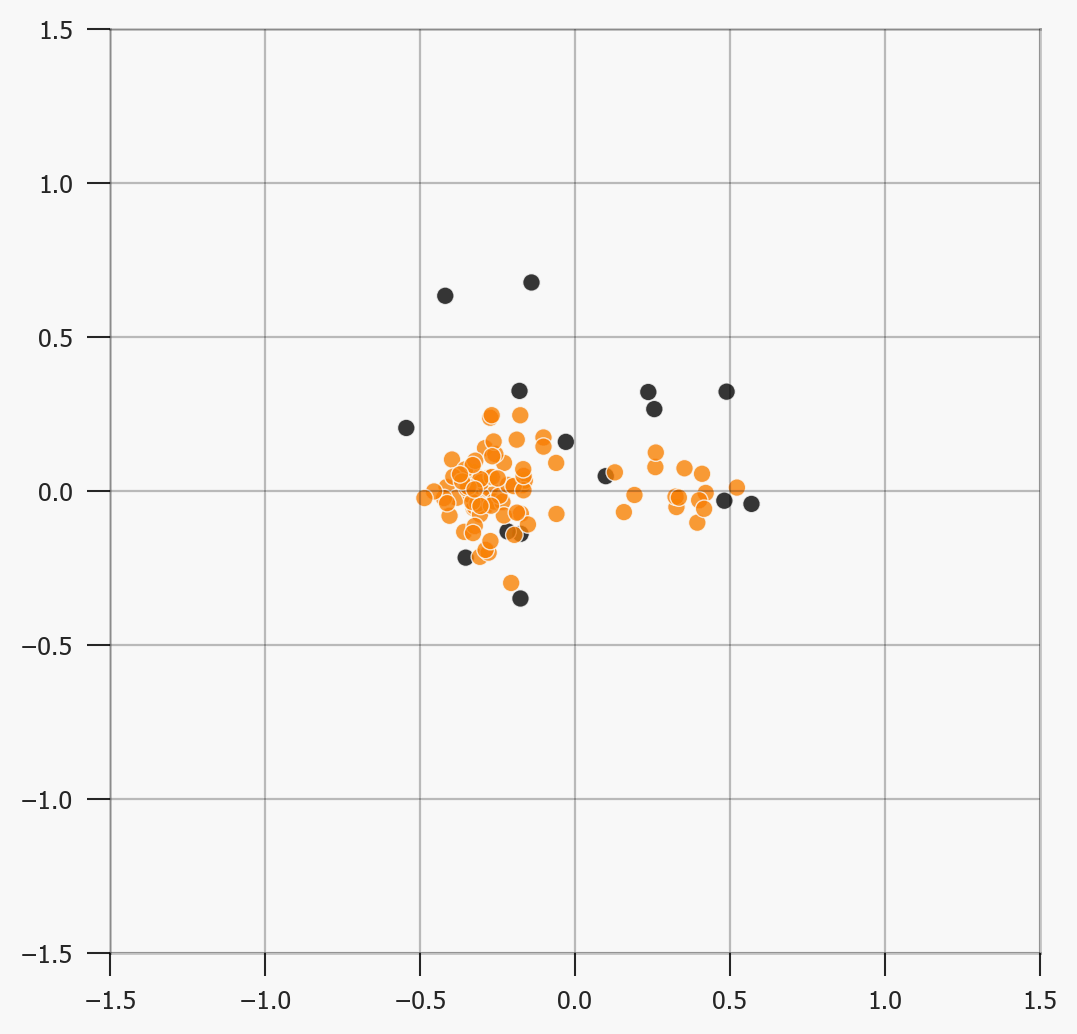}}
    \subfigure[30\%]{\includegraphics[width=0.27\linewidth]{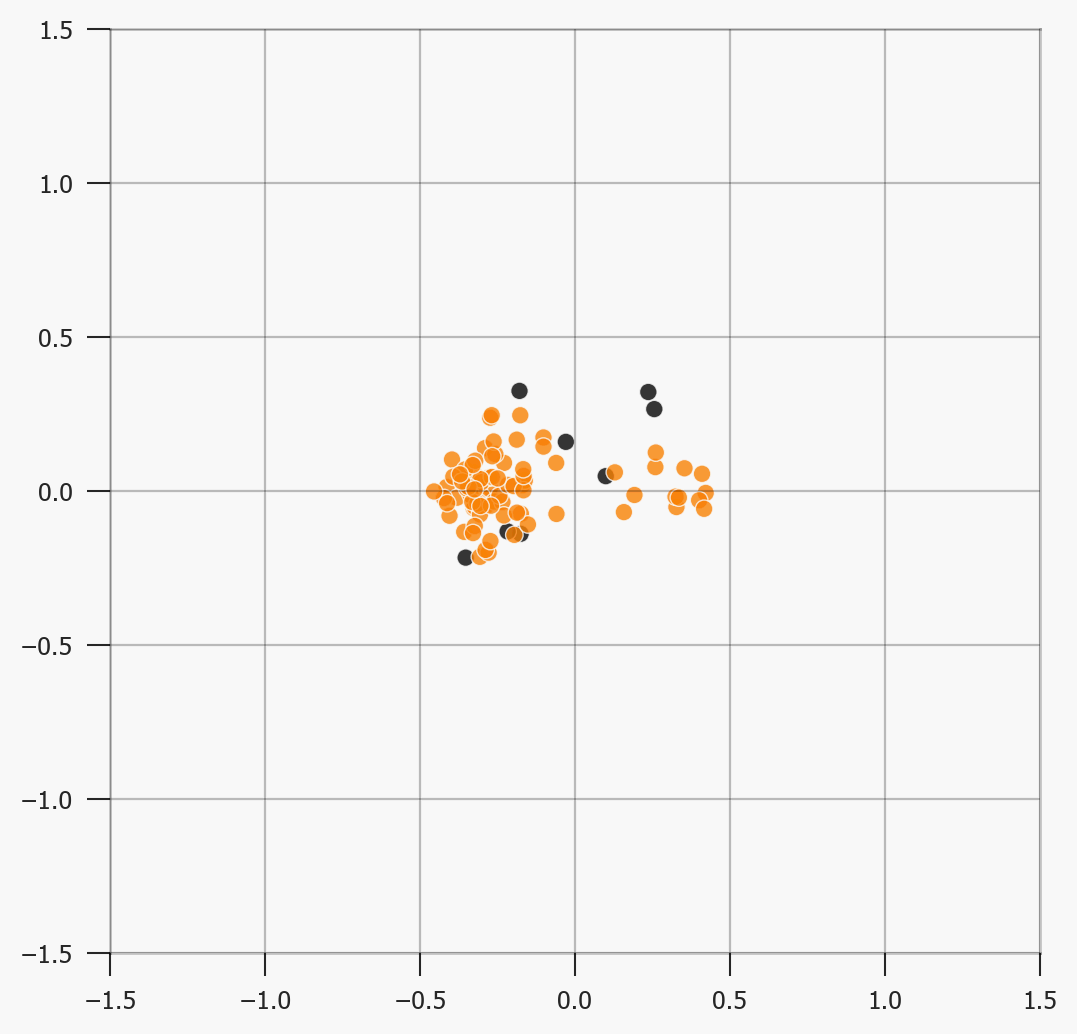}}
    \caption{Tuning $\tau_2$ with visual representations.}
    \label{fig:appendix4}
\end{figure}

In the process of group-adaptive filtering described in Equation \eqref{eq:gf}, noise identification depends on the choice of tuning parameters, denoted as $\tau_1$ and $\tau_2$. 
To investigate how to select these tuning parameters effectively, we turn to visual representations using the simulation data from Section \ref{sec:simul}. 
For simplicity, we set $\tau_1$ to be selected 10\% samples in $D_m$ and then explore different values of $\tau_2$ to be selected 30\%, 50\%, and 70\% samples in $D^*_m$. The first row of Figure \ref{fig:appendix4} displays the latent variables selected by varying $\tau_2$, without labeling noise in practice. This visual exploration helps us determine an appropriate choice for $\tau_2$. In Figures \ref{fig:appendix4} (a) and (b), some hard minority samples that are selected appear to be distant from their centers, suggesting they may be noise. Conversely, in Figure \ref{fig:appendix4} (c), all hard-minority samples are clustered around their center. This observation suggests that selecting 30\% of $D^*_m$ could be considered an optimal parameter choice.
The second row of Figure \ref{fig:appendix4} illustrates the selected minority samples based on $\tau_2$.

\newpage
\section*{Other evaluation metrics} \label{app:metrics}
This section presents the classification performances of benchmark methods using additional evaluation metrics, specifically the F1-score and geometric mean. Our proposed method consistently achieves higher performance in both of these metrics.
\subsubsection*{F1-score}
\bed
    \item F1-score $= 2 \times \frac{\mbox{Precision} \times \mbox{Recall}}{\mbox{Precision} + \mbox{Recall}}$
\eed

\begin{sidewaystable}[ph!]
\caption{Classifcation performance (F1-score) of different oversampling models using the random forest on the imbalanced datasets. The numbers within parentheses indicate their rankings, excluding BASE. The most favorable value is bolded.}
{\footnotesize\setlength{\tabcolsep}{1pt} 
  \centering
\begin{tabular}{lcccccccccc}
    \toprule
    \textsf{Dataset} & BASE & SMOTE & BSMOTE & SMOTEENN & KMSMOTE & DFBS & CVAE & DeepSMOTE & DDHS  & SMOTE-CLS \\
    \midrule
    \texttt{ecoli} & $0.784_{\pm0.016}$ & $0.793_{\pm0.013}~(3)$ & $0.777_{\pm 0.007}~(6)$ & $0.774_{\pm0.005}~(5)$ & $\mathbf{0.840_{\pm0.013}~(1)}$ & $0.778_{\pm 0.011}~(4)$ & $0.751_{\pm 0.021}~(8)$ & $0.744_{\pm 0.022}~(9)$ & $0.761_{\pm 0.015}~(7)$ & $0.820_{\pm 0.010}~(2)$ \\
    \texttt{libras\_move} & $0.733_{\pm0.029}$ & $0.624_{\pm 0.024}~(8)$ & $0.636_{\pm 0.029}~(7)$ & $0.614_{\pm 0.029}~(9)$ & $0.659_{\pm 0.028}~(5)$ & $0.702_{\pm 0.029}~(2)$ & $0.674_{\pm 0.038}~(4)$ & $0.675_{\pm 0.034}~(3)$ & $\mathbf{0.717_{\pm 0.026}~(1)}$ & $0.640_{\pm 0.039}~(6)$ \\
    \texttt{spectrometer} & $0.816_{\pm0.017}$ & $0.819_{\pm 0.015}~(6)$ & $0.822_{\pm 0.011}~(4)$ & $0.819_{\pm 0.008}~(5)$ & $\mathbf{0.836_{\pm 0.010}~(1)}$ & $0.748_{\pm 0.016}~(8)$ & $0.767_{\pm 0.021}~(7)$ & $0.833_{\pm 0.018}~(3)$ & $0.721_{\pm 0.015}~(9)$ & $0.833_{\pm 0.014}~(2)$ \\
    \texttt{oil} & $0.704_{\pm0.021}$ & $0.775_{\pm0.012}~(3)$ & $0.764_{\pm 0.024}~(4)$ & $\mathbf{0.784_{\pm0.017}~(1)}$ & $0.652_{\pm 0.018}~(9)$ &  $0.752_{\pm 0.014}~(6)$ & $0.669_{\pm 0.014}~(8)$ & $0.687_{\pm 0.018}~(7)$ & $0.756_{\pm 0.026}~(5)$ & $0.776_{\pm 0.010}~(2)$ \\
    \texttt{yeast} & $0.507_{\pm 0.008}$ & $0.668_{\pm 0.015}~(4)$ & $\mathbf{0.706_{\pm 0.014}~(1)}$ &$0.688_{\pm 0.007}~(3)$ & $0.697_{\pm 0.011}~(2)$ & $0.528_{\pm 0.011}~(6)$ & $0.514_{\pm 0.009}~(8)$ & $0.520_{\pm 0.012}~(7)$ & $0.495_{\pm 0.005}~(9)$ & $0.665_{\pm 0.016}~(5)$ \\
    \texttt{car\_eval} & $0.895_{\pm 0.006}$ & $0.934_{\pm 0.004}~(3)$ & $\mathbf{0.941_{\pm 0.003}~(1)}$ & $0.940_{\pm 0.006}~(2)$ & $0.929_{\pm 0.006}~(4)$ & $0.901_{\pm 0.007}~(6)$ & $0.875_{\pm 0.011}~(9)$ & $0.886_{\pm 0.008}~(8)$ & $0.893_{\pm 0.005}~(7)$ & $0.927_{\pm 0.006}~(5)$ \\
    \texttt{us\_crime} & $0.615_{\pm 0.009}$ & $0.676_{\pm 0.008}~(4)$ & $0.667_{\pm 0.008}~(5)$ & $0.683_{\pm 0.004}~(2)$ & $\mathbf{0.686_{\pm 0.010}~(1)}$ & $0.647_{\pm 0.006}~(6)$ & $0.619_{\pm 0.006}~(8)$ & $0.618_{\pm 0.012}~(9)$ & $0.645_{\pm 0.005}~(7)$ & $0.677_{\pm 0.010}~(3)$\\
    \texttt{scene} & $0.546_{\pm 0.005}$ & $0.626_{\pm 0.008}~(2)$ & $0.620_{\pm 0.007}~(3)$ & $0.558_{\pm 0.004}~(6)$ & $0.608_{\pm 0.007}~(4)$ & $0.546_{\pm 0.003}~(7)$ & $0.536_{\pm 0.000}~(9)$ & $0.539_{\pm 0.003}~(8)$ & $0.561_{\pm 0.006}~(5)$ & $\mathbf{0.633_{\pm 0.006}~(1)}$ \\
    \texttt{abalone} & $0.482_{\pm 0.003}$ & $0.654_{\pm 0.003}~(5)$ & $0.659_{\pm 0.003}~(4)$ & $0.587_{\pm 0.002}~(6)$ & $0.565_{\pm 0.004}~(7)$ & $0.680_{\pm 0.003}~(2)$ & $0.506_{\pm 0.005}~(8)$ & $0.499_{\pm 0.004}~(9)$ & $\mathbf{0.682_{\pm 0.002}~(1)}$ & $0.662_{\pm 0.003}~(3)$\\
    \texttt{optical\_digits} & $0.865_{\pm 0.004}$ & $0.911_{\pm 0.004}~(2)$ & $0.870_{\pm 0.004}~(6)$ & $0.893_{\pm 0.006}~(4)$ & $0.897_{\pm 0.003}~(3)$ & $0.849_{\pm 0.003}~(7)$ & $0.765_{\pm 0.012}~(8)$ & $0.669_{\pm 0.017}~(9)$ & $0.874_{\pm 0.003}~(5)$ & $\mathbf{0.915_{\pm 0.003}~ (1)}$ \\
    \texttt{satimage} & $0.796_{\pm 0.003}$ & $0.787_{\pm 0.003}~(6)$ & $0.744_{\pm 0.002}~(8)$ & $0.737_{\pm 0.003}~(9)$ & $0.765_{\pm 0.004}~(7)$ & $0.792_{\pm 0.003}~(4)$ & $0.791_{\pm 0.004}~(5)$ & $0.801_{\pm 0.003}~(2)$ & $\mathbf{0.802_{\pm 0.005}~(1)}$ & $0.795_{\pm 0.003}~(3)$ \\
    \texttt{isolet} & $0.693_{\pm 0.007}$ & $0.817_{\pm 0.002}~(3)$ & $0.799_{\pm 0.005}~(6)$ & $0.781_{\pm 0.005}~(7)$ & $0.827_{\pm 0.004}~(2)$ & $0.800_{\pm 0.004}~(5)$ & $0.567_{\pm 0.008}~(8)$ & $0.539_{\pm 0.013}~(9)$ & $0.815_{\pm 0.004}~(4)$ & $\mathbf{0.831_{\pm 0.003}~(1)}$ \\
    Average rank & - & 4.08 & 4.58 & 4.92 & 3.83 & 5.25 & 7.5 & 6.92 & 5.08 & \textbf{2.83} \\ 
    \bottomrule
  \end{tabular}
}

\label{tab:f1}
\end{sidewaystable}

\subsubsection*{Geometric mean (G-MEAN)}
\bed
    \item G-MEAN = $\sqrt{\mbox{Sensitivity (Positive accuracy)} \times \mbox{Specificity (Negative accuracy)}}$
\eed

\begin{sidewaystable}[ph!]
\caption{Classifcation performance (G-MEAN) of different oversampling models using the random forest on the imbalanced datasets. The numbers within parentheses indicate their rankings, excluding BASE. The most favorable value is bolded.}
{\footnotesize\setlength{\tabcolsep}{1pt} 
  \centering
\begin{tabular}{lcccccccccc}
    \toprule
    \textsf{Dataset} & BASE & SMOTE & BSMOTE & SMOTEENN & KMSMOTE & DFBS & CVAE & DeepSMOTE & DDHS  & SMOTE-CLS \\
    \midrule
    \texttt{ecoli} & $0.753_{\pm0.020}$ & $0.904_{\pm0.013}~(4)$ & $0.887_{\pm 0.013}~(5)$ & $\mathbf{0.919_{\pm0.009}~(1)}$ & $0.905_{\pm0.019}~(3)$ & $0.805_{\pm 0.011}~(6)$ & $0.717_{\pm 0.020}~(8)$ & $0.705_{\pm 0.022}~(9)$ & $0.740_{\pm 0.015}~(7)$ & $0.915_{\pm 0.011}~(2)$ \\
    \texttt{libras\_move} & $0.670_{\pm0.020}$ & $0.596_{\pm 0.019}~(9)$ & $0.613_{\pm 0.022}~(6)$ & $0.602_{\pm 0.024}~(8)$ & $0.616_{\pm 0.020}~(5)$ & $0.649_{\pm 0.021}~(2)$ & $0.629_{\pm 0.028} ~ (4)$ & $0.629_{\pm 0.024}~(3)$ & $\mathbf{0.659_{\pm 0.021}~(1)}$ & $0.607_{\pm 0.027}~(7)$ \\
    \texttt{spectrometer} & $0.749_{\pm0.018}$ & $0.755_{\pm 0.016}~(6)$ & $0.755_{\pm 0.014}~(5)$ & $0.759_{\pm 0.008}~(4)$ & $0.767_{\pm 0.010}~(2)$ & $0.687_{\pm 0.014}~(8)$ & $0.705_{\pm 0.021} ~ (7)$ & $0.767_{\pm 0.019}~(3)$ & $0.684_{\pm 0.014}~(9)$ & $\mathbf{0.780_{\pm 0.018}~(1)}$ \\
    \texttt{oil} & $0.648_{\pm0.016}$ & $0.772_{\pm0.013}~(2)$ & $0.760_{\pm 0.027}~(4)$ & $\mathbf{0.818_{\pm0.024}~(1)}$ & $0.610_{\pm 0.015}~(9)$ & $0.708_{\pm 0.016}~(5)$ & $0.622_{\pm 0.012}~(8)$ & $0.635_{\pm 0.015}~(7)$ & $0.704_{\pm 0.023}~(6)$ & $0.772_{\pm 0.016}~(3)$ \\
    \texttt{yeast} & $0.508_{\pm 0.005}$ & $0.698_{\pm 0.020}~(3)$ & $0.716_{\pm 0.017}~(2)$ & $\mathbf{0.809_{\pm 0.010}~(1)}$ & $0.673_{\pm 0.011}~(4)$ & $0.522_{\pm 0.008}~(6)$ & $0.512_{\pm 0.005}~(8)$ & $0.515_{\pm 0.007}~(7)$ & $0.501_{\pm 0.003}~(9)$ & $0.662_{\pm 0.015}~(5)$ \\
    \texttt{car\_eval} & $0.873_{\pm 0.008}$ & $0.954_{\pm 0.007}~(4)$ & $0.970_{\pm 0.005}~(3)$ & $\mathbf{0.981_{\pm 0.005}~(1)}$ & $0.937_{\pm 0.009}~(6)$ & $0.940_{\pm 0.008}~(5)$ & $0.835_{\pm 0.015}~(9)$ & $0.840_{\pm 0.014}~(8)$ & $0.971_{\pm 0.003}~(2)$ & $0.926_{\pm 0.011}~(7)$ \\
    \texttt{us\_crime} & $0.589_{\pm 0.007}$ & $0.650_{\pm 0.009}~(3)$ & $0.638_{\pm 0.008}~(5)$ & $\mathbf{0.822_{\pm 0.008}~(1)}$ & $0.695_{\pm 0.011}~(2)$ & $0.616_{\pm 0.005}~(6)$ & $0.592_{\pm 0.005}~(8)$ & $0.590_{\pm 0.009}~(9)$ & $0.612_{\pm 0.005}~(7)$ & $0.646_{\pm 0.011}~(4)$\\
    \texttt{scene} & $0.534_{\pm 0.003}$ & $0.611_{\pm 0.007}~(3)$ & $0.602_{\pm 0.005}~(4)$ & $\mathbf{0.664_{\pm 0.008}~(1)}$ & $0.581_{\pm 0.005}~(5)$ & $0.535_{\pm 0.002}~(7)$ & $0.528_{\pm 0.000}~(9)$ & $0.530_{\pm 0.001}~(8)$ & $0.543_{\pm 0.003}~(6)$ & $0.616_{\pm 0.006}~(2)$ \\
    \texttt{abalone} & $0.847_{\pm 0.002}$ & $0.853_{\pm 0.002}~(3)$ & $0.852_{\pm 0.002}~(5)$ & $0.778_{\pm 0.002}~(8)$ & $0.564_{\pm 0.003}~(9)$ & $\mathbf{0.857_{\pm 0.002}~(1)}$ & $0.849_{\pm 0.003}~(7)$ & $0.852_{\pm 0.002}~(5)$ & $0.853_{\pm 0.003}~(4)$ & $0.853_{\pm 0.002}~(2)$\\
    \texttt{optical\_digits} & $0.807_{\pm 0.005}$ & $0.925_{\pm 0.003}~(2)$ & $0.903_{\pm 0.003}~(7)$ & $0.917_{\pm 0.004}~(4)$ & $0.910_{\pm 0.003}~(6)$ & $0.916_{\pm 0.003}~(5)$ & $0.700_{\pm 0.012}~(8)$ & $0.619_{\pm 0.012}~(9)$ & $0.917_{\pm 0.003}~(3)$ & $\mathbf{0.927_{\pm 0.004}~ (1)}$ \\
    \texttt{satimage} & $0.756_{\pm 0.004}$ & $0.782_{\pm 0.004}~(3)$ & $0.801_{\pm 0.003}~(2)$ & $\mathbf{0.838_{\pm 0.003}~(1)}$ & $0.779_{\pm 0.006}~(6)$ & $0.779_{\pm 0.003}~(5)$ & $0.749_{\pm 0.003}~(9)$ & $0.755_{\pm 0.003}~(8)$ & $0.766_{\pm 0.003}~(7)$ & $0.780_{\pm 0.003}~(4)$ \\
    \texttt{isolet} & $0.634_{\pm 0.006}$ & $0.860_{\pm 0.003}~(3)$ & $0.803_{\pm 0.006}~(6)$ & $\mathbf{0.932_{\pm 0.003}~(1)}$ & $0.929_{\pm 0.003}~(2)$ & $0.846_{\pm 0.003}~(4)$ & $0.538_{\pm 0.005}~(8)$ & $0.501_{\pm 0.001}~(9)$ & $0.795_{\pm 0.003}~(7)$ & $0.836_{\pm 0.005}~(5)$ \\
    Average rank & - & 3.75 & 4.50 & \textbf{2.67} & 4.92 & 5.00 & 7.75 & 7.08 & 5.67 & 3.58\\ 
    \bottomrule
  \end{tabular}
}
\label{tab:gmean}
\end{sidewaystable}

\end{document}